%% file: ms.tex
\documentclass[10pt,twocolumn,letterpaper]{article}

\usepackage{iccv}
\usepackage{times}
\usepackage{epsfig}
\usepackage{graphicx}
\usepackage{amsmath}
\usepackage{amssymb}
\usepackage{array}
\usepackage{subfigure}

\usepackage{amsmath,amssymb,amsfonts,dsfont,bm,mathrsfs}
\usepackage{booktabs,multirow,adjustbox}
\usepackage{lipsum,color,float}
\usepackage{graphicx}

\usepackage[pagebackref=true,breaklinks=true,colorlinks,bookmarks=false]{hyperref}
\newcommand{\st}[1]{\textcolor{black}{#1}}

\iccvfinalcopy 

\def\iccvPaperID{7542} 

\ificcvfinal\fi

\begin{document}

\title{Improving the Fairness of Deep Generative Models without Retraining}

\author{Shuhan Tan$^1$ \quad Yujun Shen$^2$ \quad Bolei Zhou$^2$\\
$^1$Sun Yat-Sen University \ \ $^2$ The Chinese University of Hong Kong\\
{\tt\small tanshh@mail2.sysu.edu.cn, \{sy116, bzhou\}@ie.cuhk.edu.hk }
}

\maketitle

\input{sections/0_abstract.tex}
\input{sections/1_Introduction.tex}
\input{sections/2_Related_Work.tex}
\input{sections/3_Method.tex}
\input{sections/4_Experiments.tex}
\input{sections/5_Conclusion.tex}

\bibliographystyle{ieee_fullname}
\small{\bibliography{ref}}

\appendix
\twocolumn[\section*{Appendix}\input{sections_supp/dist_supp}
\input{sections_supp/hyper_parameter}]
\twocolumn[\input{sections_supp/dist_plots_2}
\input{sections_supp/api_details}] 
\twocolumn[\input{sections_supp/sample}]
\newpage
\twocolumn[\input{sections_supp/sample_plots_age_eyeglasses}]
\twocolumn[\input{sections_supp/sample_plots_gender_eyeglasses}]
\twocolumn[\input{sections_supp/sample_plots_black_age}]
\twocolumn[\input{sections_supp/sample_plots_black_gender}]
\twocolumn[\input{sections_supp/sample_plots_asian_age}]
\twocolumn[\input{sections_supp/sample_plots_asian_gender}]
\twocolumn[\input{sections_supp/sample_plots_gender_blackhair}]
\twocolumn[\input{sections_supp/sample_plots_age_smiling}]
\twocolumn[\input{sections_supp/pulse_samples}]
\twocolumn[\input{sections_supp/api_samples_male}] 
\newpage
\twocolumn[\input{sections_supp/api_samples_female}]
\end{document}

%% file: sections/0_abstract.tex
\begin{abstract}
Generative Adversarial Networks (GANs) advance face synthesis through learning the underlying distribution of observed data. Despite the high-quality generated faces, some minority groups can be rarely generated from the trained models due to a biased image generation process.
To study the issue, we first conduct an empirical study on a pre-trained face synthesis model. We observe that after training the GAN model not only carries the biases in the training data but also amplifies them to some degree in the image generation process.
To further improve the fairness of image generation, we propose an interpretable baseline method to balance the output facial attributes without retraining. The proposed method shifts the interpretable semantic distribution in the latent space for a more balanced image generation while preserving the sample diversity.
Besides producing more balanced data regarding a particular attribute (\textit{e.g.}, race, gender, \textit{etc.}), our method is generalizable to handle more than one attribute at a time and synthesize samples of fine-grained subgroups.
We further show the positive applicability of the balanced data sampled from GANs to quantify the biases in other face recognition systems, like commercial face attribute classifiers and face super-resolution algorithms.%
\footnote{Project page is at \url{https://genforce.github.io/fairgen}.}
\end{abstract}

%% file: sections/1_Introduction.tex
\section{Introduction}

Artificial Intelligence (AI) is being applied to a wide range of applications in our daily life, such as employee hiring, loan granting, and criminal searching~\cite{joseph2016fairness}.
The decisions made by AI algorithms, which sometimes matter to affect the life of people, are required to be unbiased and trustworthy.
Unfortunately, current AI systems are known to have a discriminatory nature due to imbalanced training data~\cite{torralba2011unbiased,buolamwini2018gender,corbett2018measure,chen2018my} and various algorithmic factors~\cite{barocas2017fairness,srivastava2017veegan}.
Such biases within the AI models may lead to unfair treatment of people, especially those from minority groups.
As a result, fairness, together with other ethical issues, becomes a crucial AI research topic.

\input{sections/1.1_teaser_figure}

Many studies have been made to analyze and improve the fairness in classification models~\cite{Hardt2016Equality,Hee2017Improving, ryu2018inclusivefacenet,Creager2019Flexibly}.
Their primary goal is to learn a system that can give equally accurate prediction, with respect to a specific task, on every single group of people.
Along with the recent development of Generative Adversarial Networks (GANs)~\cite{gan,brock2018large,Karras2019Style,Karras2019stylegan2}, deep models have shown strong capability in many generative tasks, like style transfer~\cite{li2017universal}, semantic manipulation~\cite{stargan}, image super-resolution~\cite{wang2018esrgan}, \textit{etc}.
Therefore, it becomes critical to also examine the fairness of these generative models~\cite{kurenkov2020lessons}.
Concretely, we investigate whether an algorithm can give similar performance on different groups of people from the perspective of whether each group can be generated by the models with equal probability.

In fact, many recent generative models have been found to generate imbalanced images~\cite{zhao2018Bias, Salminen2020Analyzing, jain2020imperfect}.
To improve the fairness of image generation, existing approaches typically fall into two folds, \textit{i.e.}, pre-processing and in-processing.
The pre-processing methods try to eliminate the bias from the training data perspective (\textit{e.g.}, collecting new data or selecting a subset of the data collection to make it balanced)~\cite{mcduff2019characterizing}, while the in-processing methods target at improving the data coverage learned by the model via introducing new training objectives~\cite{xu2018fairgan,sattigeri2019fairness,choi2019fair,yu2020inclusive}.
Both kinds of approaches, however, require training new models, making them hard to apply to the numerous systems that have already been released to the public.
On the other hand, as evidenced by some recent works, pre-trained GANs can provide rich information to facilitate various downstream tasks~\cite{shen2020interfacegan,Menon_2020_CVPR,yang2019semantic,gu2020image,pan2020dgp,xu2021generative}.
Accordingly, it would be of great use if we can improve the fairness of existing models without touching the model weights.

In this work, we first conduct an empirical study on the fairness of a state-of-the-art pre-trained face synthesis GAN model. 
Fig.~\ref{fig:teaser} gives an example on the model bias related to race and gender, where the bias in the training data (blue bars) is carried and amplified by the GAN model (orange bars). Based on this observation, we propose an interpretable baseline method to alleviate the biases in pre-trained GANs, which does not require any retraining or access to the original training data. Instead, it examines the sampling process of GANs and shifts the interpretable semantic distribution in the latent space for each subgroup of interest. As shown in Fig.~\ref{fig:teaser} (green bars) and other experiments, our baseline method can very well improve the fairness of the models. 

\st{We summarize our contributions as follows.
\begin{itemize}
    \vspace{-5pt}
    \setlength{\itemsep}{2pt}
    \setlength{\parsep}{0pt}
    \setlength{\parskip}{0pt}
    \item We conduct an empirical study to reveal relationship of the bias in the training data and a well-learned GAN model. We observe that bias in the training data is carried and further amplified by the GAN model.
    \item We propose an interpretable baseline, which is able to improve the fairness of GANs not only regarding a single attribute but also across multiple attributes.
    \item We demonstrate the positive applicability of the GAN model calibrated by our method for quantifying the biases in other AI systems, \textit{i.e.}, attribute classifiers and super-solution algorithms.
\end{itemize}
}

%% file: sections/1.1_teaser_figure.tex
\begin{figure}[t]
    \centering
    \subfigure{
        \includegraphics[width=1.0\linewidth]{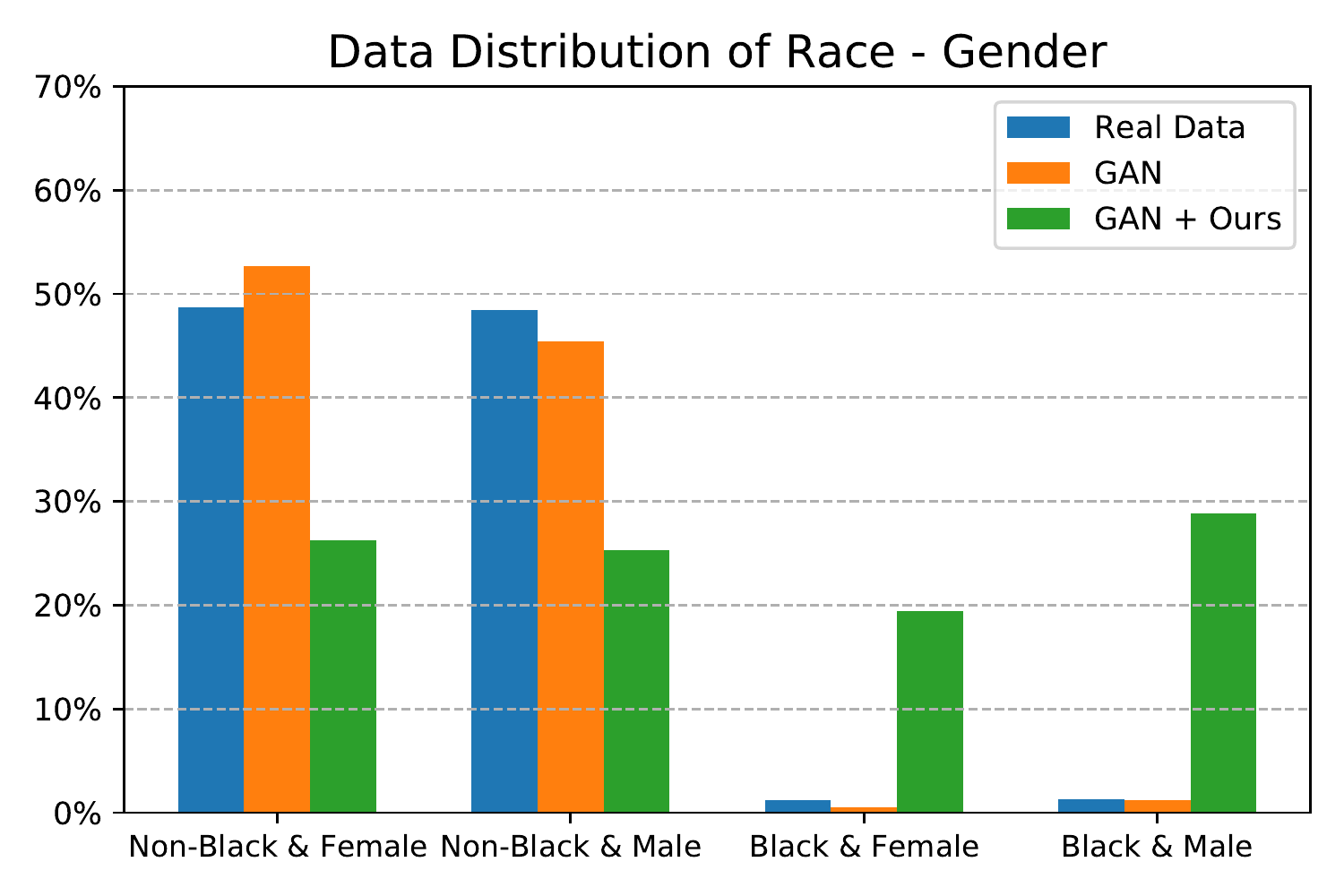}
    }
    \subfigure{
        \includegraphics[width=1.0\linewidth]{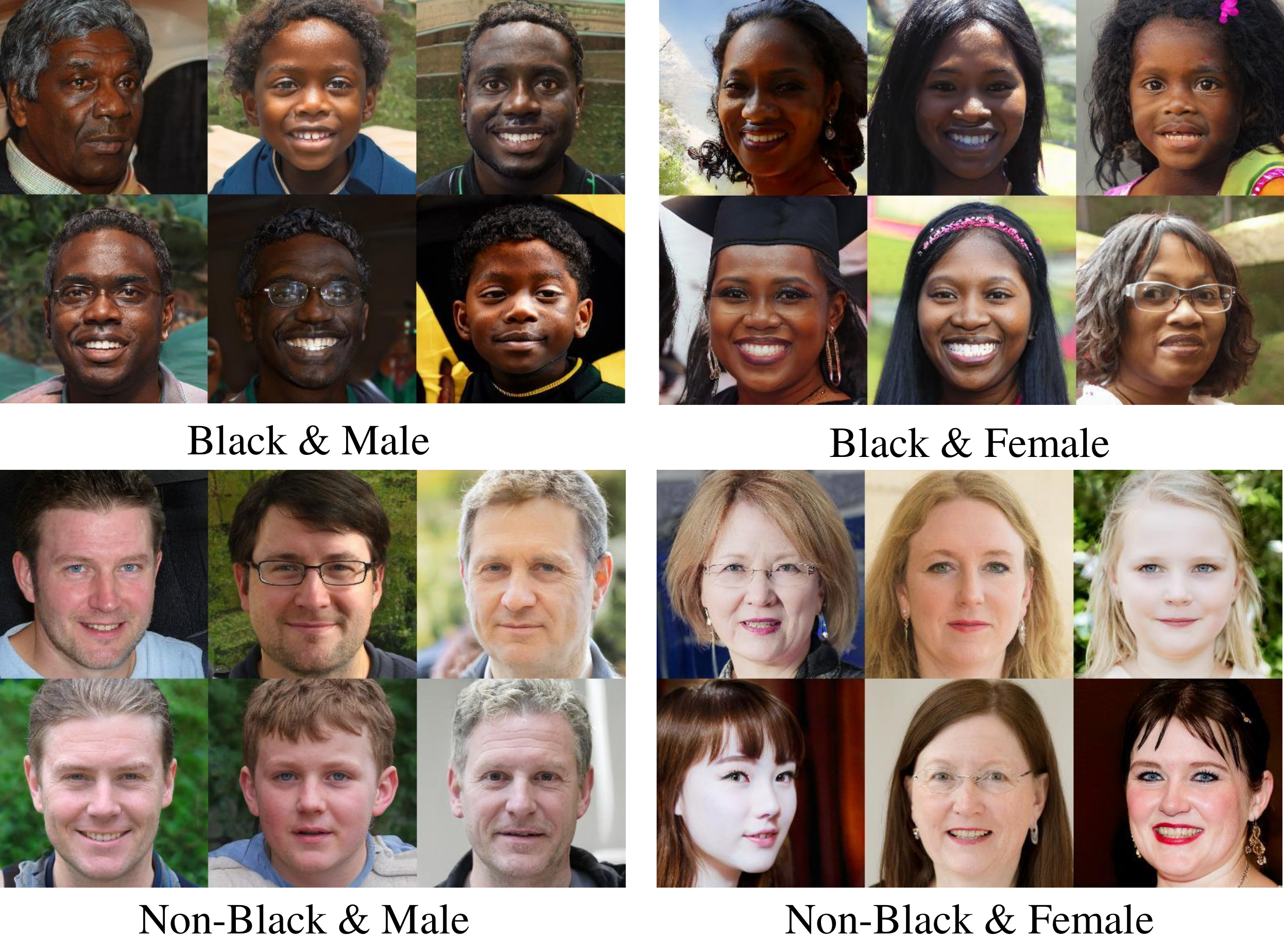}
    }
    \caption{
        \textbf{Top:} The joint distribution of two face attributes, \textit{i.e.}, race and gender, on the training data, the native synthesis from GANs, and the far more balanced synthesis after improving the model fairness with our method.
        \textbf{Bottom:} Visual samples for each subgroup with our proposed sampling scheme.
        The \textit{fixed} StyleGAN2 model pre-trained on FF-HQ dataset~\cite{Karras2019stylegan2} is used.
    }
    \vspace{-10pt}
    \label{fig:teaser}
\end{figure}

%% file: sections/2_Related_Work.tex
\section{Related Work}

\noindent\textbf{AI Fairness.}
AI fairness has attracted wide attention in recent years~\cite{barocas2017fairness,buolamwini2018gender,kurenkov2020lessons}.
Most existing works focus on studying the fairness of discriminative models (\textit{e.g.}, face attribute classifiers), where fairness is achieved when the model makes predictions in a non-discriminatory way.
For example, an age predictor is expected to achieve similar performance on different genders.
Three types of approaches have been proposed: the pre-processing methods try to collect balanced training data~\cite{Zemel2013Learning,Lum2016ASF,Christos2016Fair}, the in-processing methods introduce constraints or regularizers into the training process~\cite{Blake2017Learning,Zafar2017FairnessCM,pmlr-v80-agarwal18a}, and the post-processing methods attempt to modify the posteriors of pre-trained models~\cite{Feldman2015ComputationalFP,Hardt2016Equality}.
Compared to discriminative models, which are primarily designed to make inference on existing data, generative models are able to create new data, enabling a lot more application scenarios~\cite{stargan,wang2018esrgan,gu2020image,pan2020dgp}.
But little efforts have been made to explore the fairness of generative models and existing methods typically focus on the pre-processing stage and the in-processing stage.
This work fills this hole by improving the fairness of well-learned GAN models without retraining.

\noindent\textbf{Fairness of Generative Models.}
Studying the fairness of generative models is as important as studying that of discriminative models.
It has been shown that the data produced by a fair generative model can benefit various downstream classifiers~\cite{xu2018fairgan,sattigeri2019fairness}.
However, Zhao \textit{et al.}~\cite{zhao2018Bias} find that GAN will usually carry and amplify the bias existing in the training data.
Jain \textit{et al.}~\cite{jain2020imperfect} further analyze this problem from the perspective of mode collapse. 
Some attempts have been made to reduce the model bias and learn a more balanced image generation~\cite{choi2019fair,yu2020inclusive,mcduff2019characterizing}.
Choi \textit{et al.}~\cite{choi2019fair} propose to learn a weighting function to reweigh the importance of each instance during the training of GANs.
Yu \textit{et al.}~\cite{yu2020inclusive} mitigate bias by increasing the data coverage learned by GANs.
McDuff \textit{et al.}~\cite{mcduff2019characterizing} carefully balance the training set such that all groups of interests possess similar number of samples.
Our \textbf{improvements} are:
1) We propose a fair sampling pipeline by shifting the latent distribution of a well-learned GAN model instead of training new models or tuning the model weights, which can be often costly. In this way, our approach can be applied to numerous existing models with minor effort.
2) We do not rely on the full access of the original training dataset, which is often not available for end-users.
3) Our algorithm is flexible such that we can easily include a new subgroup of interests and still maintain the fairness.

%% file: sections/3_Method.tex
\section{Toward Fair Image Generation with GANs}

\input{sections/1.2_distribution}
\input{sections/4.0_Fair_Sampling_Table}

\subsection{Problem Setting}
\noindent \textbf{Background}. We assume there exists an unknown data distribution ${P_\text{data}(\mathcal{X})}$ over observed $d$-dimensional image data $\mathcal{X} \subseteq \mathbb{R}^{d}$. 
The goal for GAN is to learn distribution $P_\theta({\mathcal{X}})$ such that $P_\theta({\mathcal{X}}) = {P_\text{data}(\mathcal{X})}$. Given a well-trained GAN model, its generator can be seen as a mapping function $g_\theta: \mathcal{Z} \rightarrow \mathcal{X}$. Here, $\mathcal{Z} \subseteq \mathbb{R}^{n}$ denotes the $n$-dimensional latent space, which is usually assumed as Gaussian distribution $\mathcal{N}(\mathbf{0}, \mathbf{I}_d)$. 
With this model trained on the dataset $D_{\text{train}}$ sampled from ${P_\text{data}(\mathcal{X})}$, we are able to obtain a generated distribution $P_{\theta} (\mathcal{X}) \approx {P_\text{data}(\mathcal{X})}$. 
By sampling latent codes $\textbf{z}$, we can generate a realistic dataset $D_{\theta} = \{ g_{\theta} (\mathbf{z}_i)\}_{i=1}^N$
with data size $N$.

Besides, we assume that each of the image sample $x$ in $\mathcal{X}$ contains specific semantic attributes, like age and gender of the face in the image.
In the context of fairness, we focus on $m$ target binary attributes $\mathcal{A}_t$, which we aim to achieve fairness over their distributions. Also, we have $m'$ binary context attributes $\mathcal{A}_c$ that do not require fairness. 
Suppose we have a attribute classifier as the semantic scoring function $f_S: \mathcal{X} \rightarrow \mathcal{S} \subseteq \mathbb{R}^{m+m'}$, where $\mathcal{S}$ is the score space for all the attributes. It can be mapped to binary labels with unit step function $H(\cdot)$, such that $H (x) = 1$ if $x \geq 0$, otherwise $H (x) = 0$.
This allows us to quantify the bias in terms of specific attributes by mapping the latent code $\mathbf{z}$ to its attribute label $\mathbf{a}$ with $\mathbf{a} = H(f_S(g_{\theta}(\mathbf{z}))) $. 

\subsection{Measuring Data Bias and Generation Bias}

\st{In this section, we conduct an empirical study on the relationship between the bias in the training data and in a well-trained GAN model. To this end, we also develop metrics to measure the data generation bias.}

\noindent \textbf{Generation Bias}. Real-world datasets often follow a long-tail distribution, thus carry various kinds of biases. The GAN model trained on real data $D_{\text{train}}$ will therefore produce a dataset $D_{\theta}$ that highly likely carries the biases in the real data. Meanwhile, training GAN is prone to the mode collapse issue~\cite{srivastava2017veegan} which potentially brings in more biases. 
If the GAN models is biased, then in $D_{\theta}$ the marginal distribution of target attributes $P_{D_{\theta}}(\mathcal{A}_t )$ will be highly imbalanced. We measure the imbalance with
\begin{equation}\label{imbalance_score}
    f_u (D) = KL \left( P_{D_{\theta}} (\mathcal{A}_t) \, \middle\| \,\mathcal{U}(\mathcal{A}_t) \right),
\end{equation}
where $KL$ is the Kullback-Leibler divergence and $\mathcal{U}$ denotes the uniform distribution. A higher $f_u(D)$ indicates a imbalanced distribution in $D$.

To empirically show the bias introduced by the training process of GAN model, we computed the imbalance measurement $f_u$ for a state-of-the-art face synthesis GAN model (StyleGAN2~\cite{Karras2019stylegan2}), which is well trained on the FF-HQ dataset~\cite{Karras2019Style}. We first show the bias results w.r.t a single facial attribute (\textit{e.g.} Gender) in Table~\ref{table:fair_1}. We observe that: 1) there is a strong correlation between the imbalance degree in the training data and in the GAN's output, indicating that bias is carried from the training data to the GAN model; 2) the imbalance in the GAN's output is consistently more significant than in the training data, which shows that bias is amplified during GAN's training process. Then, we show bias results w.r.t the combination of two and three facial attributes in Table~\ref{table:fair_2} and Table~\ref{table:fair_3} respectively. We find that when we consider multiple facial attributes simultaneously, the bias in the GAN output becomes much more severe. In the next subsection, we will develop an effective baseline method to diminish such biases.

\noindent \textbf{Measuring the Fairness of Image Generation}. 
To reduce the bias in $D_{\theta}$, one could either acquire a balanced training set~\cite{mcduff2019characterizing} or retrain the GAN model with new objectives~\cite{choi2019fair, yu2020inclusive}. However, both approaches are costly and require access to the original training dataset and hyper-parameters.
Instead, we propose a new sampling strategy that could generate a fair dataset $D_{\text{fair}}$ from the same GAN model. 

On one hand, our main objective is to make the marginal distribution of the \textit{target} attributes in the fair dataset $P_{D_{\text{fair}}} (\mathcal{A}_t)$ as close to the uniform distribution as possible, \textit{i.e.}, to reduce $f_u(D_{\text{fair}})$ in Eq.~\eqref{imbalance_score}.
On the other hand, to ensure we do not introduce new bias w.r.t other attributes, we need to preserve the conditional distribution of the \textit{context} attributes $\mathcal{A}_c$ in $D_{\theta}$. 
Thus we formally define the fairness discrepancy $f$ as
\begin{equation}\label{fairness_score}
    f (D_{\text{fair}}) = f_u(D_{\text{fair}}) + \beta KL \left( P_{D_{\text{fair}}}(\mathcal{A}_c | \mathcal{A}_t
) \, \middle\| \, P_{D_{\theta}}(\mathcal{A}_c | \mathcal{A}_t) \right).
\end{equation}
where the target fair dataset $D_{\text{fair}}$ is sampled from a generative model $g_\theta$ w.r.t attributes $\mathcal{A}_t$ and $\mathcal{A}_c$, and $\beta$ is used to balance the importance between the main objective of target fairness and the preservation constraint on context attributes. The lower $f(D_{\text{fair}})$ is, the more fair the generated dataset is with respect to $\mathcal{A}_t$.

\subsection{Improving the Fairness of Image Generation via Shifting Latent Distribution}\label{subsec:fairgen}

\input{sections/3.1_Framework}
Based on the previous empirical analysis, we can see that pretrained GAN model carries and amplifies the bias in the imbalanced training data. It is impossible to fix all the biases in the models, and we are fully aware that any new debiasing methods might introduce some sort of new bias implicitly. Thus the goal of this work is not to fully solve the bias issue in GANs, instead, we aim at providing a new baseline to improve the fairness of GAN's image generation to some degree through shifting the latent variable distribution. This baseline method, termed as Latent Distribution Shifting (\textbf{LDS}), utilizes the interpretable semantic dimensions identified in the latent space, then shifts the latent distribution for a more fair output.

The objective of our method is to sample a set of latent codes $\mathbf{Z}_\text{fair}=\{ \mathbf{z}_i \}_{i=1}^N$, such that the GAN generated dataset $D_{\text{fair}} = \{ g_{\theta} (\mathbf{z}_i)\}_{i=1}^N$ achieves low $f(D_{\text{fair}})$. 
To construct the fair latent code set $\mathbf{Z}_\text{fair}$ that makes $P_{D_{\text{fair}}} (\mathcal{A}_t)$ close to $\mathcal{U}(\mathcal{A}_t)$, we propose to sample a set of latent codes \textit{conditioned} on each of the possible value $ \mathbf{a} \in \mathcal{A}_t$.

Specifically, for $\mathcal{A}_t$ with $m$ binary attributes, we have $K=2^m$ possible attribute values. 
For the $i_\text{th}$ possible attribute value $\mathbf{a}$, our goal is to sample a set of latent codes $\mathbf{Z}_i$ such that $H(f_S(g_{\theta}(\mathbf{z}))) = \mathbf{a}, \forall \mathbf{z} \in \mathbf{Z}_i$ and $\left|\mathbf{Z}_i\right|=N / K$. 
Then we can simply compose a fair latent code set with $\mathbf{Z}_\text{fair} = \{\mathbf{Z}_i\}_{i=1}^{K}$ such that $P_{D_{\text{fair}}} (\mathcal{A}_t) = \mathcal{U}(\mathcal{A}_t)$. This process is illustrated in Fig~\ref{fig:framework}.
The most critical component of this approach is attribute-conditioned latent code sampling. We split this process into three steps:
\vspace{-2pt}
\begin{enumerate}
    \setlength{\itemsep}{0pt}
    \setlength{\parsep}{0pt}
    \setlength{\parskip}{0pt}
    \item Create an intermediate code set $\mathbf{Z}_{\text{edit}}$ by manipulating random latent codes towards specified condition $\mathbf{a}$. 
    \item Filter and fit the distribution of $\mathbf{Z}_{\text{edit}}$ with the attribute scoring function and a Gaussian Mixture Model $q_{\Phi} (\mathbf{z})$. 
    \item Construct $\mathbf{Z}_i$ by sampling $\mathbf{z} \sim q_{\Phi} (\mathbf{z})$.
\vspace{-2pt}
\end{enumerate}
\vspace{-2pt}

We introduce each step in more details as follows.

\subsubsection{Shifting Semantic Distribution in Latent Space} \label{method:InterfaceGAN}

\noindent \textbf{Manipulating Attributes in the Latent Space}. 
InterFaceGAN is based on the assumption that for any binary semantic attribute $\mathcal{A}_{t, i}$, there exists a hyperplane in the latent space $\mathcal{Z}$ as the separation boundary. Latent codes on the same side of the boundary have the same attribute value, which turns into the opposite when the latent codes cross the boundary. 

Assume the boundary for semantic $\mathcal{A}_{t, i}$ has the unit normal vector $\mathbf{n}_i$, we define its signed distance with a latent code $\mathbf{z}$ as $d(\mathbf{z}, \mathbf{n}_i) = \mathbf{n}_i^T \mathbf{z}$. Given $f_i$ as the scoring function for attribute $\mathcal{A}_{t, i}$, when $\mathbf{z}$ move towards or away from $\mathbf{n}_i$, both $d(\mathbf{z}, \mathbf{n}_i)$ and $f_i(g_\theta(\mathbf{z}))$ would vary accordingly. Furthermore, when $\mathbf{z}$ move across the boundary, both $d(\mathbf{z}, \mathbf{n}_i)$ and $f_i(g_\theta(\mathbf{z}))$ would change their numerical signs. Therefore, the authors assume a linear relationship such that

\begin{equation} \label{equ:liner_semantic_relation}
f_i(g_\theta(\mathbf{z})) = \lambda d(\mathbf{z}, \mathbf{n}_i),
\end{equation}
where $\lambda > 0$ measures the ratio of changing speeds of semantic score and distance.
According to this relationship, to manipulate the attribute score, we can easily vary the original code with
$\mathbf{z}_{\text{edit}} = \mathbf{z} + \alpha \mathbf{n}_i$, as $f_i(g_\theta(\mathbf{z}_\text{edit})) = f_i(g_\theta(\mathbf{z})) + \lambda \alpha$. Please refer to Sec.~\ref{exp:fair_sampling} for the detail about how we obtain $\mathbf{n}_i$ in practice.

Our goal is to make $f_i(g_\theta(\mathbf{z}_{\text{edit}, i})) = \lambda \alpha$, where $\lambda \alpha > 0$ is a predefined scoring threshold such that we can classify the synthesised image with high confidence. Towards this goal, we first move $\mathbf{z}$ onto the decision boundary and then move it away from the boundary with magnitude $\alpha$. Formally, we have
\begin{equation} \label{equ:attr_magnitude_single}
    \mathbf{z}_{\text{edit}, i} = \mathbf{z} - d(\mathbf{z}, \mathbf{n}_i) \mathbf{n}_i + \alpha \mathbf{n}_i.
\end{equation}
According to Eq.~\eqref{equ:liner_semantic_relation}, we can prove that $f_i(g_\theta(\mathbf{z}_{\text{edit}, i})) = \lambda \alpha$. The same conclusion holds for $\mathbf{a}_i=0$, where $\alpha < 0$. This enables us to edit $\mathbf{a}_i$ for any $\mathbf{z} \sim \mathcal{N}(\mathbf{0}, \mathbf{I}_d)$.

The next step is to set attribute value $\mathbf{a}$ for all $m$ target attributes $\mathcal{A}_m$. To this end, we need the normal vectors of any pair of the semantic boundary to satisfy $\mathbf{n}_i^{\text{T}} \mathbf{n}_j \approx 0$, which we find to be the case for the GAN model and attributes we consider. This is because these attributes are already well disentangled in the latent space of the pre-trained model. In this way, we can set
\begin{equation} \label{equ:attr_magnitude_multiple}
    \mathbf{z}_{\text{edit}} = \mathbf{z} - \sum_{i=1}^m  [ d(\mathbf{z}, \mathbf{n}_i) \mathbf{n}_i - \alpha_i \mathbf{n}_i ],
\end{equation}
where $\alpha_i>0$ if $\mathbf{a}_i=1$ otherwise $\alpha_i<0$. It is also easy to prove that $f_i(g_\theta(\mathbf{z}_{\text{edit}})) \approx \lambda \alpha_i$ for arbitrary attribute $i$.
In this way, if Eq.~\eqref{equ:liner_semantic_relation} is satisfied, we are able to obtain 
\begin{equation} \label{equ:manipulation_obj}
H(f_S(g_{\theta}(\mathbf{z}_\text{edit}))) = \mathbf{a}.    
\end{equation}

For a specified attribute subgroup $\mathbf{a}$, we can construct an intermediate code set $\mathbf{Z}_{\text{edit}}$ with $N_{\text{edit}}$ samples. This is done by simply sampling $N_{\text{edit}}$ latent codes $\mathbf{z}$ from $\mathcal{N}(\mathbf{0}, \mathbf{I}_d)$ and mapping them into $\mathbf{z}_{\text{edit}}$ with Eq.~\eqref{equ:attr_magnitude_multiple}.

\vspace{-5pt}

\subsubsection{Conditional Latent Space Modeling}
To support re-sampling latent code from a specific subgroup, we utilize a Gaussian Mixture Model (GMM) to fit its distribution in the latent space with the set of latent code $\mathbf{Z}_{\text{edit}}$ created with Eq.~\eqref{equ:attr_magnitude_multiple}.

However, as the linear relationship in Eq.~\eqref{equ:liner_semantic_relation} may not be the case for some latent codes and attributes in practice, not all samples from $\mathbf{Z}_{\text{edit}}$ satisfy Eq.~\eqref{equ:manipulation_obj}. To obtain more accurate distribution modeling, we further use $f_S$ to filter out less confident samples for creating a new set: $\mathbf{Z}_{\text{edit}}' = \{\mathbf{z} \in \mathbf{Z}_{\text{edit}} | H(f_S(g_{\theta}(\mathbf{z}))) = \mathbf{a}\}$. 

We then train a GMM model on $\mathbf{Z}_{\text{edit}}'$ with expectation-maximization (EM) algorithm to obtain a probabilistic model of latent codes $q_{\Phi}(\mathbf{z})$ conditioned on the specified subgroup $\mathbf{a}$. This model enables us to sample an arbitrary number of high-quality images from a certain subgroup.

To construct the fair latent code set $\mathbf{Z}_\text{fair}$, we firstly prepare the GMM model for each of the possible value in $\mathcal{A}_t$. We then compose $\mathbf{Z}_\text{fair}$ with the same-size latent code set sampled from each of the GMM models.

%% file: sections/1.2_distribution.tex
\begin{table*}[htbp]
  \centering
    \caption{Fairness analysis on different datasets with single attribute}
    \vspace{2pt}
    \begin{tabular}{c*{7}{|p{1.7cm}<{\centering}}}
    \hline
    Attributes & Eyeglasses & Age & Smiling & Gender & Black & Asian & White \\
    \hline
    FFHQ Dataset & 0.191 & 0.034 & 0.002 & $9.40\times10^{-7}$ & 0.576 & 0.279 & 0.042 \\
    GAN & 0.246 & 0.059 & 0.040 & 0.002 & 0.603 & 0.319 & 0.057 \\
    \hline
    StyleFlow~\cite{abdal2020styleflow} & 0.101 & 0.001 & 0.003 & 0.004 & - & - & - \\
    Ours & $5.65 \times10^{-4}$& $9.68 \times10^{-6}$ & 0 & $1.28\times10^{-6}$ & $3.45\times10^{-4}$ & $5.13\times10^{-6}$ & $2.00\times10^{-8}$ \\
    \hline
    \end{tabular}%
  \label{table:fair_1}%
\end{table*}%

%% file: sections/4.0_Fair_Sampling_Table.tex
\begin{table*}[!ht]
  \centering
    \caption{Fairness analysis on different datasets with two attributes.}
    \begin{tabular}{c*{7}{|p{1.7cm}<{\centering}}}
    \hline
    \multirow{2}{*}{Attributes} & \multirow{2}{*}{Age-Gender} & Age-Eyeglasses & Gender-Eyeglasses & Black-Gender & Asian-Gender & \multirow{2}{*}{Black-Age} & \multirow{2}{*}{Asian-Age} \\
    \hline
    FFHQ Dataset & 0.0930 & 0.2994 & 0.2228 & 0.5762 & 0.2790 & 0.6097 & 0.3125 \\
    GAN & 0.1205  & 0.4079  & 0.2808  & 0.6075  & 0.3305  & 0.6622  & 0.3919  \\
    \hline
    StyleFlow~\cite{abdal2020styleflow} & 0.1811 & 0.1755 & 0.0980 & - & - & - & - \\
    Ours & \textbf{0.0201}  & \textbf{0.0079} & \textbf{0.0013} & \textbf{0.0102} & \textbf{0.0007}  & \textbf{0.0033} & \textbf{0.0018} \\
    \hline
    \end{tabular}%
  \label{table:fair_2}%
\end{table*}%

%% file: sections/3.1_Framework.tex
\begin{figure*}[!t]
    \centering
    \includegraphics[width=1.0\linewidth]{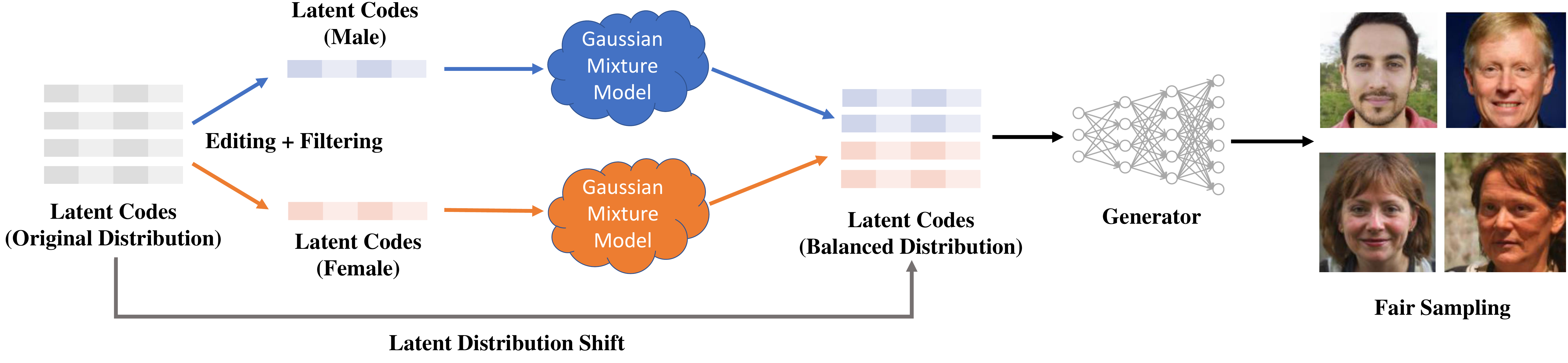}
    \caption{
        Overview of the proposed baseline to shift the latent distribution of a well-trained GAN model.
        Starting from the original distribution, we manage to collect a set of latent codes, which are able to synthesize balanced images across all subgroups of interests.
        The detailed procedure of the latent code collection can be found in Sec.~\ref{subsec:fairgen}.
        Then, for each subset of latent codes, a Gaussian Mixture Model (GMM) is used to fit the sub-distribution.
        Finally, we integrate these GMMs together as a balanced distribution, which naturally supports conditional sampling for any particular subgroup and hence improves the model fairness.
    }
    \label{fig:framework}
\end{figure*}

%% file: sections/4_Experiments.tex
\input{sections/4.1_Fair_Sampling}
\section{Experiments}
\label{exp:general}
We evaluate LDS with a well-trained face synthesis GAN model, StyleGAN2~\cite{Karras2019stylegan2}. Specifically, in Sec.~\ref{exp:fair_sampling} we investigate how effective LDS can improve the fairness of an existing GAN model w.r.t different attribute settings. Then, in Sec.~\ref{exp:application}, to show the potential impact of LDS, we utilize our generated fair dataset to reveal and quantify the biases in two commercial facial classification APIs as well as a state-of-the-art super-resolution model. Finally, in Sec.~\ref{exp:analysis}, we conduct an ablation study of the design choices used in our model.

\noindent\textbf{Implementation Details.} 
For the GAN model $g_\theta$ we study, we use the official StyleGAN2 model\footnote{\url{https://github.com/NVlabs/stylegan2}.} trained on the FF-HQ dataset~\cite{Karras2019Style}. This model is able to generate realistic human face images with $1024\times1024$ resolution. For common attributes (\textit{age, gender} and \textit{eyeglasses}) we use an off-the-shelf classifier trained on Celeb-A dataset~\cite{liu2015faceattributes} as the scoring function $f_S$; for the racial attributes, we use a classifier trained on LFW dataset~\cite{Gary2007labeled}. Indeed these classifiers themselves might be biased or give some wrong predictions; for generality, we assume they are reasonably good to use.

The style-based generator in StyleGAN2 learns to first map the latent code from $\mathcal{Z}$ space to another high-dimensional space $\mathcal{W}$. As shown in~\cite{shen2020interfacegan}, latent codes in $\mathcal{W}$ space have much stronger disentanglement property than in $\mathcal{Z}$, and latent code manipulation quality is also better in $\mathcal{W}$ space. Therefore, in our framework, we directly operate and sample codes from the $\mathcal{W}$ space, which is simply mapped from $\mathcal{Z}$ space with StyleGAN2 mapping module.

To obtain the semantic boundaries $\mathbf{n}$ used in Eq.~\eqref{equ:attr_magnitude_single}, we first synthesize a dataset $\mathcal{D}_{\textit{direct}}$ with $50K$ images by directly sampling from the original latent space. Then, we use the scoring function $f_S$ to obtain the attribute scores for all the sensitive attributes. For each attribute, we sort the corresponding scores and choose the ones with top $2\%$ highest scores as positive examples and top $2\%$ lowest scores as negative examples. This is to select the most representative examples as the scoring function may not be absolutely accurate. Finally, we use these examples to train a linear SVM to obtain the decision boundary, the normal direction of which results in $\mathbf{n}$. The SVM is trained to take the latent codes in $\mathcal{W}$ space as input, and output binary labels obtained with $f_S$.

For InterFaceGAN, we set the magnitude of code editing factor$\left|\alpha\right|=3.0$. For the conditional latent space modeling of each attribute subgroup, we manipulate and generate $\mathcal{Z}_\textit{edit}$ with $N_{\text{edit}}=2.5K$ samples; then we use a GMM model with $k=10$ components to fit the distribution. 
We provide an empirical analysis of these hyper-parameters in the \textbf{Supplementary Material}.

\input{sections/4.2_API_Gender_Compare} 
\input{sections/4.3_PULSE_compare}

\subsection{Fair Image Generation} \label{exp:fair_sampling}
We firstly show the existing bias in the GAN model w.r.t different attribute settings. Then, we present both the quantitative and qualitative results that show our LDS method can significantly improve the fairness of image generation while preserving the image quality.

\noindent\textbf{Experiment Setting.} For evaluation, We use 3 common attributes (\textit{age, gender} and \textit{eyeglasses}) and the race attributes (\textit{Black}, \textit{Asian} and \textit{White}). To form different sampling tasks, we combine $n$ of the attributes as a pair of target attributes $\mathcal{A}_t$ to form subgroups for each task while leaving the other attributes in the context set $\mathcal{A}_c$. For each of the compared sampling methods, we sample 10K images in total and evaluate on this generated dataset. We then compute the generative sampling fairness discrepancy $f$ with Eq.~\eqref{fairness_score} for each of the datasets, where we set $\beta=0.1$ in $f$. We conduct experiments with $n=1, 2, 3$, respectively.

\noindent\textbf{Quantitative Evaluation.}
We compare LDS with two baselines. 1) \textbf{GAN}: we directly sample latent codes from the original distribution of the GAN. 2) \textbf{StyleFlow}~\cite{abdal2020styleflow}, which trains conditional continuous normalizing flows to support attribute-conditional sampling from the GAN model. Here we use the official-released StyleFlow model\footnote{\url{https://github.com/RameenAbdal/StyleFlow.}} to uniformly sample latent codes from each attribute subgroup in a way similar to our method. Note that this model currently does not support latent code sampling conditioned on the race attributes.

We show the fairness results with different numbers of attributes in Tab.~\ref{table:fair_1}, Tab.~\ref{table:fair_2}, and Tab.~\ref{table:fair_3} respectively. The results show that: 1) LDS can significantly improve the fairness of GAN model across various attribute combinations. Furthermore, it is able to easily handle more than one attribute at one time. For example, for \textit{age-eyeglasses} task in Tab.~\ref{table:fair_2}, LDS significantly decreases the bias score $f$ of GAN model from 0.4079 to 0.0079, showing the high effectiveness of LDS. 2) LDS consistently outperforms StyleFlow across multiple attribute combinations.

\noindent\textbf{Qualitative Evaluation.}
Fig.~\ref{fig:fair_qualitative_1} and Fig.~\ref{fig:fair_qualitative_2} plot the generated images from the codes sampled by LDS for subgroups regarding two and three face attributes. It suggests that LDS performs well to generate images for all the subgroups with correct attributes while retaining high image quality and diversity. This quality enables us to use the generated balanced dataset to examine the fairness of other visual tasks and models. Please refer to the \textbf{Supplementary Material} for image examples from more attribute subgroups.

\subsection{Quantifying Bias in Existing Models} \label{exp:application}
The fair image generation achieved by LDS can be useful in many applications. Here we apply our method to reveal and quantify the potential biases in the existing face classifiers and a super-resolution model. 

\noindent\textbf{Bias in Face Classifiers.} We first study the bias in existing face classifiers. To make it more practical, we select two state-of-the-art commercial face attribute classification APIs (Face++ Detect API and Azure Facial Recognition). To analyze the bias problem, we focus on gender classification under different attribute conditions: \textit{age, eyeglasses} and \textit{race}. 

Specifically, for each attribute subgroup (\textit{e.g.}, Young Male), we use our method to generate a subgroup dataset with $2.5K$ images. Then, we run the face classifiers on these datasets and compare the error rate of different groups.

We show the results in Tab.~\ref{tab:api_gender}, which well quantifies the bias problem in the two APIs. In this table, besides the error rate of gender in each subgroup, we also compute the average error rates for male and female people over all the subgroups. We first observe that for both APIs, the gender classification accuracy for females is significantly lower than for the males. Also, people with black skin color are more likely to be wrongly classified, while accuracy is more balanced w.r.t age. We show some of the failure cases observed by our model in Fig.~\ref{fig:API}.

\noindent\textbf{Bias in Super-resolution Model.} In this part, we study the bias problem of super-resolution method PULSE~\cite{Menon_2020_CVPR}. This recent neural network model takes a low-resolution (LS) image as input and outputs a high-resolution (HS) image. It has been found that for certain minority groups, their rare attribute values in the LS input will often be changed to more common values in the HS output~\cite{kurenkov2020lessons}. We aim to use the images generated with LDS to examine which attributes will be more likely to be altered by PULSE.

To this end, we first generate images from a certain subgroup with LDS, and then input its down-sampled LS ($32\times32$) version to PULSE to obtain a HS ($1024\times1024$) output. Then, we use the scoring function $f_S$ to obtain the attribute values of the HS images. Finally, we compare this result with the original image's attribute value, and compute the rate of PULSE alternation for each of the attributes. Here we select four attributes \textit{Gender}, \textit{Eyeglasses}, \textit{Age} and \textit{Race}. We exhibit some examples of the image attribute alternated by PULSE in Fig.~\ref{fig:PULSE}. We can see the PULSE wrongly alters the gender, glasses and race attributes through the super-resolution process.

We show the rate of attribute alternation of PULSE in Tab.~\ref{tab:PULSE}. Here each number represents the alternation rate in a subgroup. We have the following fairness analysis: 1) For the race attributes, the alternation rate is much higher for people with Black ($76.4\%$) and Asian ($34.3\%$) race, while few images ($0.5\%$) with White race people are alternated. This indicates PULSE is prone to output people with white race.  2) For the gender attribute, the alternation rate is higher for female people, indicating PULSE is prone to output male faces. 3) For the eyeglasses attribute, we observe a very high alternate rate ($89.8\%$) for input images with eyeglasses, which shows that PULSE often fails to preserve the eyeglasses. 

It is worth noting that none of the above bias analysis requires additional human labeling or careful dataset balancing for each of the attributes, making the bias analysis based on LDS much more convenient to run than the previous bias analysis works~\cite{buolamwini2018gender, mcduff2019characterizing}. 
With the support to sample images from any attribute subgroup, LDS enables us to do such detailed analysis with low cost on the biases in existing models w.r.t different subgroups. 
In addition, we are able to easily extend the analysis to other attributes when provided with scoring function $f_S$ of new attributes. This makes LDS a general and flexible tool for studying the bias of visual models.

\subsection{Ablation Study} \label{exp:ablation}
In this section, we analyze the impact of different design choices in LDS on the fairness score. Specifically, we focus on three variations of LDS by ablating different components: 1) \textbf{Ours w/o Edit}: We use $f_S$ to filter latent codes for different subgroups with the latent code \textit{directly} sampled from the original distribution; then fit GMMs with these latent codes; 2) \textbf{Ours w/o Filter}: We \textit{skip} the subgroup attribute filter before using GMM to fit the shifted latent codes for each subgroup; 3) \textbf{Ours w/o GMM}: We generate the dataset by directly using the latent coded generated in Sec.~\ref{method:InterfaceGAN}. 

We show the results on three tasks in Figure~\ref{tab:fair_ablation}, which proves that all the components in our method are important to fairness performance of the generated dataset. Particularly, we find the latent code shifting (editing) process is very important in all the cases. This is because we are able to obtain more diverse sets of latent codes for the rare subgroups compared with original latent code distribution.

\input{sections/4.4_Analysis} 

%% file: sections/4.1_Fair_Sampling.tex
\begin{table}[!ht]
  \centering
    \caption{Fairness analysis on different datasets with three attributes.}
    \begin{tabular}{c*{3}{|p{1.7cm}<{\centering}}}
    \hline
    \multirow{3}{*}{Attributes} & Age- & Black- & Asian- \\
     & Glasses-& Age- & Age- \\
     & Gender & Gender & Gender \\
    \hline
    FFHQ & 0.3690 & 0.6694 & 0.3721 \\
    GAN & 0.4771 & 0.7260 & 0.4575 \\
    \hline
    StyleFlow~\cite{abdal2020styleflow} & 0.3294 & - & - \\
    Ours & 0.0170 & 0.0159 & 0.0077 \\
    \hline
    \end{tabular}%
  \label{table:fair_3}%
  \vspace{-5pt}
\end{table}%

\begin{figure*}[!ht]
    \centering
    \includegraphics[width=1.0\linewidth]{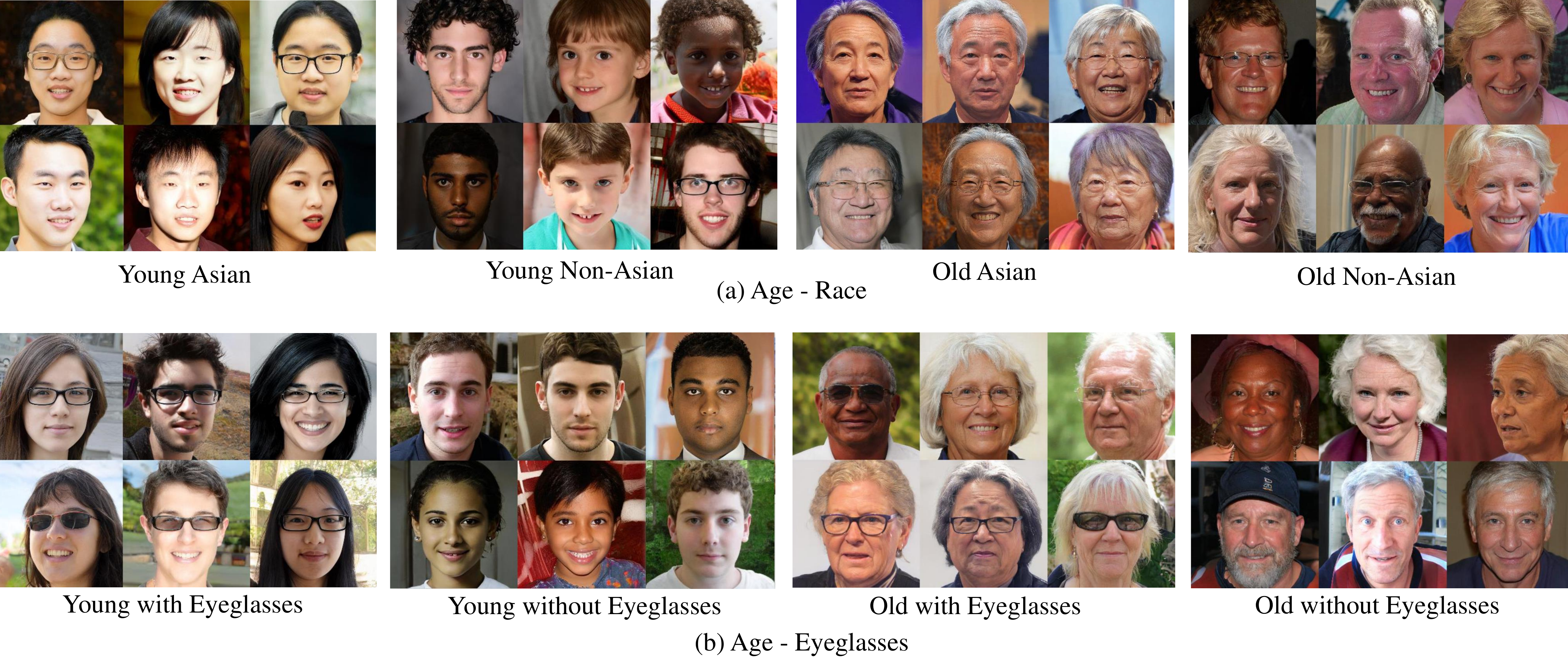}
    \caption{Qualitative results for fair image generation in GANs with two-attribute values.}
    \label{fig:fair_qualitative_1}
\end{figure*}

\begin{figure*}[!ht]
    \centering
    \includegraphics[width=1.0\linewidth]{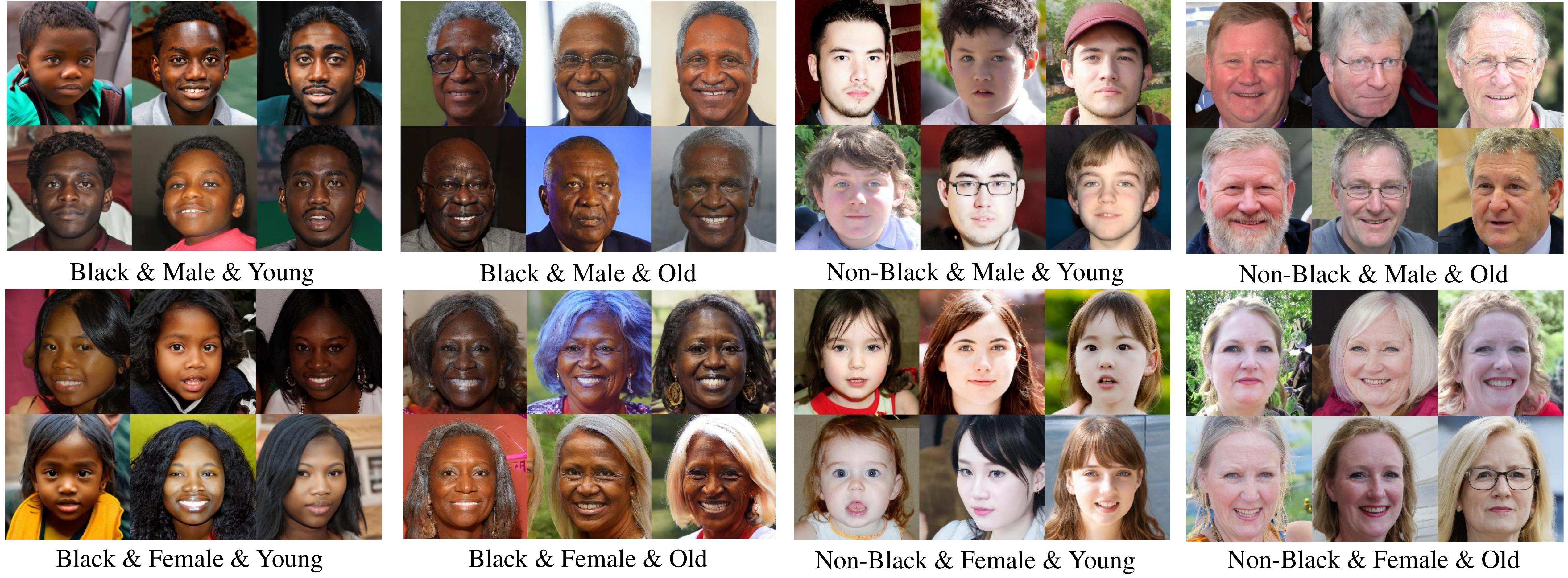}
    \caption{
        Qualitative results of fair image generation with respect to three different facial attributes, \textit{i.e.}, race, age, and gender.
        %
    }
    \label{fig:fair_qualitative_2}
    \vspace{-2pt}
\end{figure*}

%% file: sections/4.2_API_Gender_Compare.tex
\begin{figure*}[htbp]
\centering
\subfigure[Mis-classified Male images]{
\includegraphics[width=8.32cm]{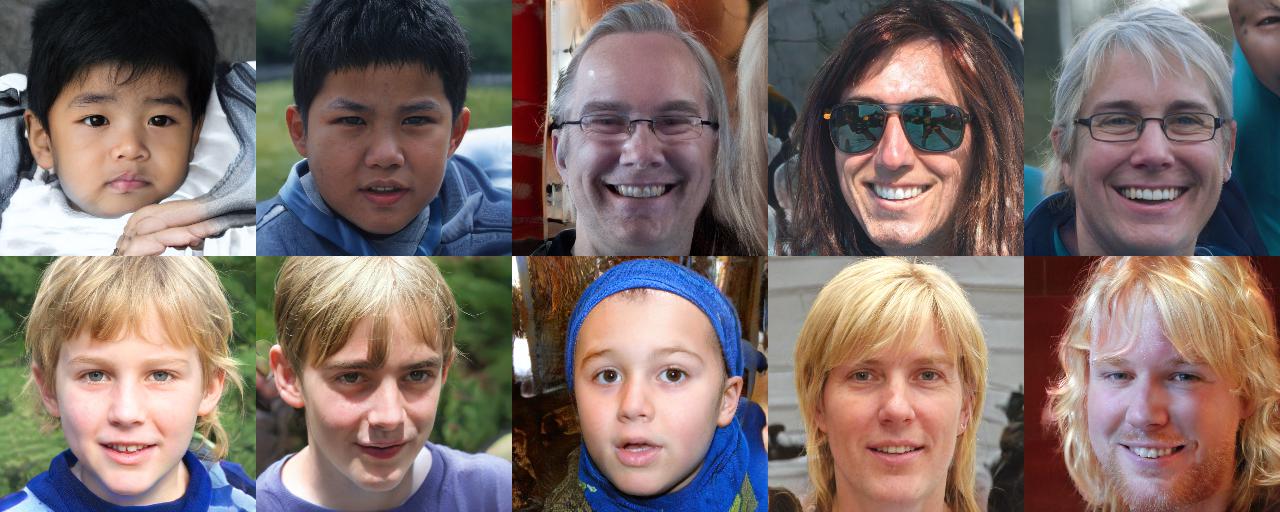}
}
\quad
\subfigure[Mis-classified Female images]{
\includegraphics[width=8.32cm]{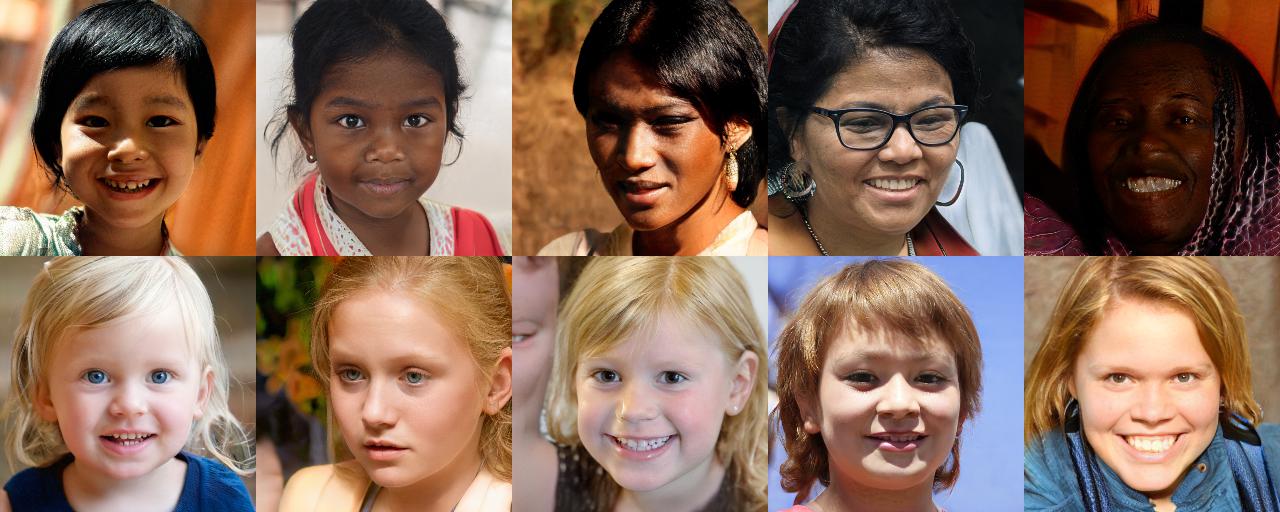}
}
\caption{Mis-classified images by APIs.}
\vspace{-6pt}
\label{fig:API}
\end{figure*}

\begin{figure*}[!htbp]
\centering
\vspace{-3pt}
\subfigure[Gender alteration]{
\includegraphics[width=5.30cm]{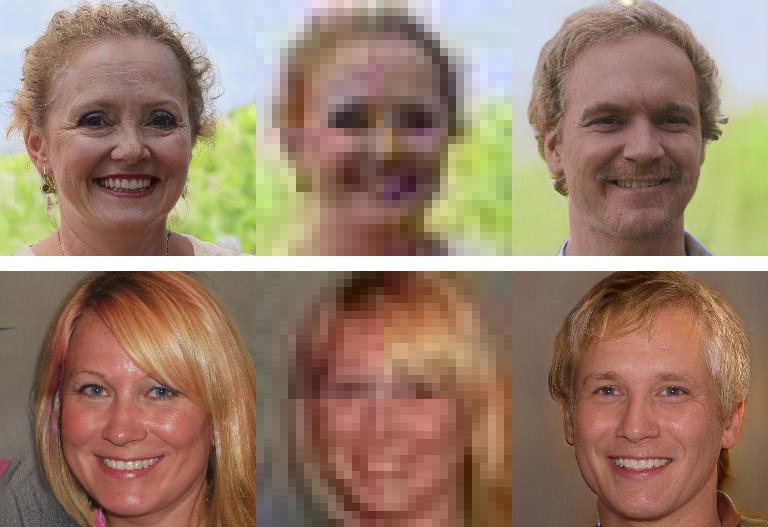}
}
\quad
\subfigure[Eyeglasses alteration]{
\includegraphics[width=5.30cm]{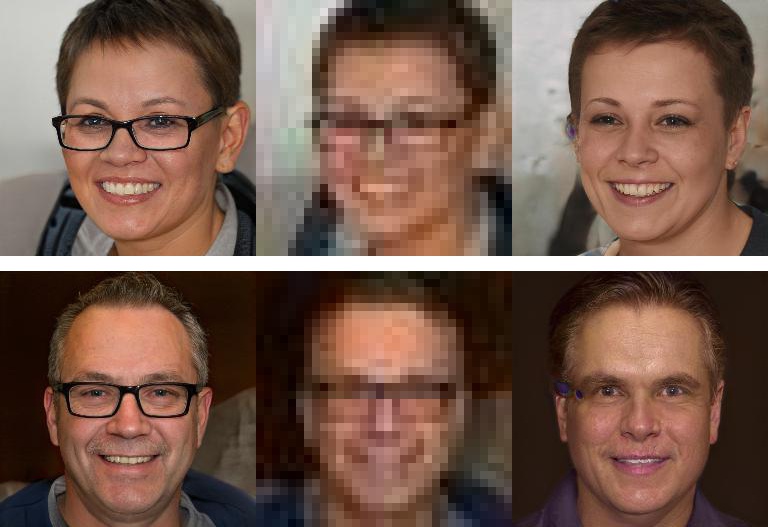}
}
\quad
\subfigure[Race alteration]{
\includegraphics[width=5.30cm]{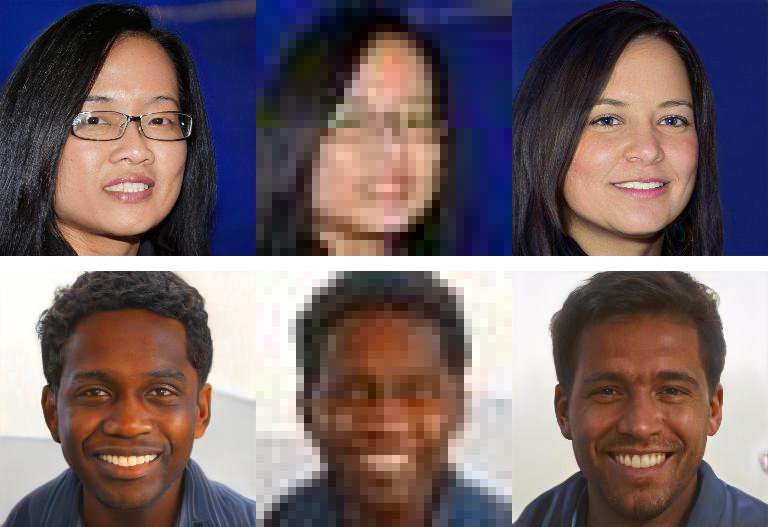}
}
\caption{Examples of attribute alteration by the super-resolution model. From left to right we showcase 1) the original image generated by our method; 2) the LR image down-sampled from the original image; 3) HR image output by PULSE given the LR image.}
\label{fig:PULSE}
\end{figure*}

\begin{table*}[!t]
  \centering
    \caption{Gender classification error rate (in percentage) in different attribute subgroups combined with gender.}
    \begin{tabular}{c|cc||cc|cc|ccc}
    \hline
      Sensitive Attributes
      & \multicolumn{2}{c||}{Gender} & \multicolumn{2}{c|}{Age} & \multicolumn{2}{c|}{Eyeglasses} & \multicolumn{3}{c}{Race}\\
     \hline
    Subgroups & Male & Female & Young & Old & With & Without & Black & Asian & White \\
     \hline
    Face++ Detect API& 0.81  & 4.19  & 2.02  & 1.76  & 0.58  & 3.40  & 3.50  & 0.64  & 2.28\\
    Azure Facial Recognition & 0.71  & 2.52  & 1.08  & 1.00  & 0.47  & 2.36  & 1.62  & 0.84 & 0.44\\
    \hline
    \end{tabular}%
    \vspace{-6pt}
  \label{tab:api_gender}%
\end{table*}%

%% file: sections/4.3_PULSE_compare.tex
\begin{table*}[ht]
  \centering
    \caption{Percentage of attribute alternation by the super-resolution model.}
    \begin{tabular}{c*{9}{|p{1.0cm}<{\centering}}}
    \hline
    Attributes & \multicolumn{2}{c|}{Gender} & \multicolumn{2}{c|}{Age} & \multicolumn{2}{c|}{Glasses} & \multicolumn{3}{c}{Race} \\
    \hline
    Groups & Male & Female & Young  & Old & With & Without & Black & Asian & White\\
    \hline
    Value alternation rate & 3.4  & \textbf{7.8} & 3.3 & \textbf{29.5} & \textbf{89.8} & 0.3  & \textbf{76.4} & 34.3 & 0.5 \\
    \hline
    \end{tabular}%
    \vspace{-2pt}
  \label{tab:PULSE}%
\end{table*}%

%% file: sections/4.4_Analysis.tex
\begin{table}[!ht]
  \centering
    \vspace{-4pt}
    \caption{Ablation study of our method.}
    \vspace{-2pt}
    \begin{tabular}{c*{3}{|p{1.5cm}<{\centering}}}
    \hline
    \multirow{2}{*}{Attributes} & age-eyeglasses & gender-eyeglasses & black-gender \\
    \hline
    Ours w/o Edit & 0.0497  & 0.0139  & 0.0292\\
    Ours w/o Filter & 0.0200  & 0.0053  & 0.0404\\
    Ours w/o GMM & 0.0122  & 0.0031  & 0.0267 \\
    Ours & \textbf{0.0079} & \textbf{0.0013} & \textbf{0.0102} \\
    \hline
    \end{tabular}%
    \vspace{-4pt}
  \label{tab:fair_ablation}%
\end{table}%

%% file: sections/5_Conclusion.tex
\section{Conclusion}
%
In this work, following the empirical study on the bias in the training set and the trained generative model, we develop a baseline method to eliminate the bias within a well-trained GAN model by discreetly altering the sampling strategy yet retaining the model weights. Our method has shown great potential in turning numerous publicly available GAN models unbiased and helping detect the unfairness in other AI systems. We hope our work can raise the awareness of the fairness in generative models, and our method will be a starting point to diminish and eventually eliminate all the potential biases in generative models.
%
%
%
%
%
%

%% file: sections_supp/dist_supp.tex
\section{Data Distribution Comparisons}

\setlength{\parindent}{2em}
\par 
In this section, we show more comparisons of the data distributions of different datasets w.r.t the target attributes. Specifically, we compare the data distribution of the real training dataset FFHQ \cite{Karras2019Style}, the dataset directly sampled with GAN \cite{Karras2019stylegan2}, and the dataset sampled with our method.

\par
In Figure \ref{fig:dist_plots_1} and \ref{fig:dist_plots_2} we show results for the other four tasks in addition to the task we show in the paper. We observe that our method is able to consistently remove the bias in the GAN model without retraining across multiple tasks.


\input{sections_supp/dist_plots}

\newpage


\newpage

%% file: sections_supp/dist_plots.tex
\vspace*{-1.5em} 

\begin{figure}[H]
\hsize=\textwidth
\centering
\subfigure[Age - Gender]{
\includegraphics[width=8.32cm]{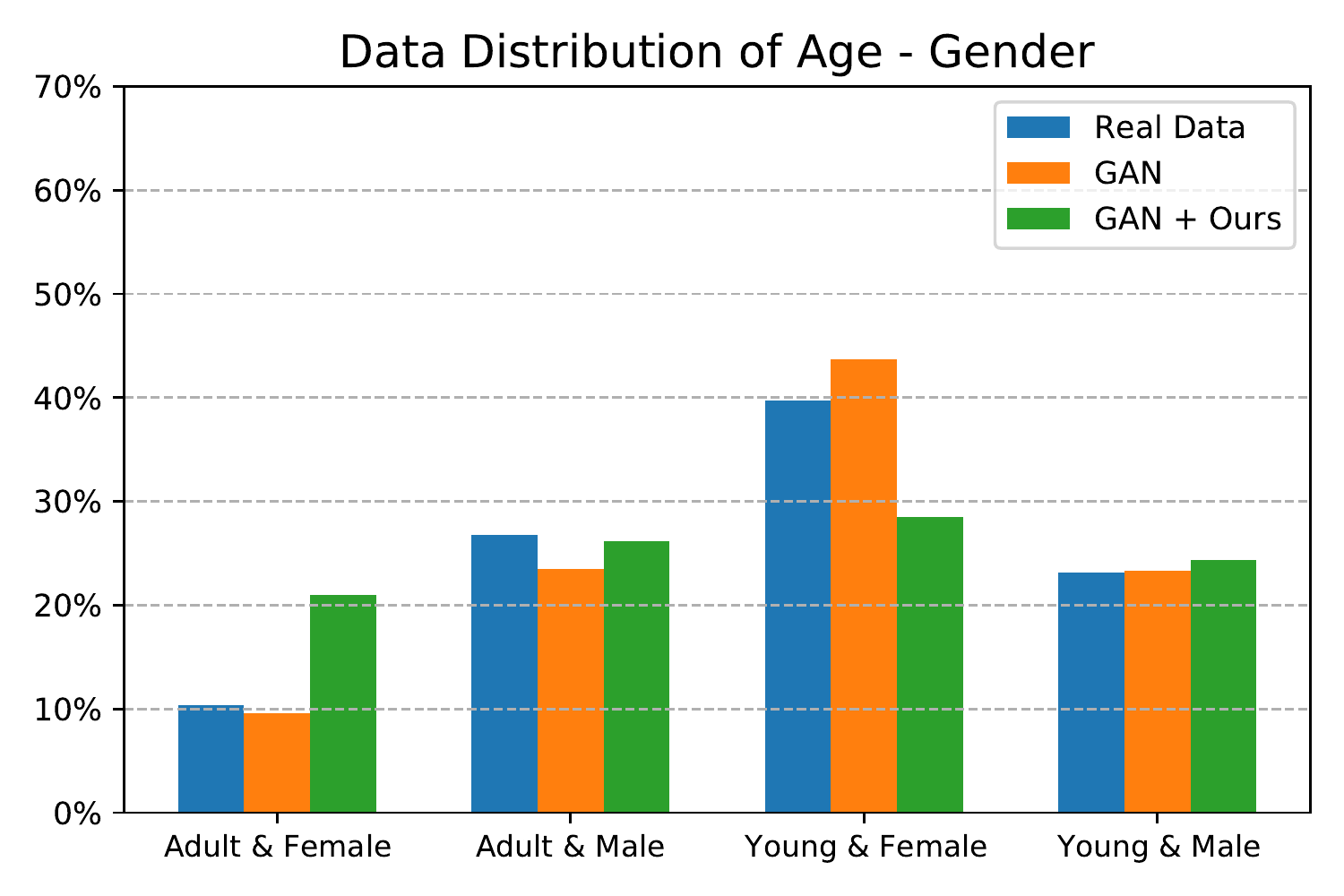}
}
\quad
\subfigure[Gender - Eyeglasses]{
\includegraphics[width=8.32cm]{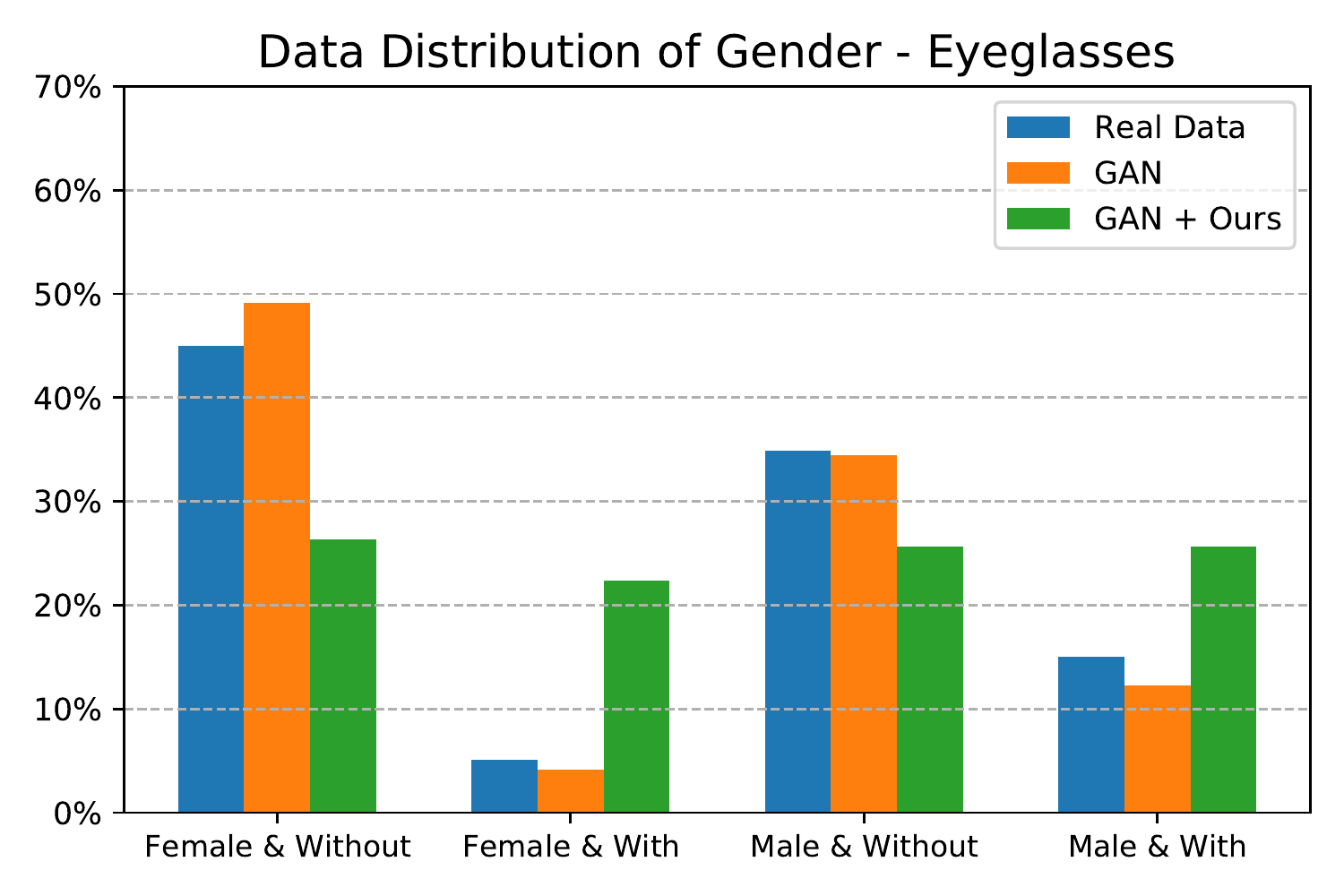}
}
\quad
\subfigure[Gender - Black\_hair]{
\includegraphics[width=8.32cm]{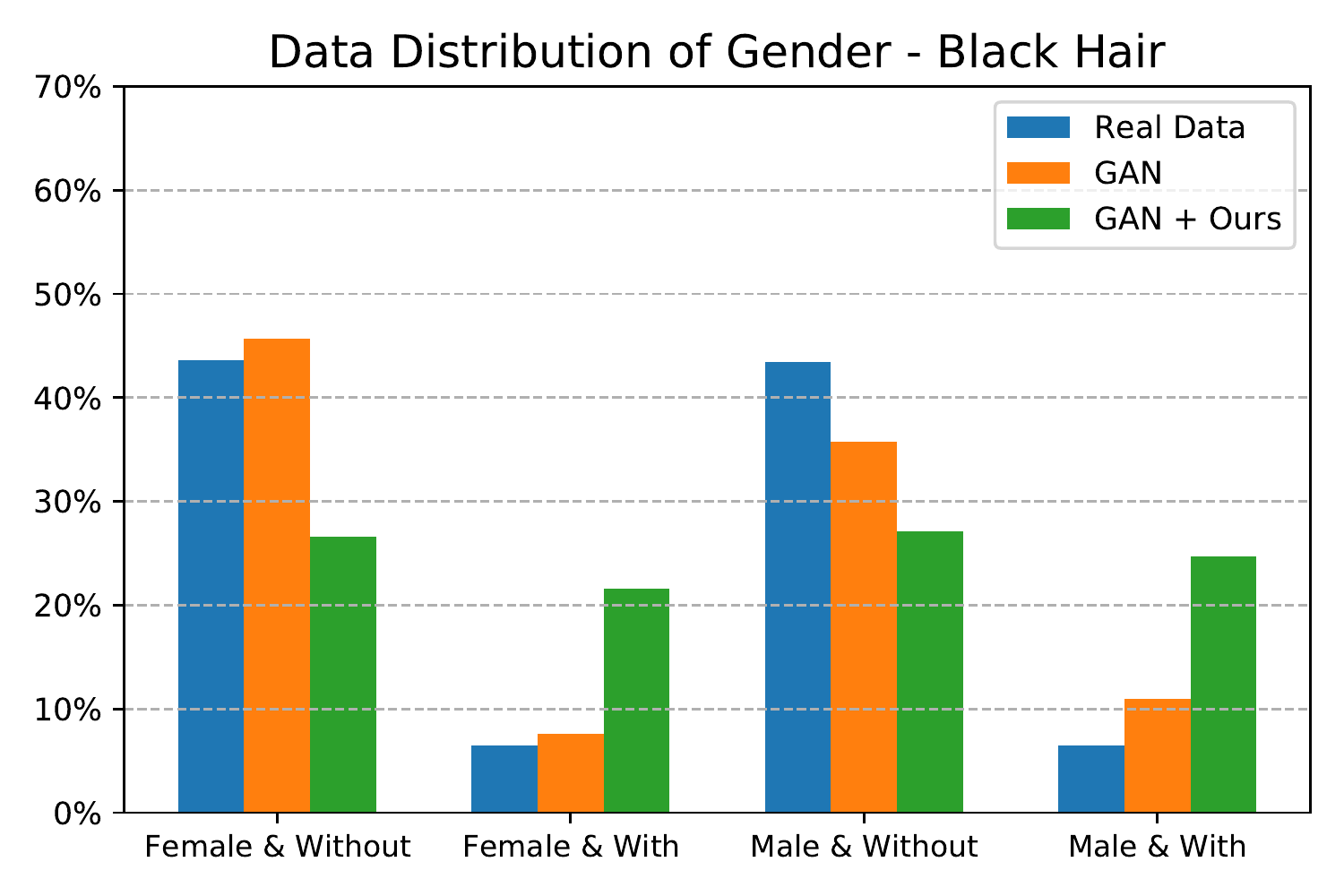}
}
\quad
\subfigure[Age - Smiling]{
\includegraphics[width=8.32cm]{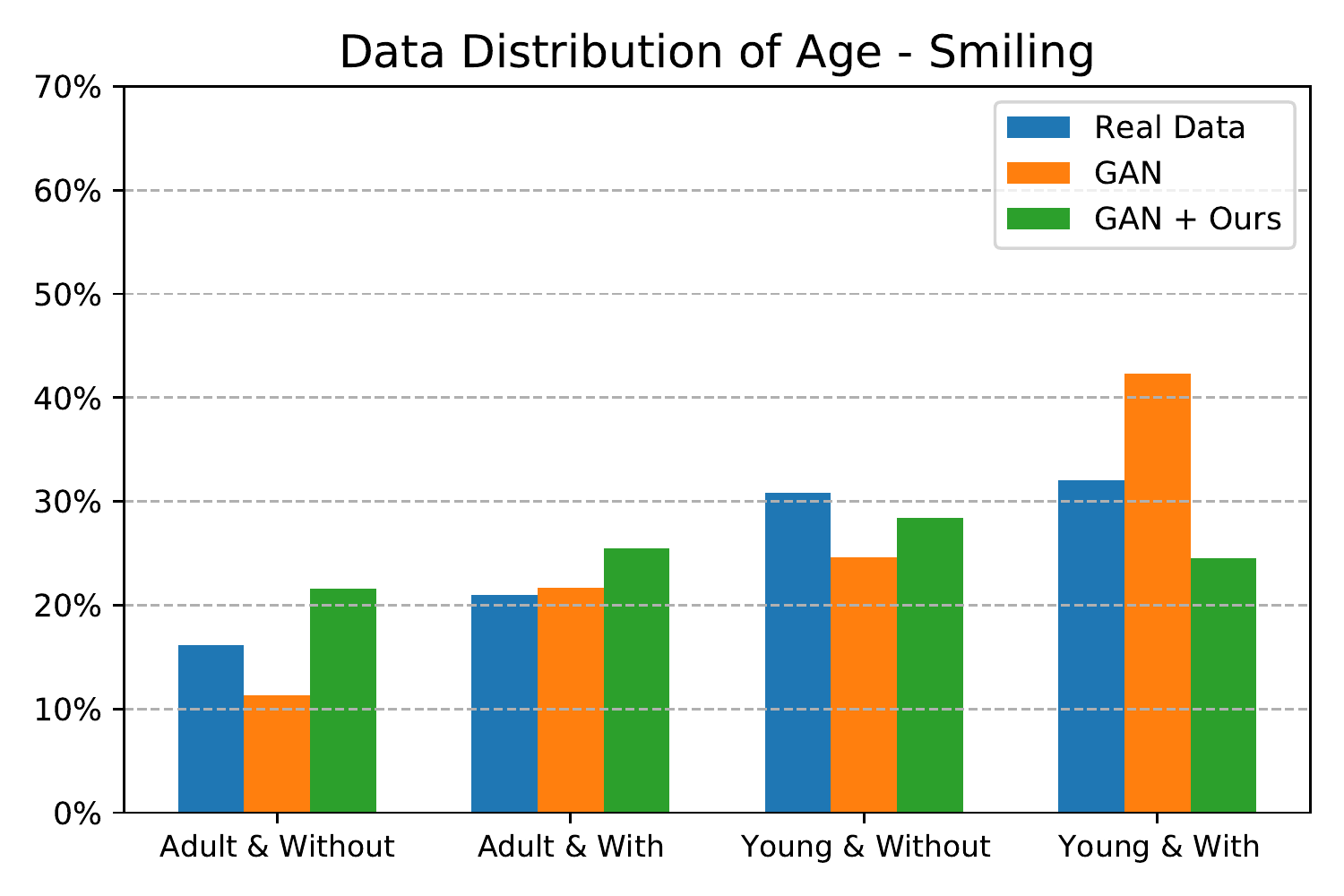}
}
\caption{Comparisons of data distributions in different datasets.}
\label{fig:dist_plots_1}
\end{figure}




%% file: sections_supp/hyper_parameter.tex
\vspace{-1.5em}
\section{Hyper-parameter Study} \label{exp:analysis}
In this section, we analyze the impact of hyper-parameter choices in FairGen on the fairness score. In particular, we analyze 1) the magnitude of InterfaceGAN manipulation $\left|\alpha\right|$; 2) the size of edited latent code set $N_\text{edit}$; 3) the number of components used in GMM models $k$. 

\begin{figure}[H]
\hsize=\textwidth
\subfigure[Magnitude of latent space shift $\left|\alpha\right|$]{
\includegraphics[width=5.25cm]{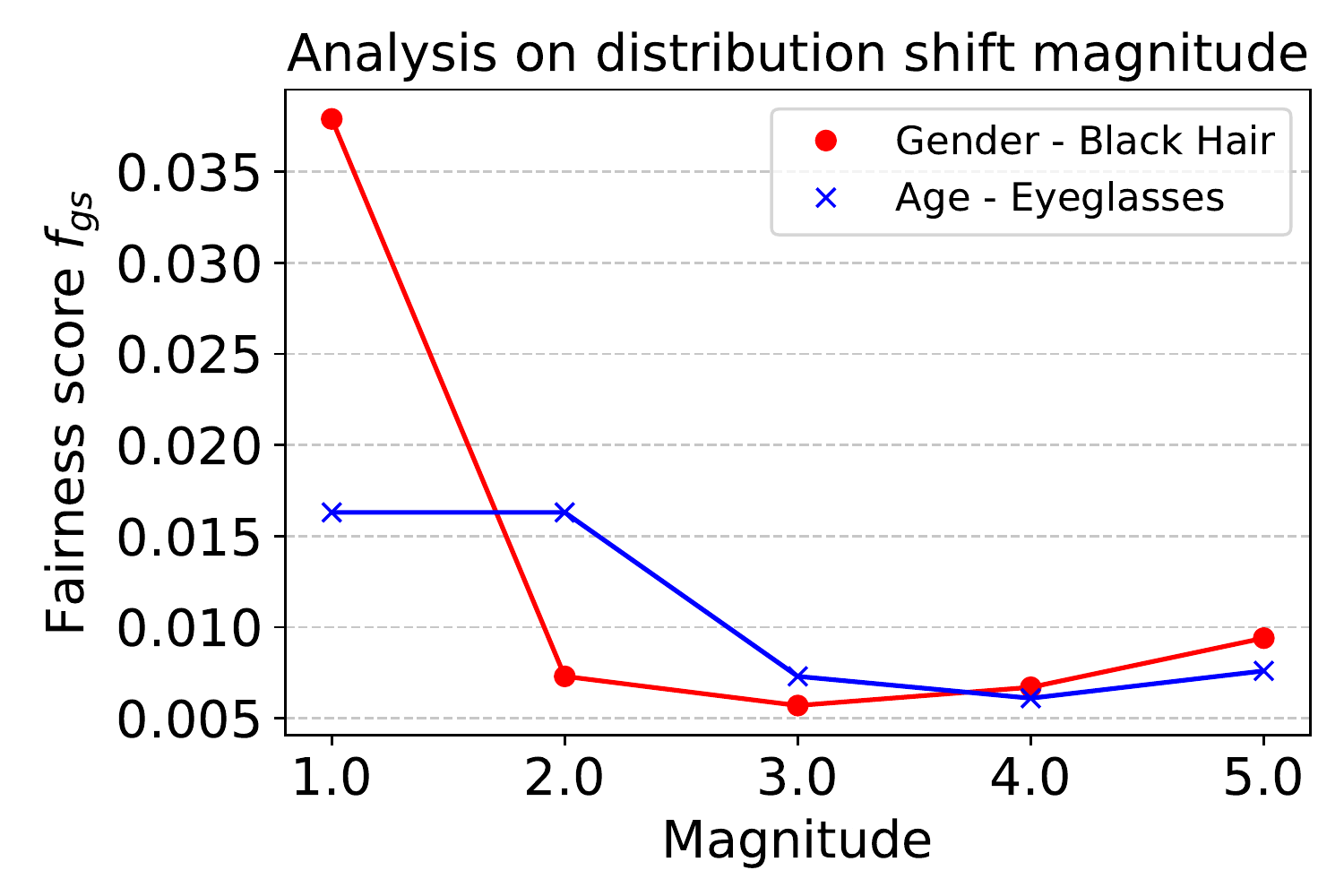}
}
\quad
\subfigure[Size of edited latent code set $N_\text{edit}$]{
\includegraphics[width=5.25cm]{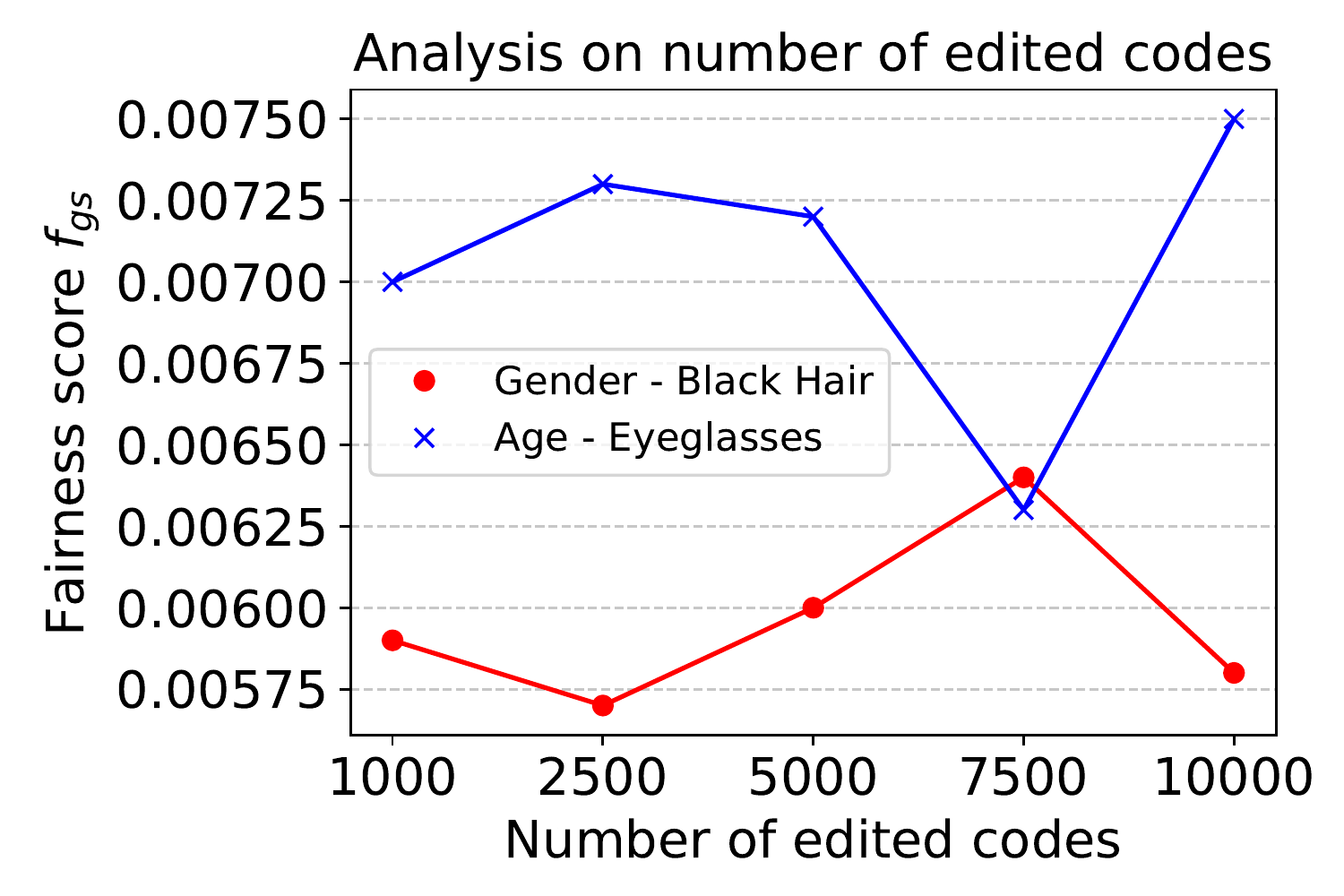}
}
\quad
\subfigure[Number of GMM components $k$]{
\includegraphics[width=5.25cm]{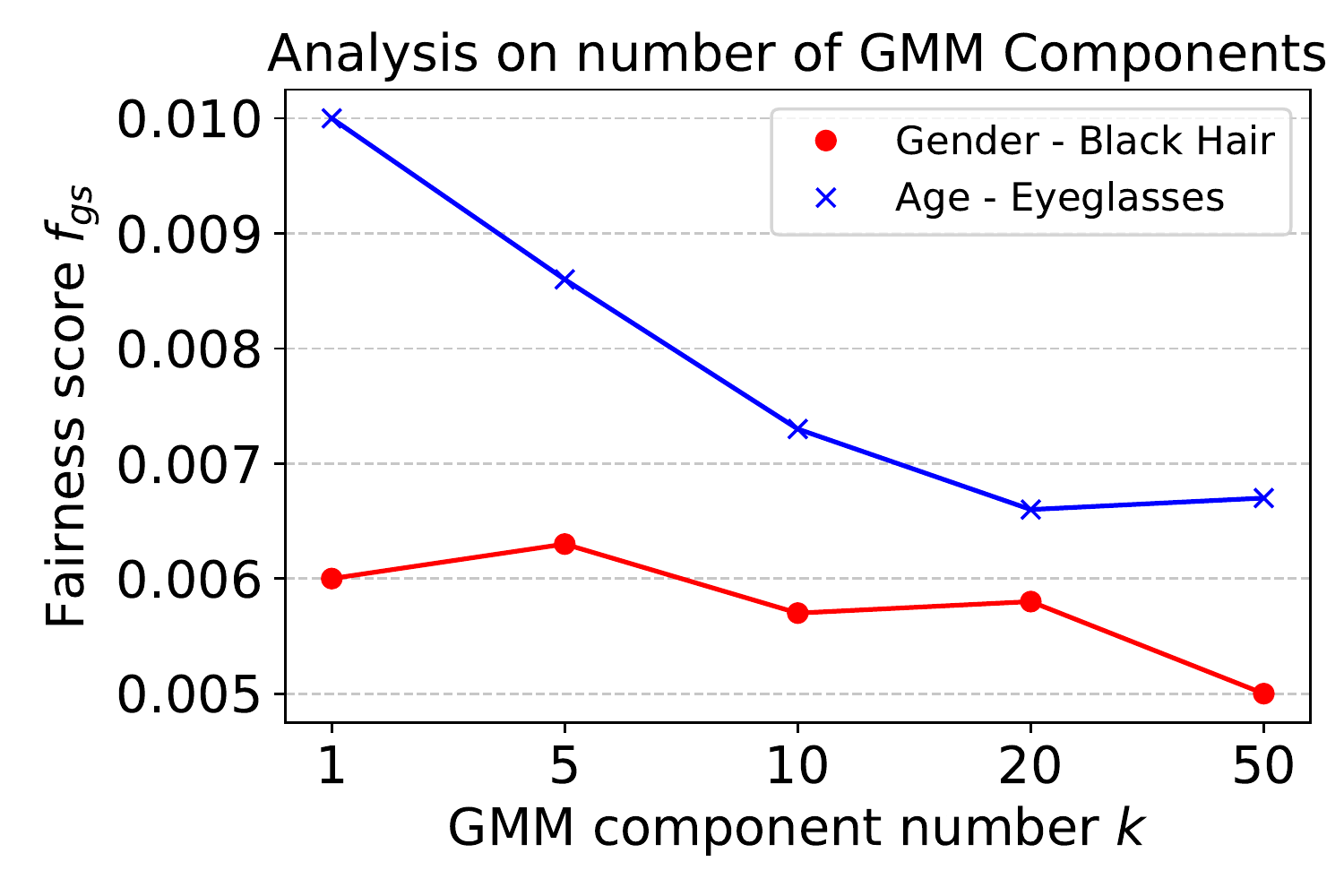}
}
\caption{Results of the analysis of each of the hyper-parameters in our framework.}
\label{fig:Analysis}
\end{figure}

We plot the results in Table~\ref{fig:Analysis}. We observe that we need a large enough $\left|\alpha\right|$ to make sure the semantics of the shifted latent codes are correct. Secondly, the size of edited codes $N_\text{edit}$ has a relatively small impact on the fairness score compared to other parameters. Finally, The more components we have in GMM, the better result we will normally obtain as more components provide a more accurate approximation of code distribution.

\vspace{-1.5em}

%% file: sections_supp/dist_plots_2.tex
\vspace*{2em} 

\begin{figure}[H]
\hsize=\textwidth
\centering
\subfigure[Age - Eyeglasses]{
\includegraphics[width=8.32cm]{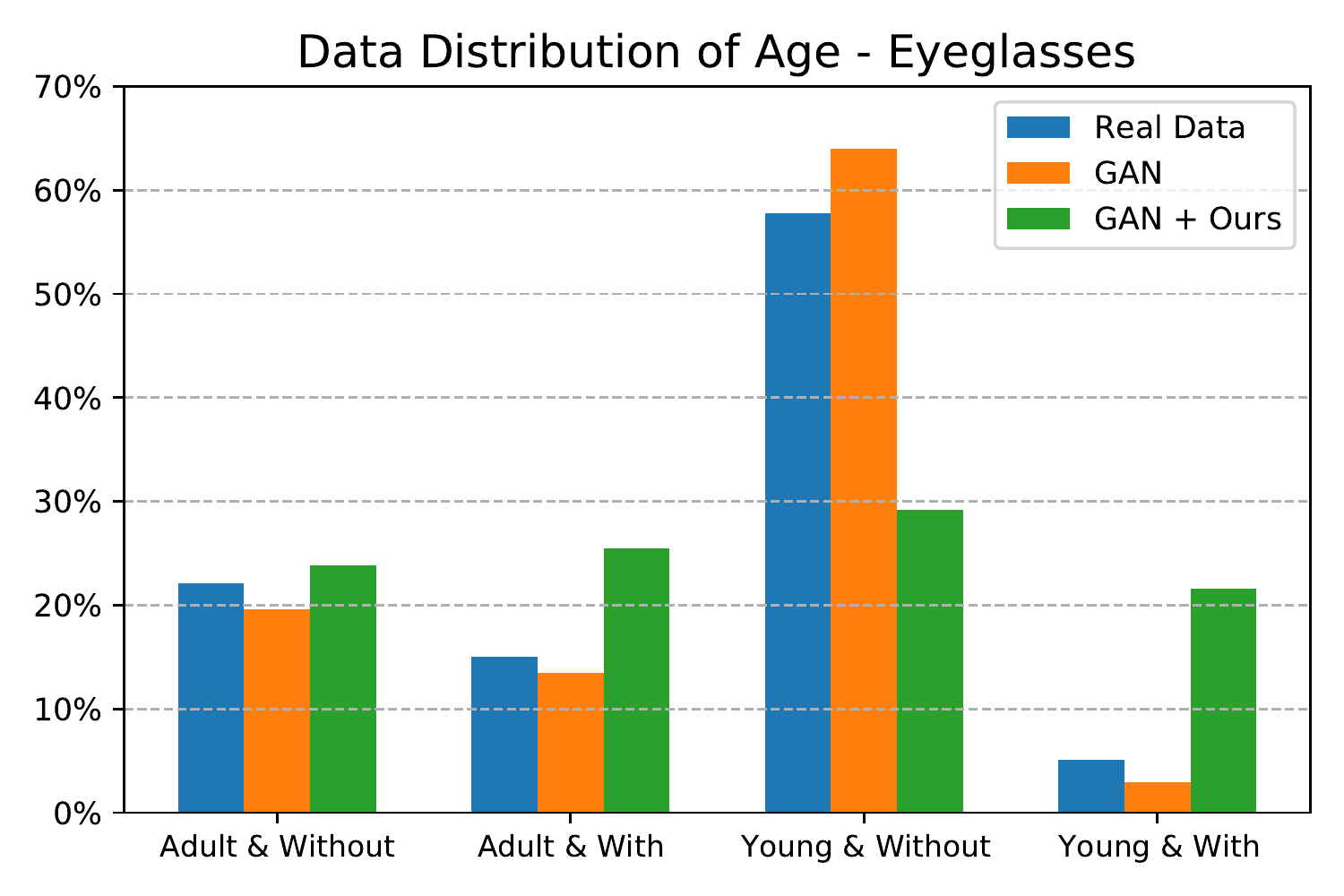}
}
\subfigure[Black - Age]{
\includegraphics[width=8.32cm]{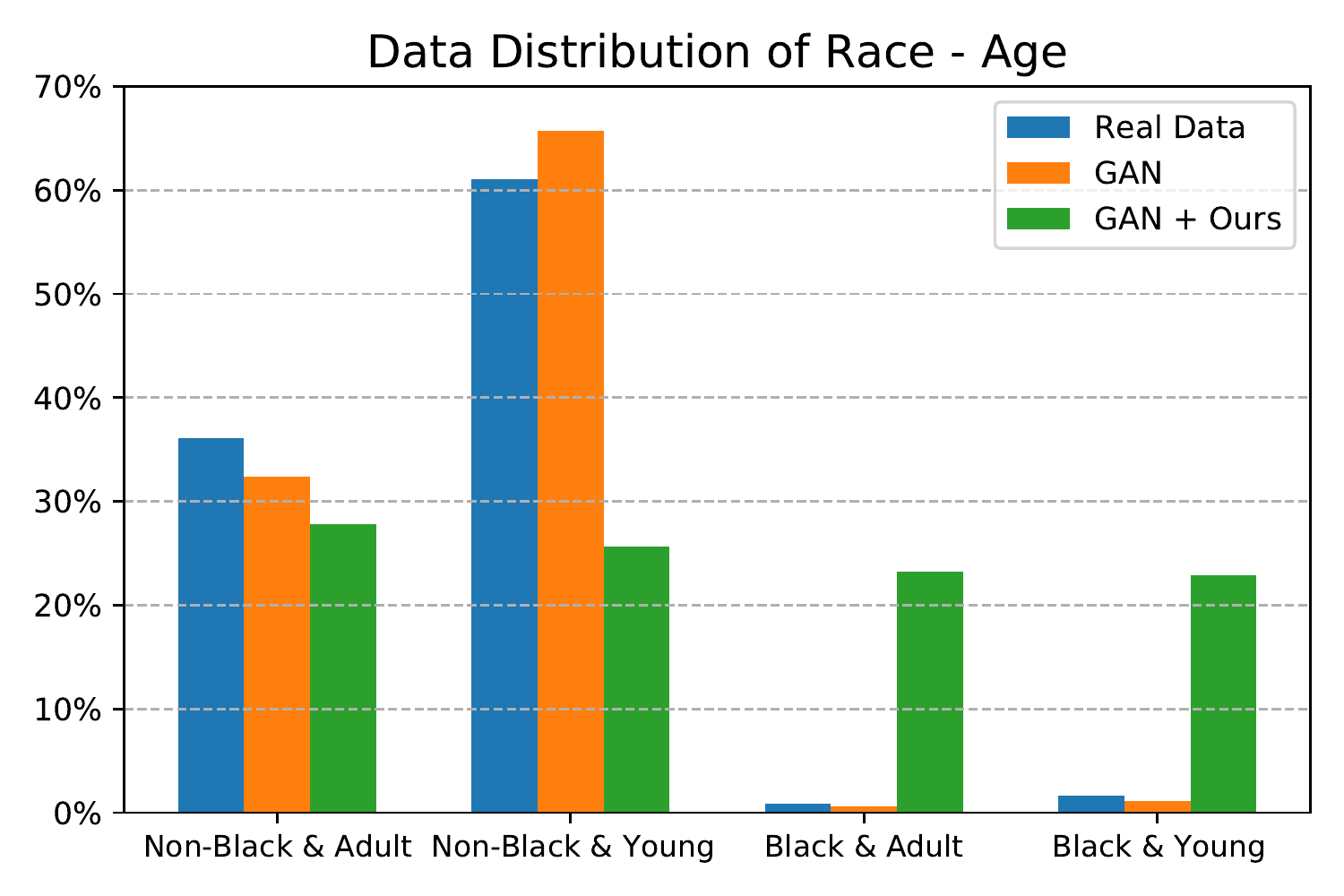}
}
\subfigure[Asian - Age]{
\includegraphics[width=8.32cm]{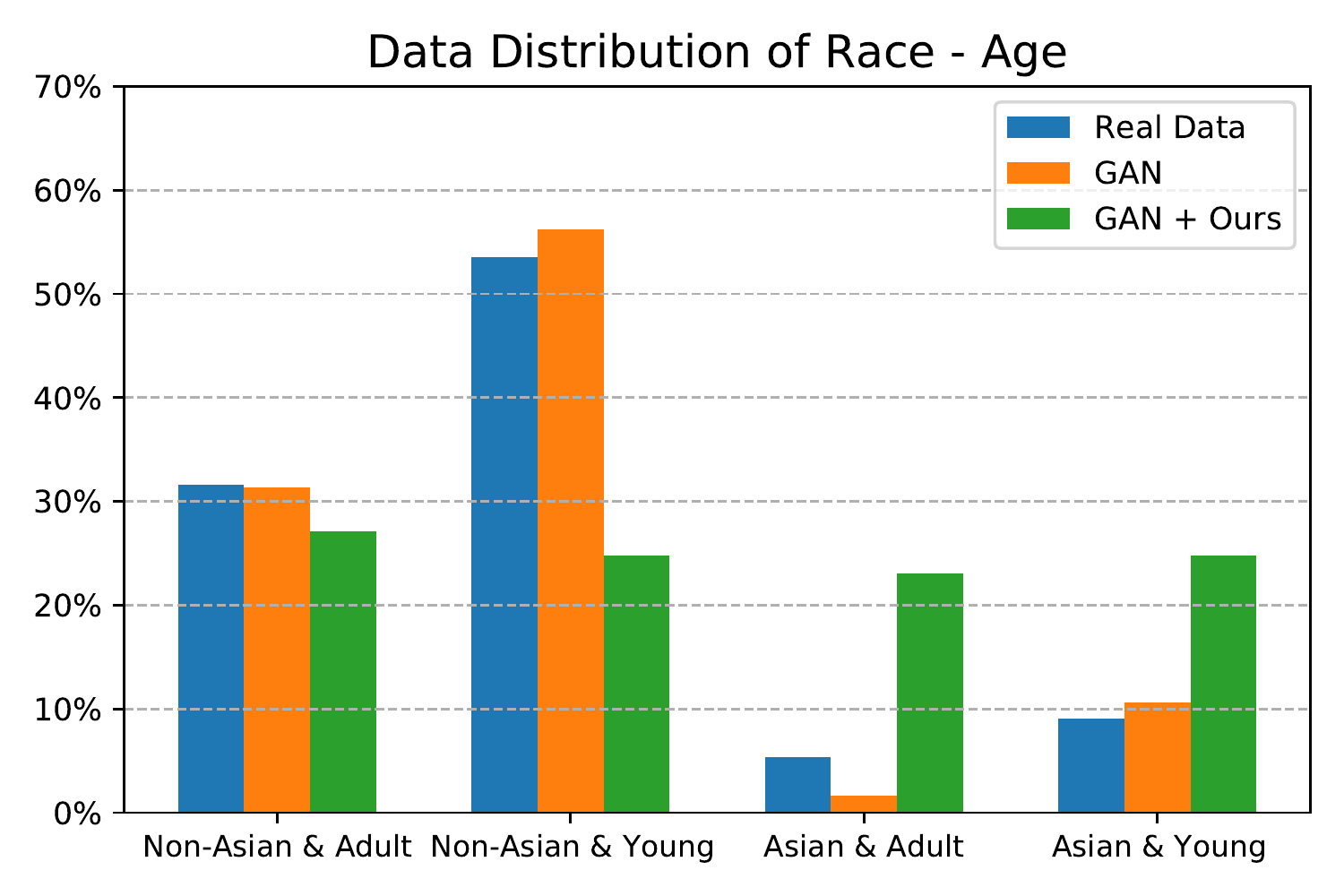}
}
\subfigure[Asian - Gender]{
\includegraphics[width=8.32cm]{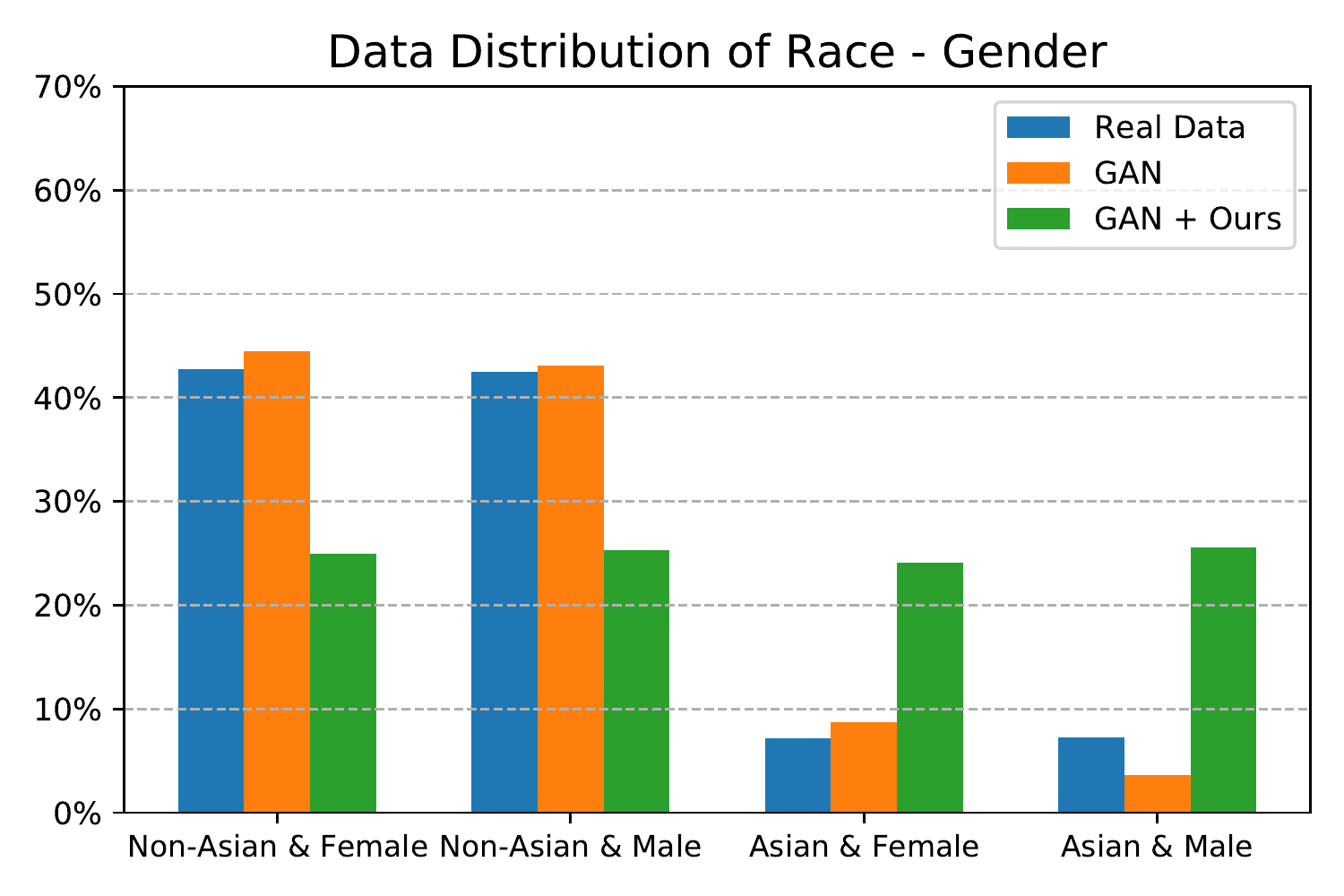}
}
\caption{Comparisons of data distributions in different datasets (continued).}
\label{fig:dist_plots_2}
\end{figure}

\vspace*{3em} 

%% file: sections_supp/api_details.tex
\section{Commercial API details}

\setlength{\parindent}{2em}
\par 
We detail how we test the commercial APIs in our experiments.

For the evaluation of facial gender classification, we first obtain access to two commercial APIs: MEGVII's Face++ Detect API (\url{https://www.faceplusplus.com/face-detection/}) and Microsoft's Azure Facial Recognition service (\url{https://azure.microsoft.com/en-us/services/cognitive-services/face/}). Then, we utilize FairGen to generate facial images in different subgroups. We use the gender attribute value of each subgroup as the ground-truth gender label for the images in that subgroup. After that, we use both APIs to detect and analyze the face in each image, and then compare the predicted gender attribute with the ground-truth label to obtain the accuracies in the paper. Note that the APIs might fail to detect the face in a few images (around 0.5\%), which are ignored during gender classification accuracy computation.

It is worth noting that here we are not claiming the defects or bugs in the commercial products. We just would like to raise the awareness of the potential bias in the existing applications through this small and humble academic work. 

\vspace{2em}

%% file: sections_supp/sample.tex
\section{Conditional Generated Samples}

\setlength{\parindent}{2em}
\par 
In this section, we show more examples of the attribute subgroup images generated by our method for each task.

\input{sections_supp/sample_plots_age_gender}

%% file: sections_supp/sample_plots_age_gender.tex
\begin{figure}[H]
\hsize=\textwidth
\centering
\subfigure[Young Male]{
\includegraphics[width=8.32cm]{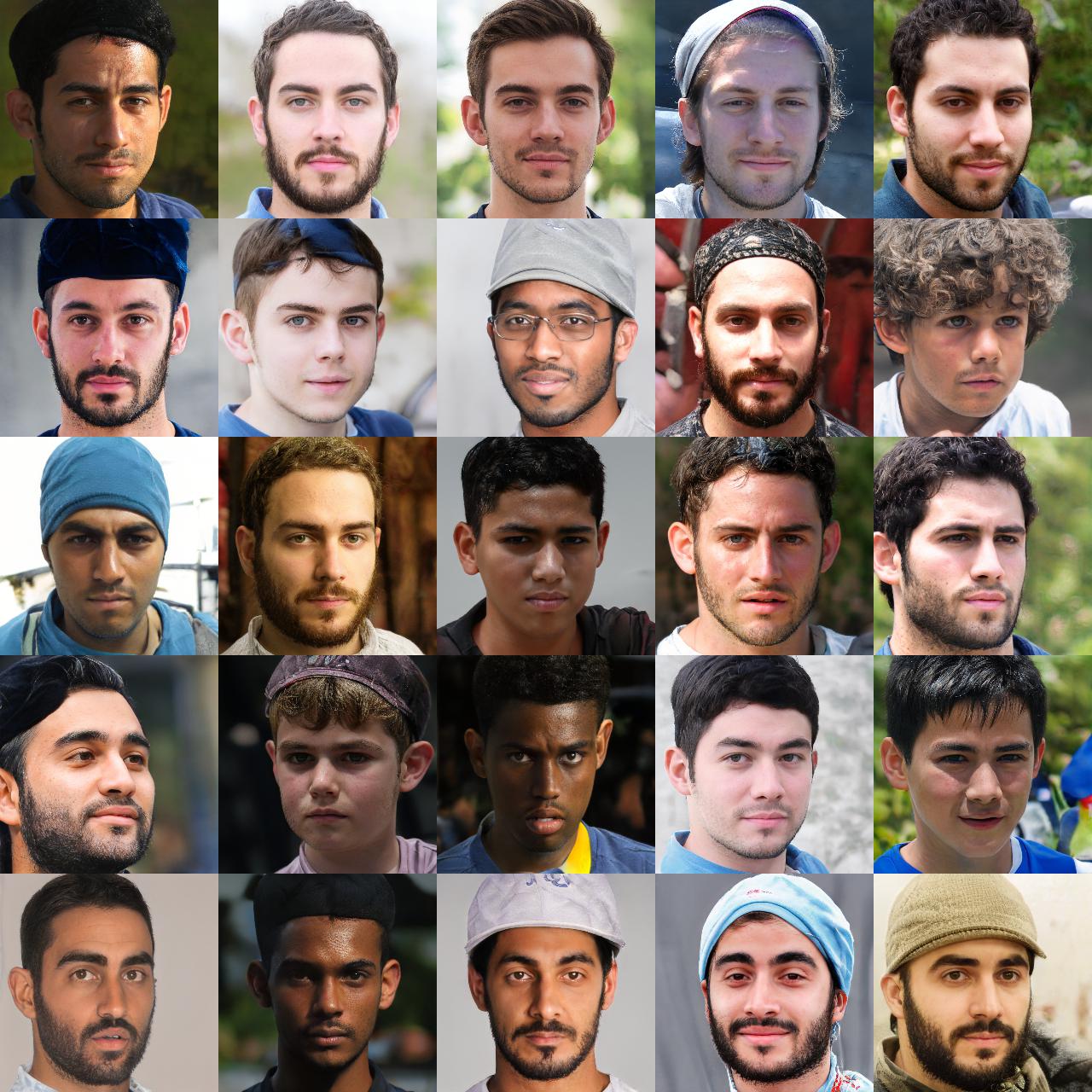}
}
\quad
\subfigure[Young Female]{
\includegraphics[width=8.32cm]{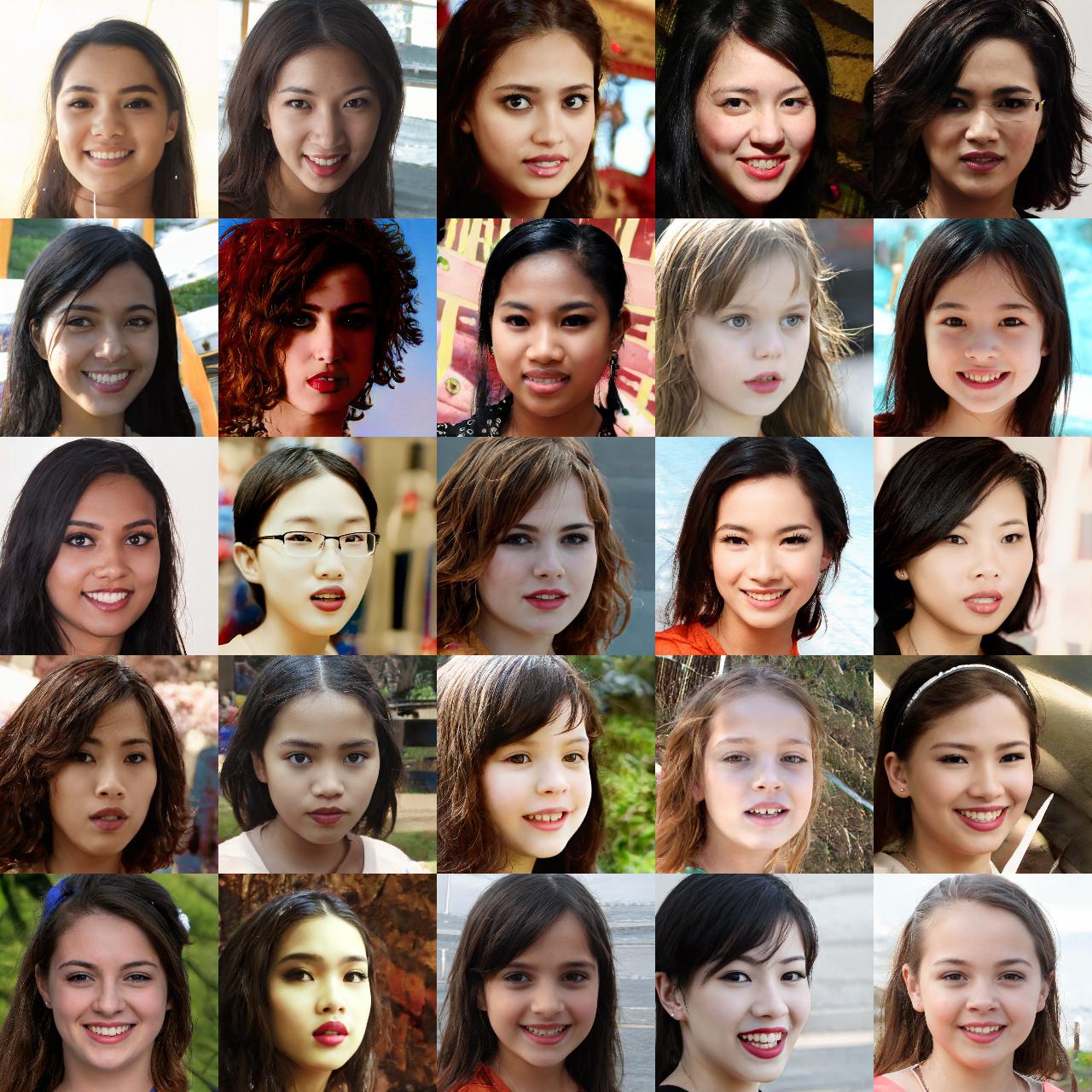}
}
\quad
\subfigure[Old Male]{
\includegraphics[width=8.32cm]{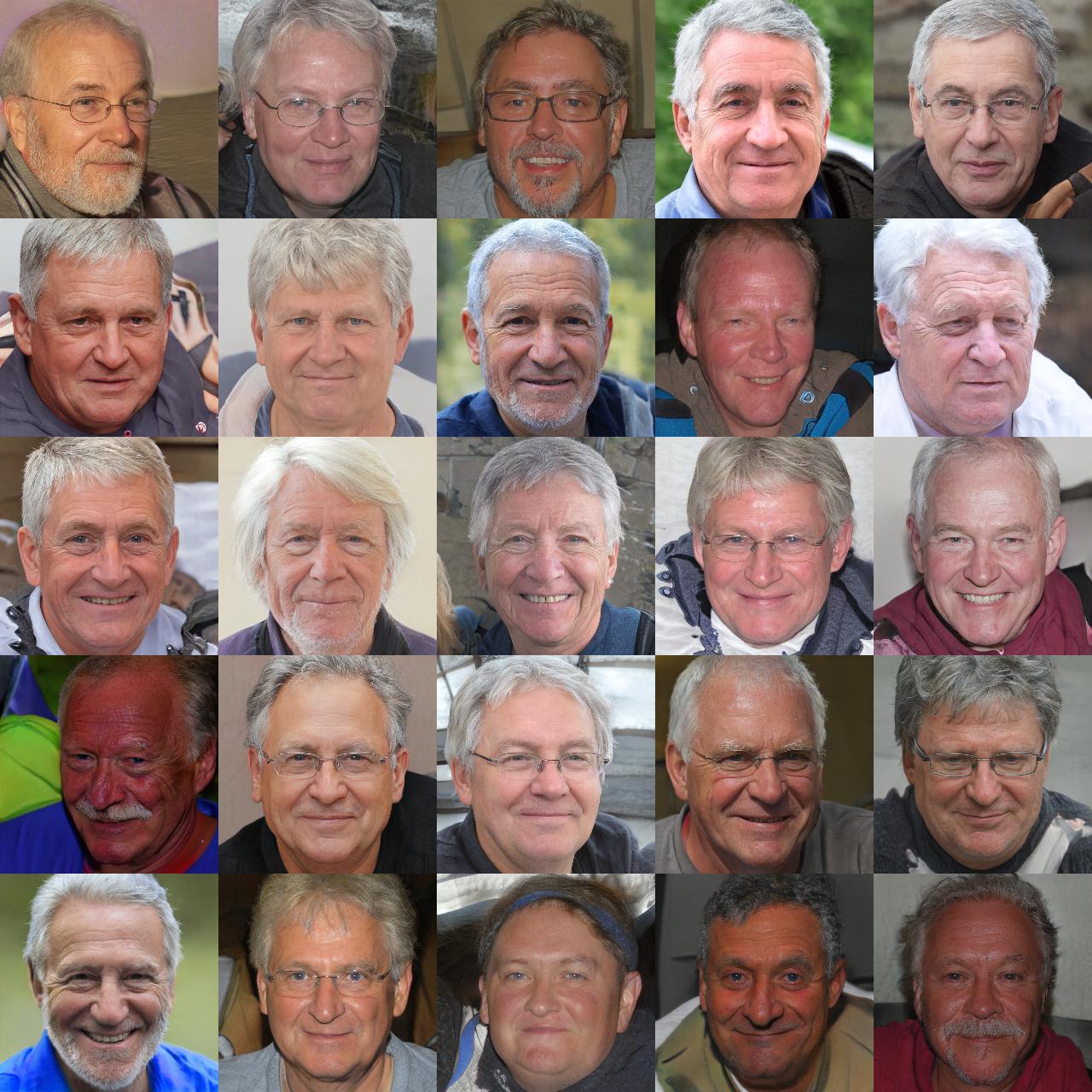}
}
\quad
\subfigure[Old Female]{
\includegraphics[width=8.32cm]{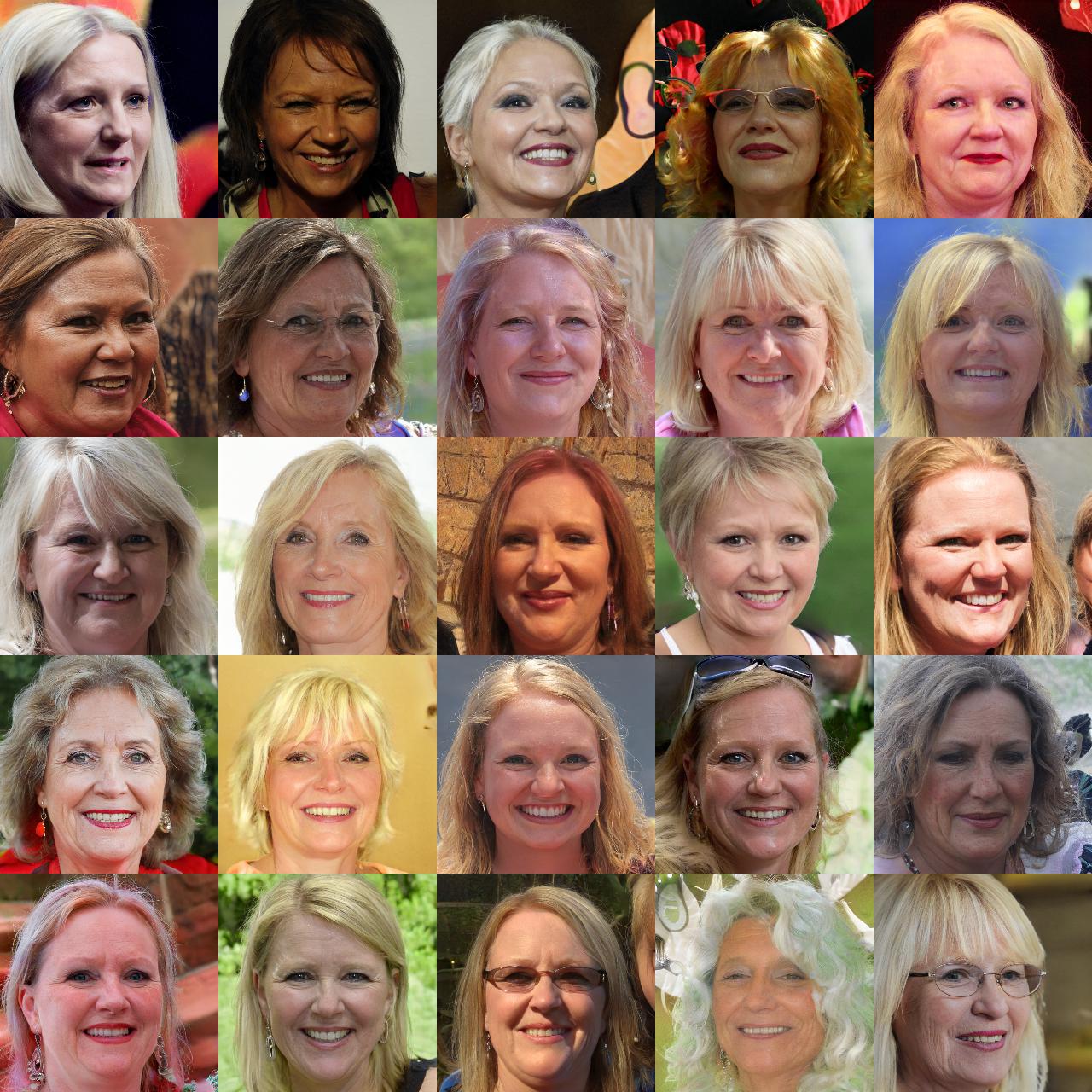}
}
\caption{Qualitative results for fair image generation in GANs with Age and Gender.}
\end{figure}

\vspace*{\fill} 

%% file: sections_supp/sample_plots_age_eyeglasses.tex
\vspace*{3.2em} 

\begin{figure}[H]
\hsize=\textwidth
\centering
\subfigure[Young with Eyeglasses]{
\includegraphics[width=8.32cm]{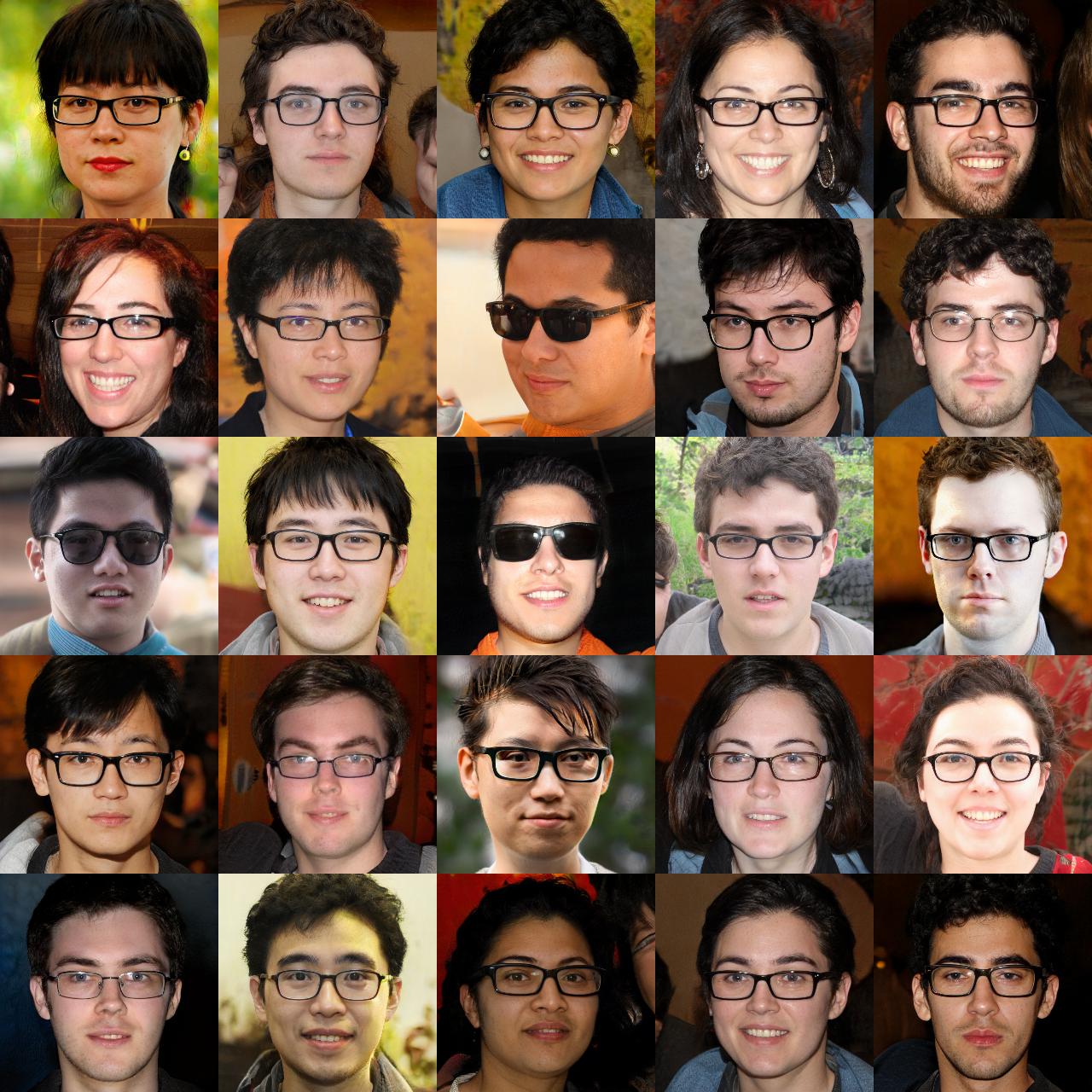}
}
\quad
\subfigure[Young without Eyeglasses]{
\includegraphics[width=8.32cm]{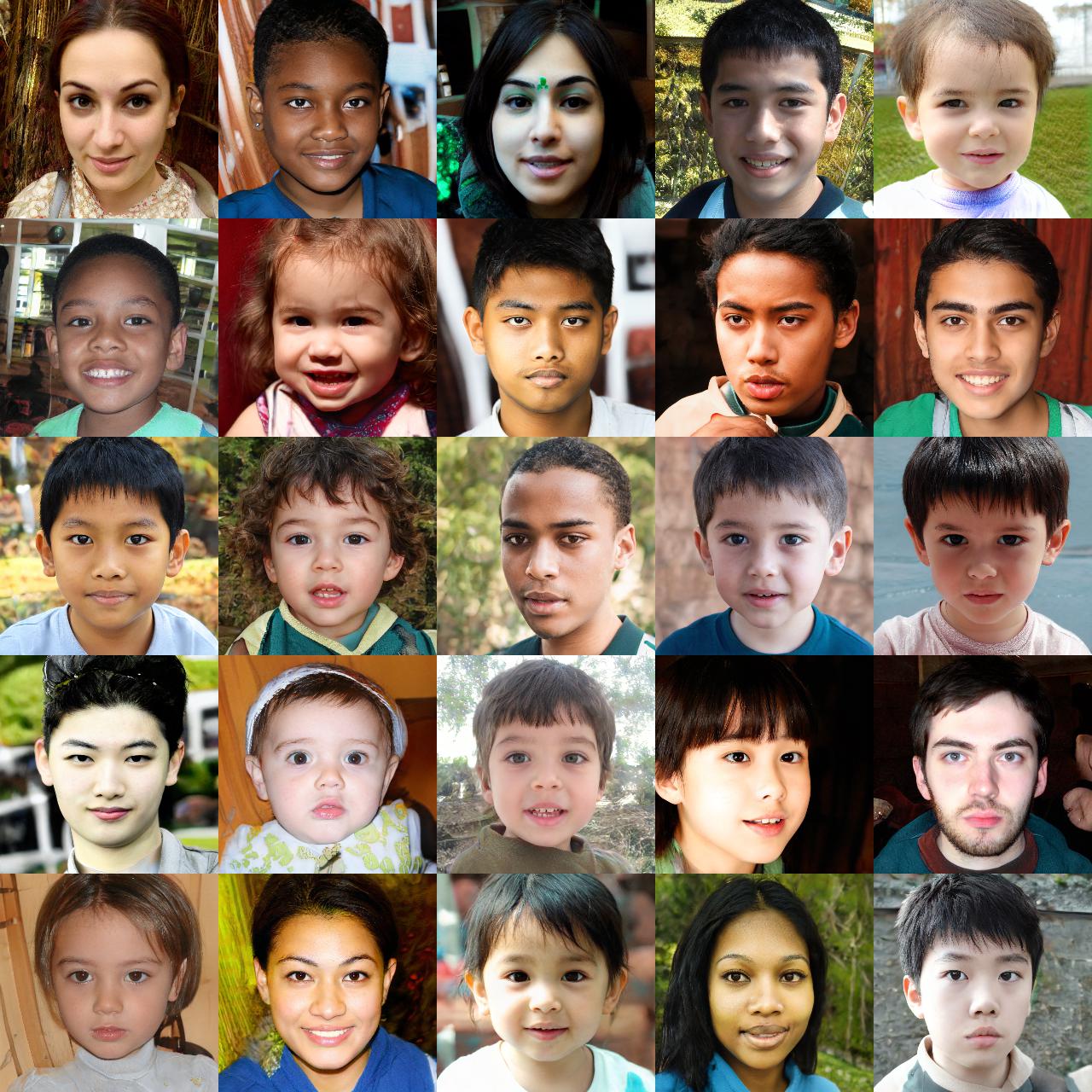}
}
\quad
\subfigure[Old with Eyeglasses]{
\includegraphics[width=8.32cm]{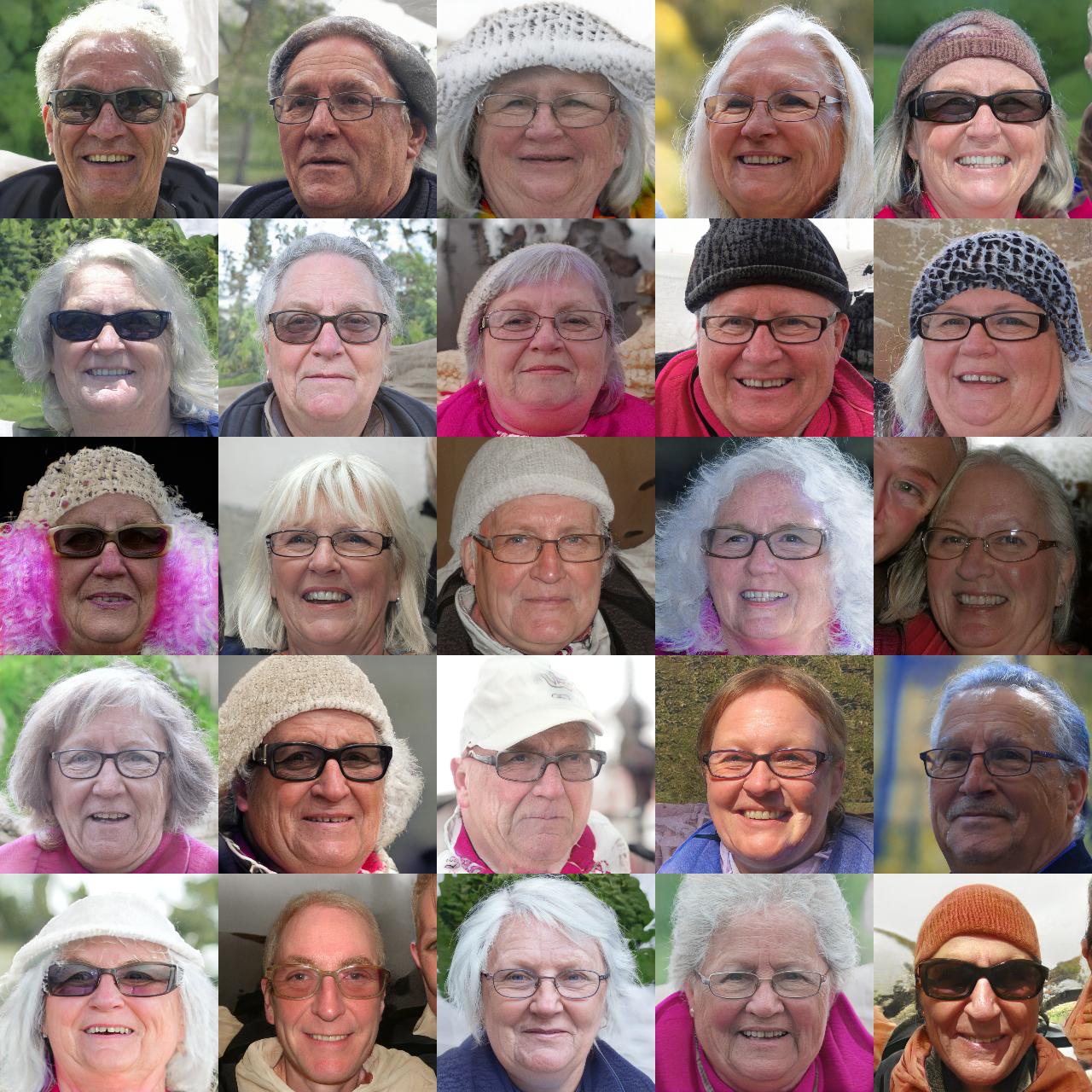}
}
\quad
\subfigure[Old without Eyeglasses]{
\includegraphics[width=8.32cm]{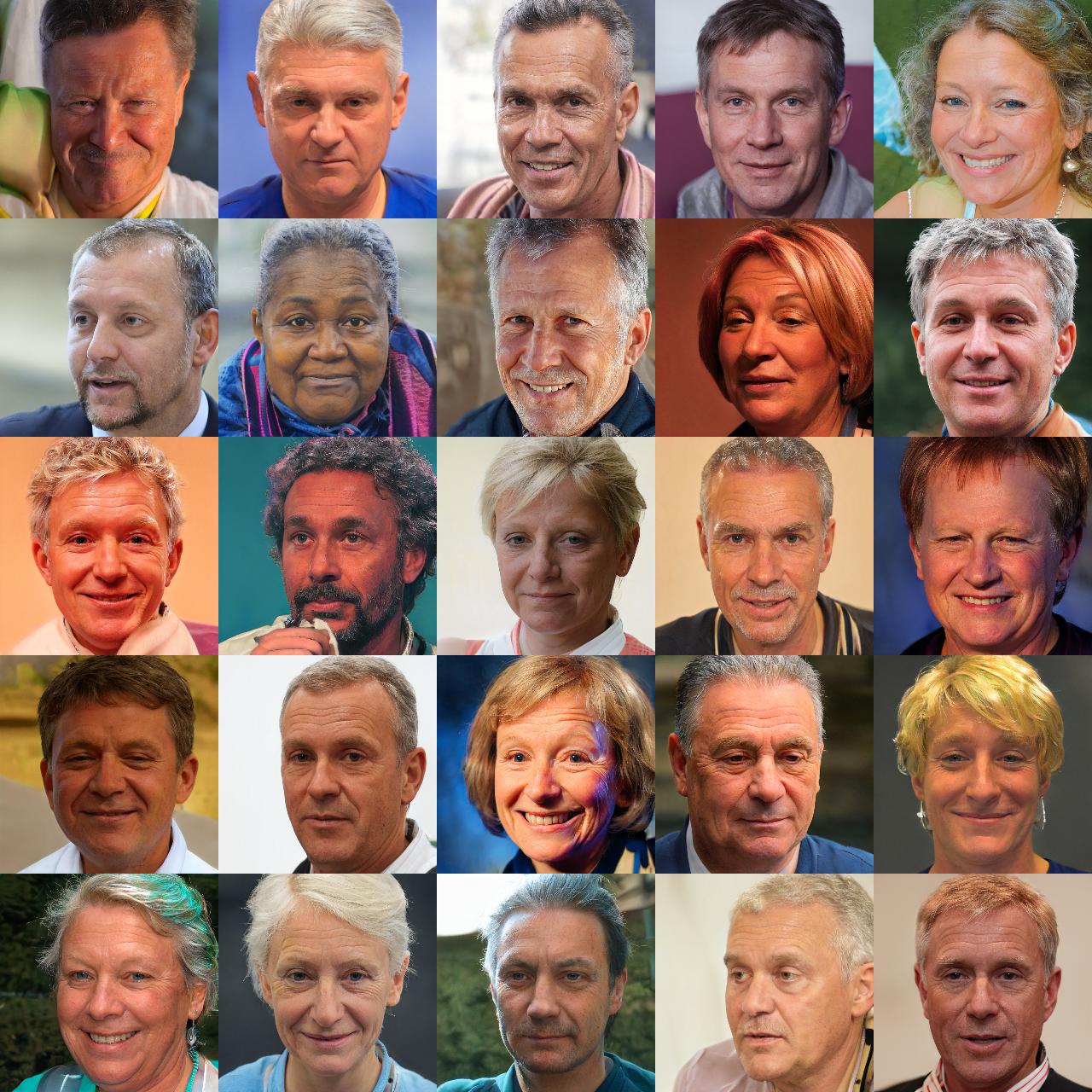}
}
\caption{Qualitative results for fair image generation in GANs with Age and Eyeglasses.}
\end{figure}

\vspace*{\fill}

%% file: sections_supp/sample_plots_gender_eyeglasses.tex
\vspace*{3.2em} 

\begin{figure}[H]
\hsize=\textwidth
\centering
\subfigure[Male with Eyeglasses]{
\includegraphics[width=8.32cm]{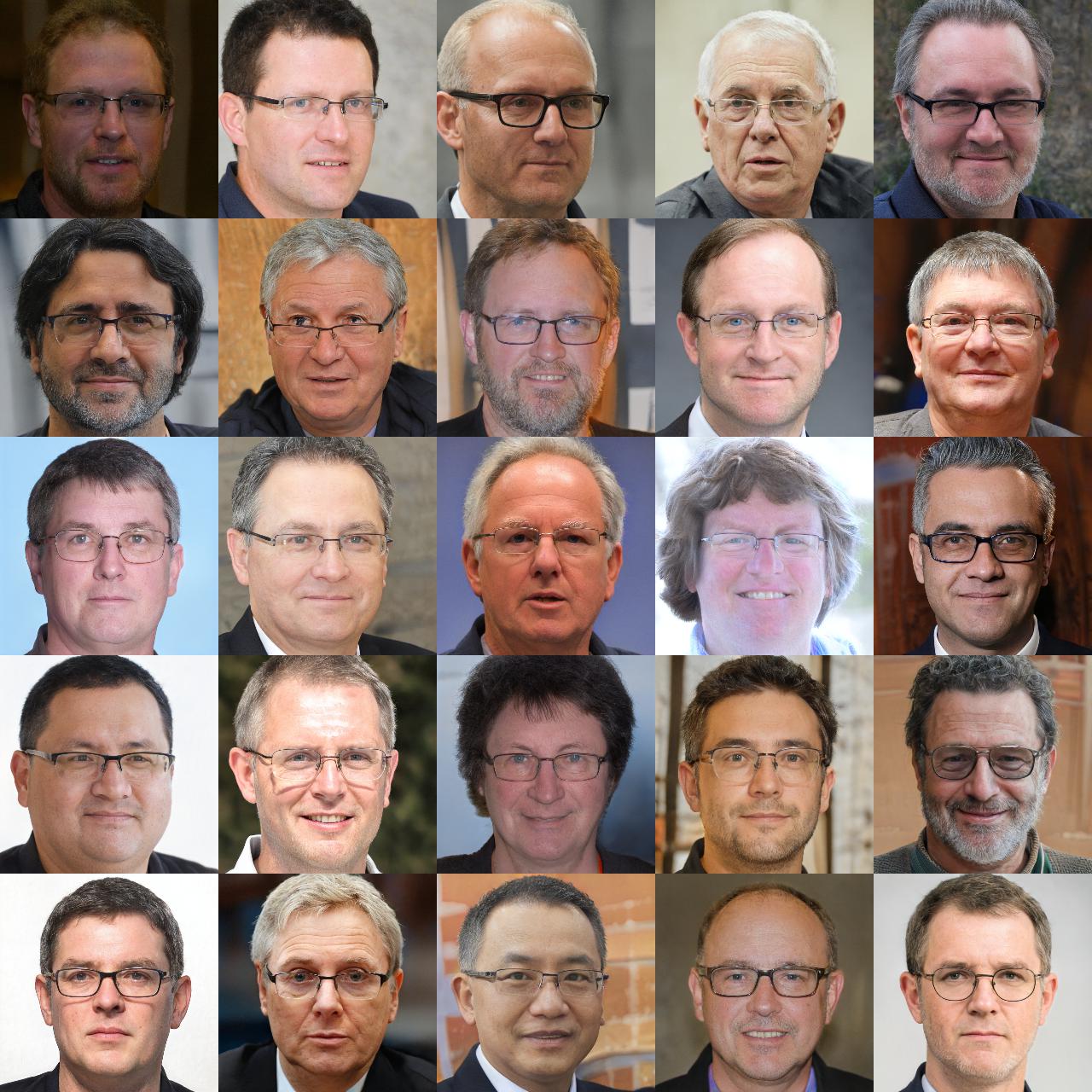}
}
\quad
\subfigure[Male without Eyeglasses]{
\includegraphics[width=8.32cm]{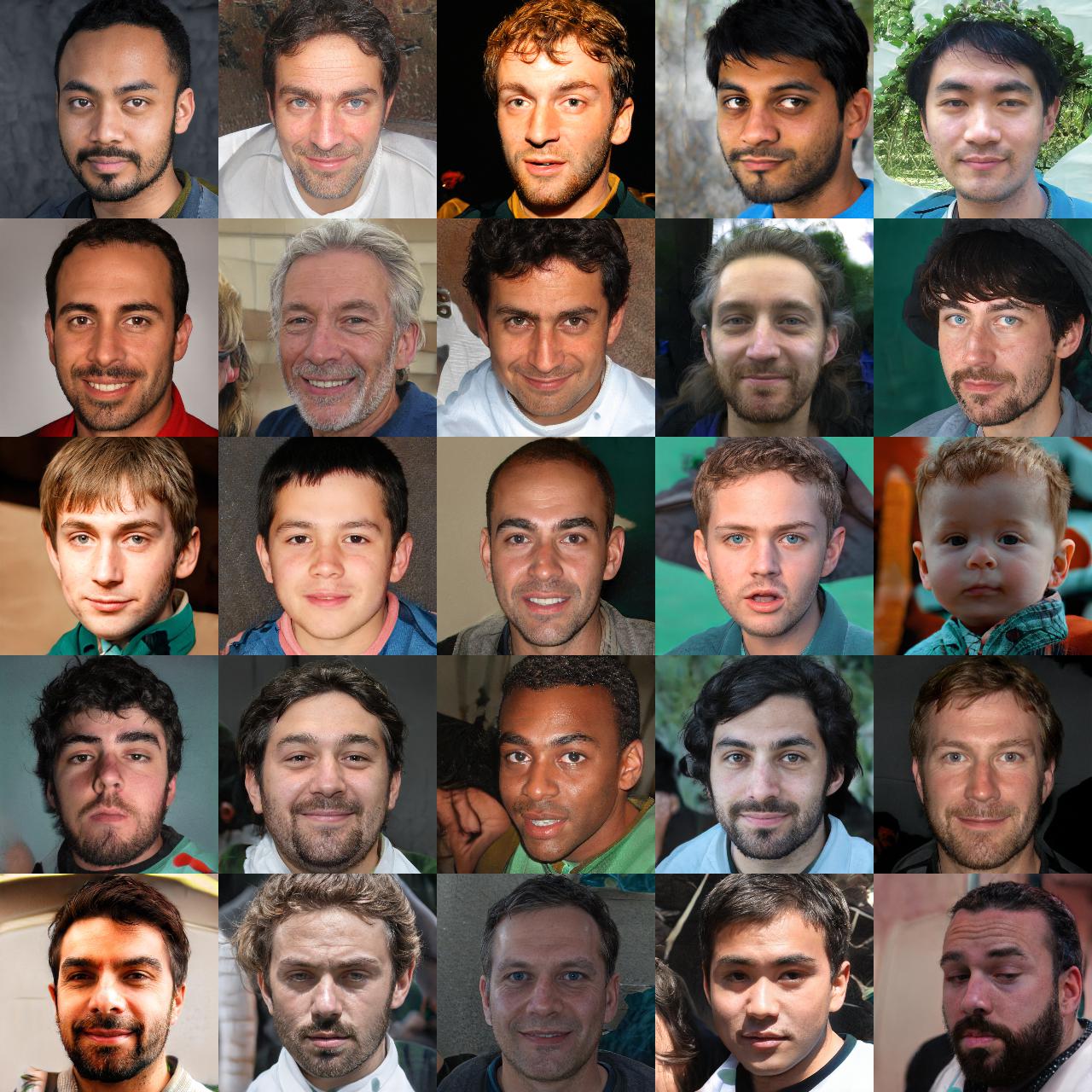}
}
\quad
\subfigure[Female with Eyeglasses]{
\includegraphics[width=8.32cm]{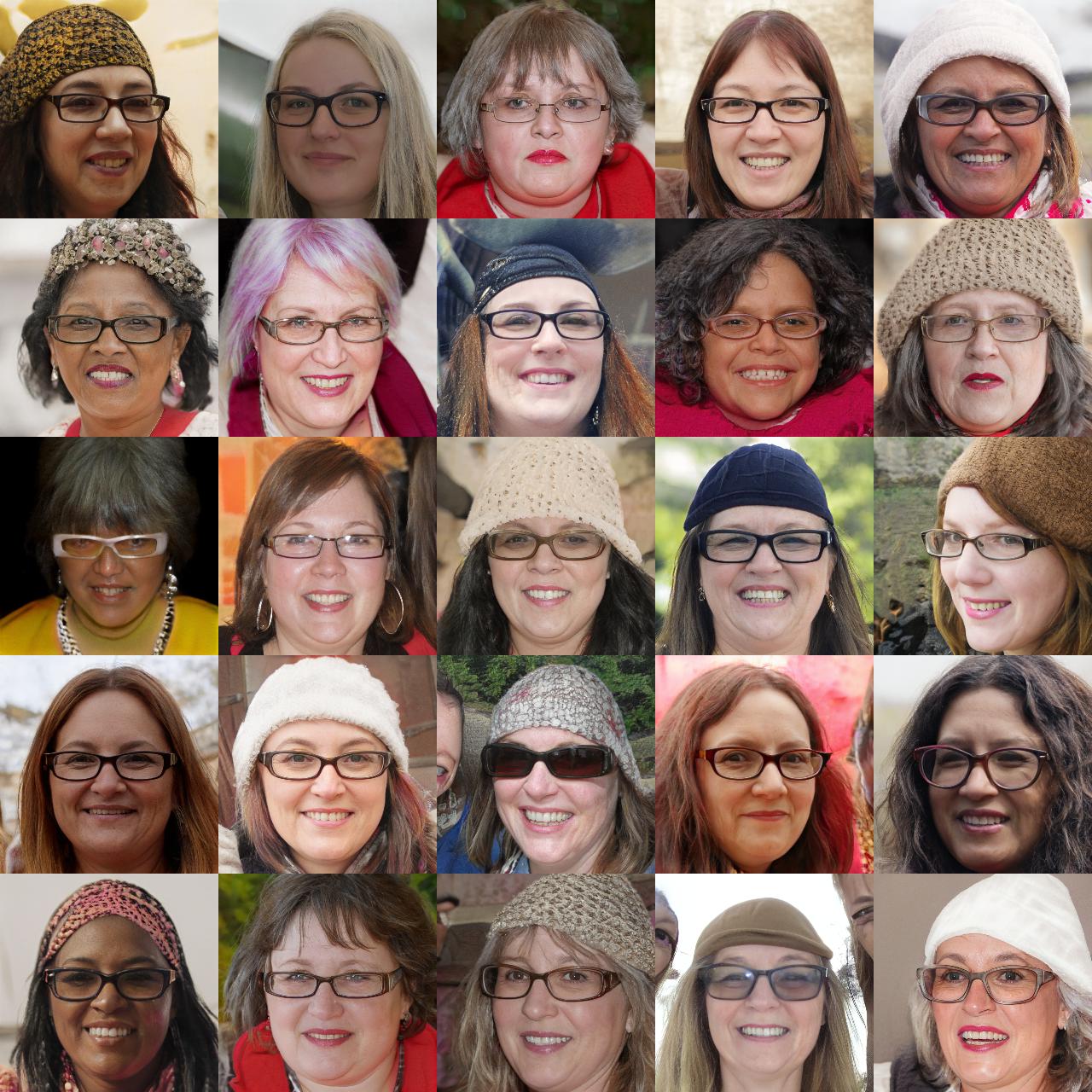}
}
\quad
\subfigure[Female without Eyeglasses]{
\includegraphics[width=8.32cm]{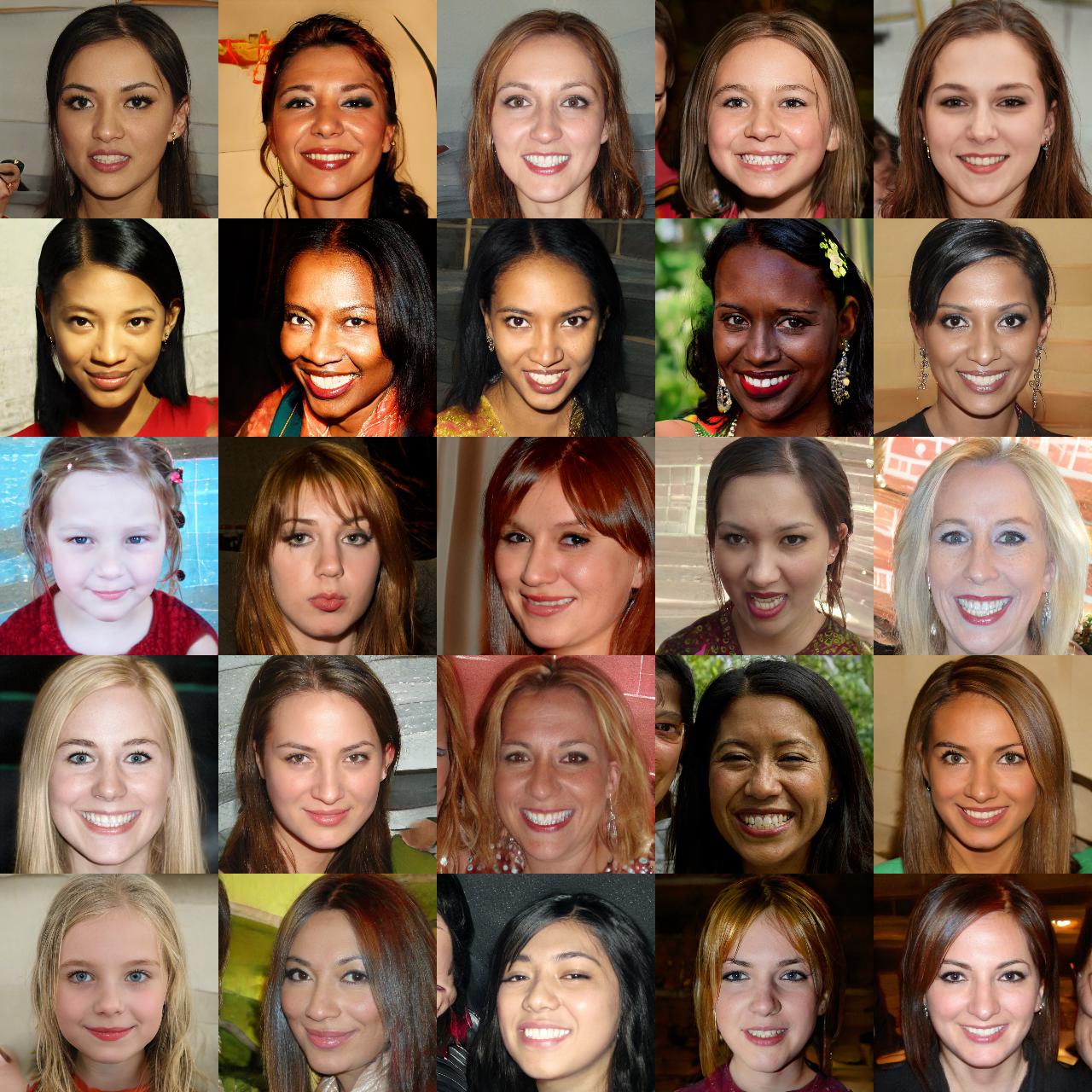}
}
\caption{Qualitative results for fair image generation in GANs with Gender and Eyeglasses.}
\end{figure}


\vspace*{\fill} 

%% file: sections_supp/sample_plots_black_age.tex
\vspace*{3.2em} 

\begin{figure}[H]
\hsize=\textwidth
\centering
\subfigure[Young Black]{
\includegraphics[width=8.32cm]{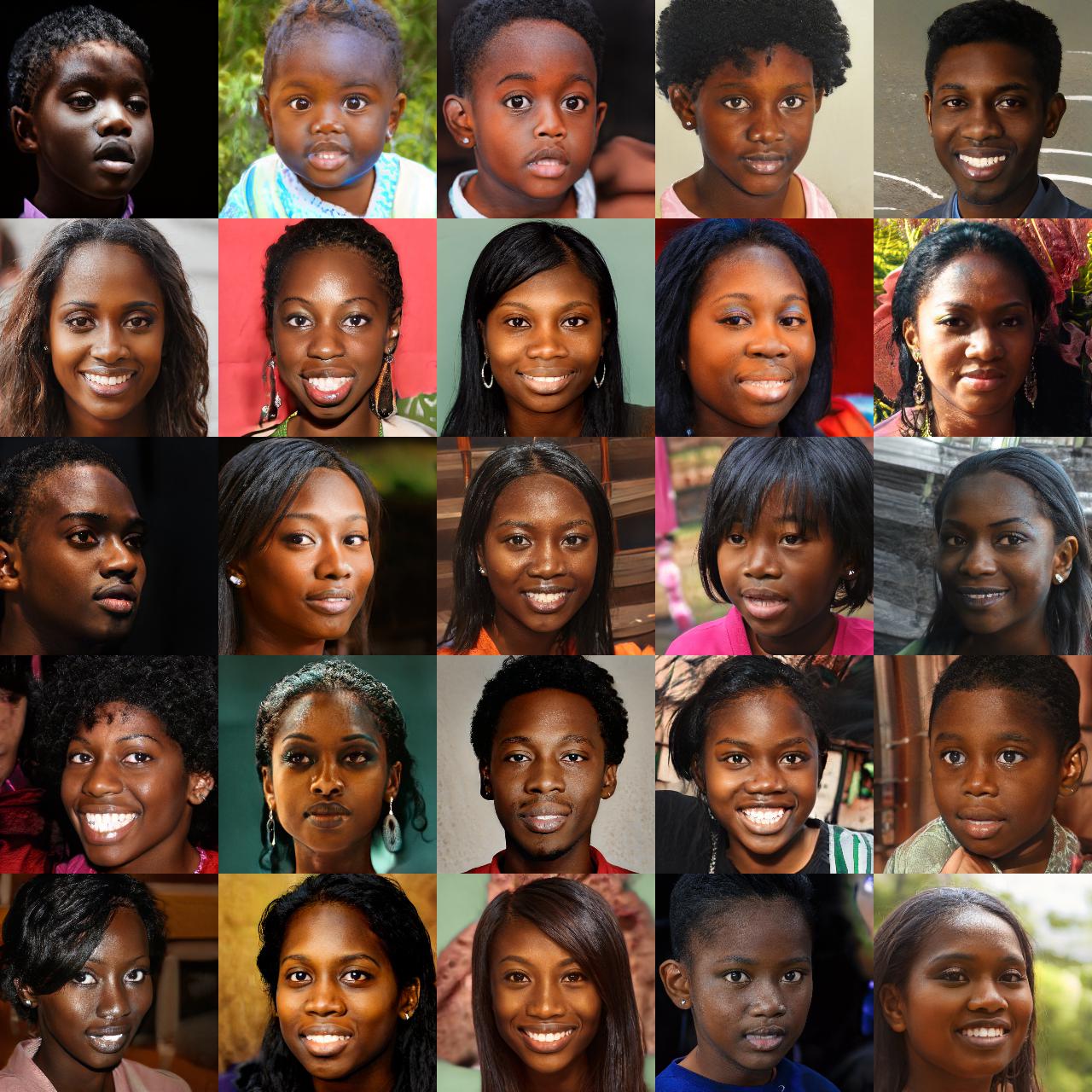}
}
\quad
\subfigure[Old Black]{
\includegraphics[width=8.32cm]{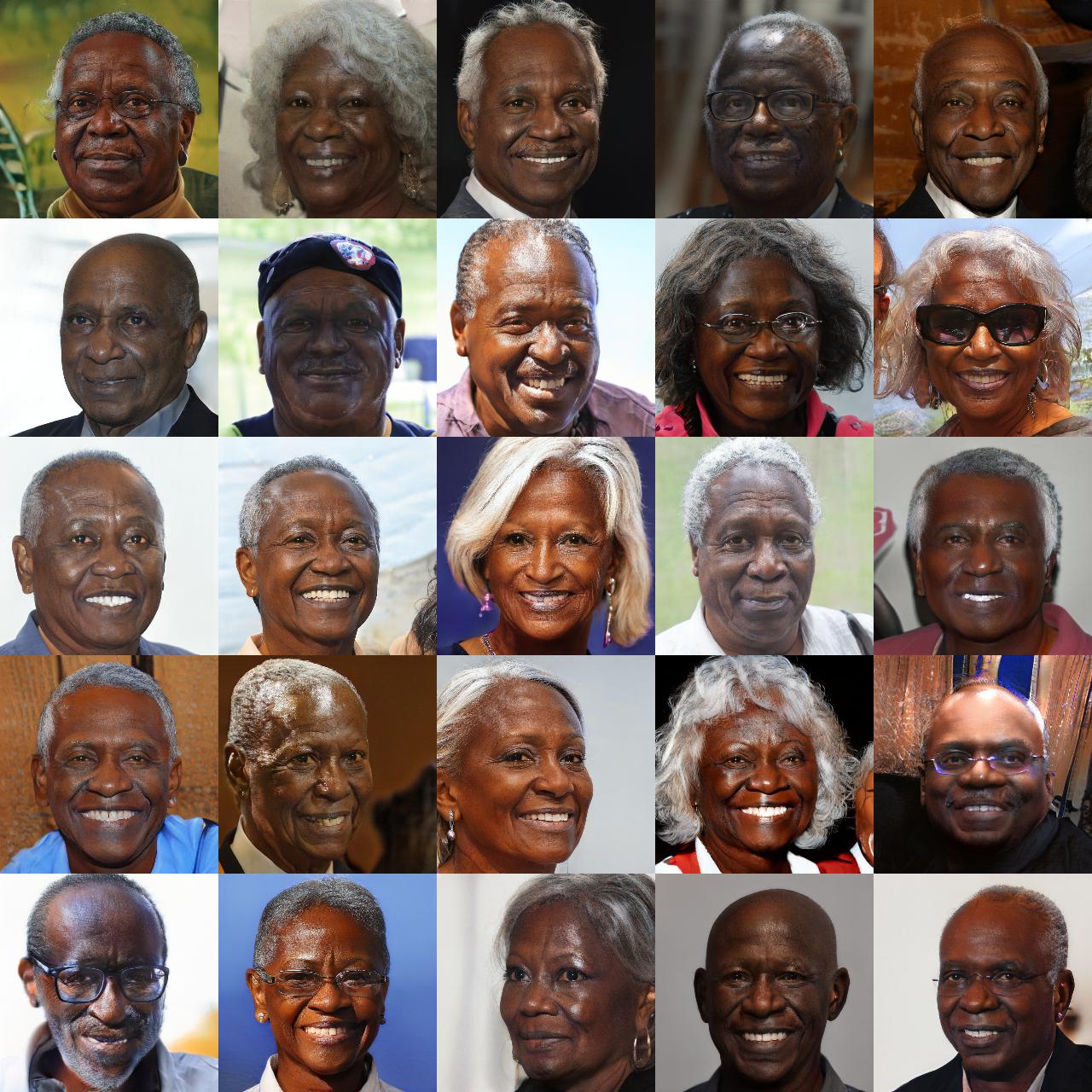}
}
\quad
\subfigure[Young Non-Black]{
\includegraphics[width=8.32cm]{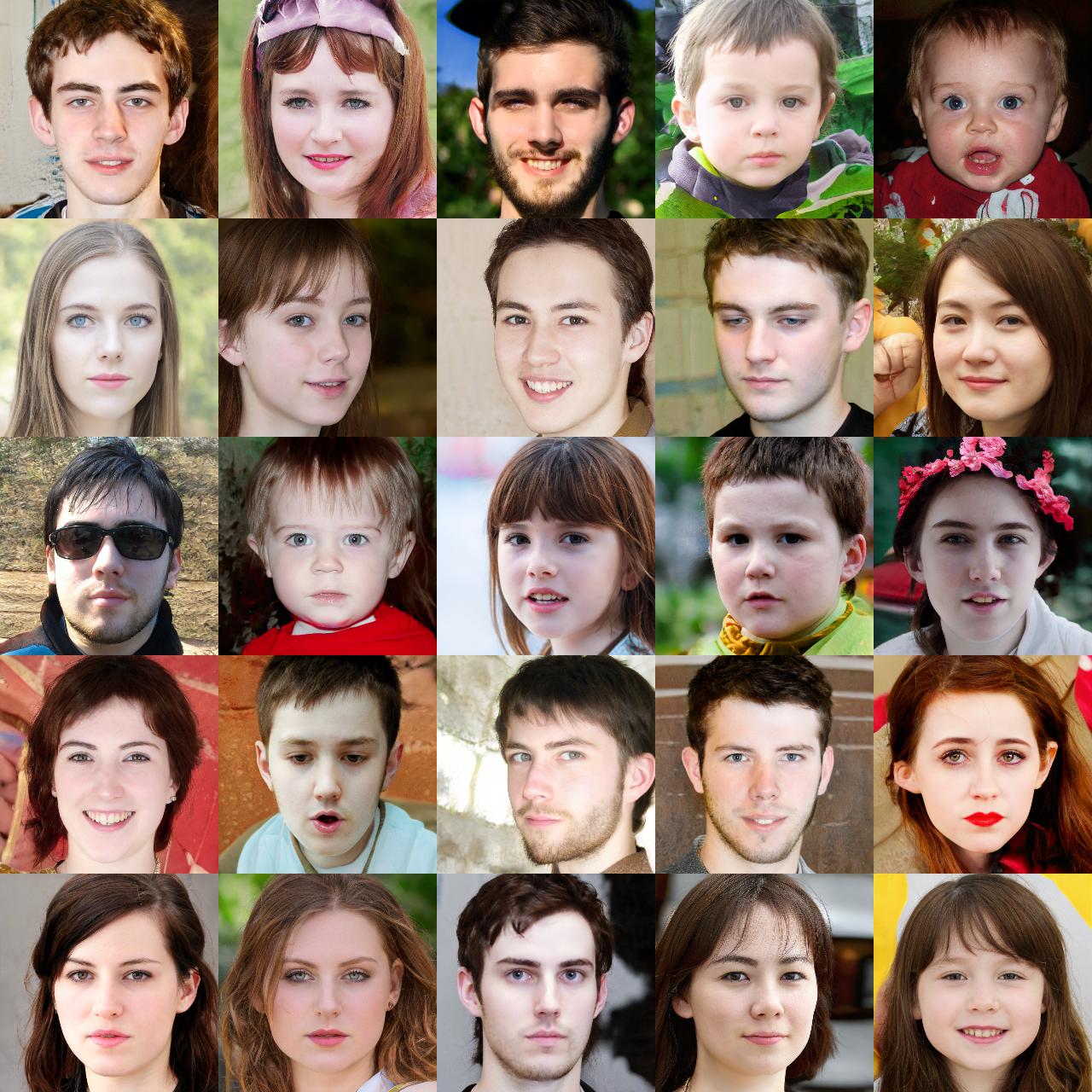}
}
\quad
\subfigure[Old Non-Black]{
\includegraphics[width=8.32cm]{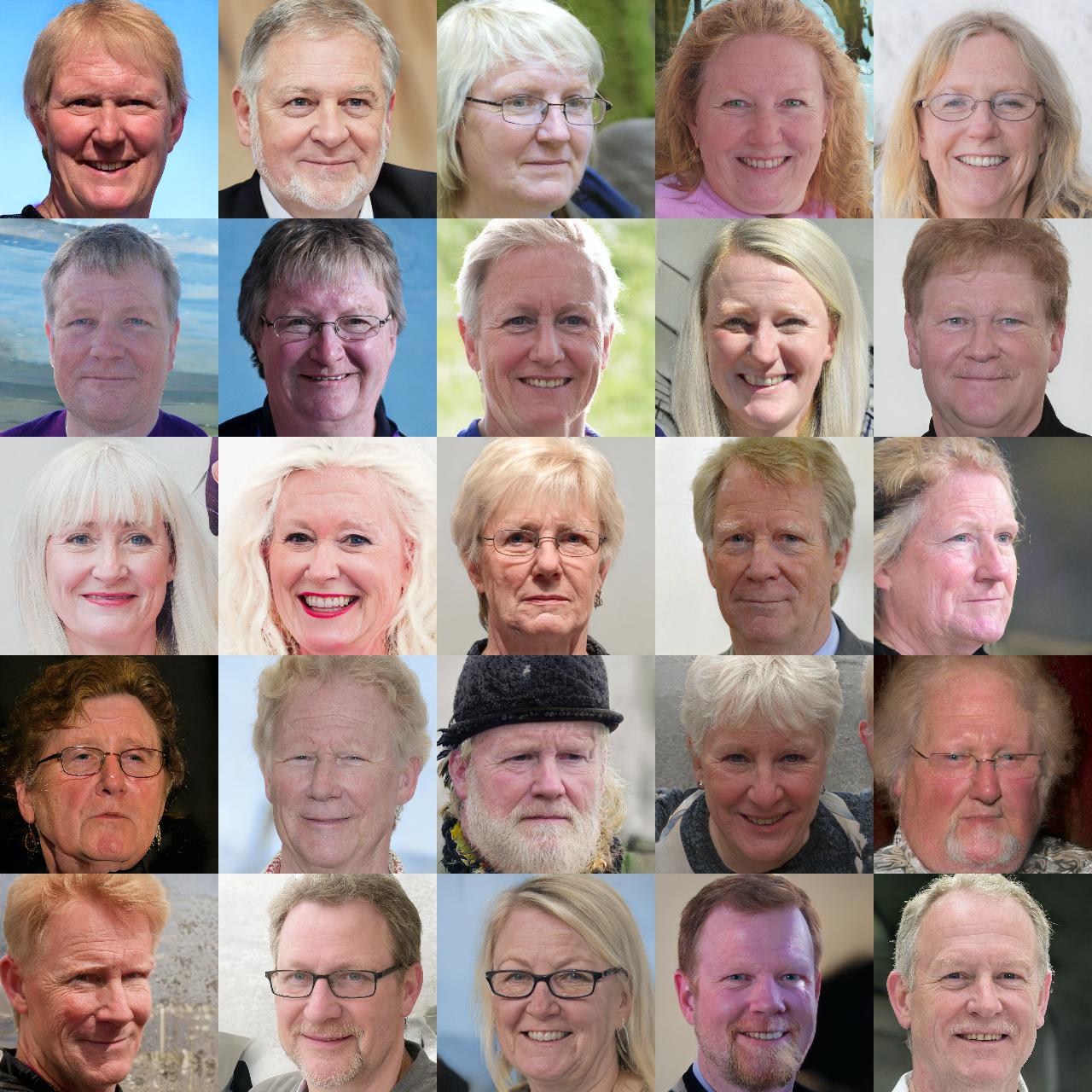}
}
\caption{Qualitative results for fair image generation in GANs with Black and Age.}
\end{figure}

\vspace*{\fill}  

%% file: sections_supp/sample_plots_black_gender.tex
\vspace*{3.2em} 

\begin{figure}[H]
\hsize=\textwidth
\centering
\subfigure[Male Black]{
\includegraphics[width=8.32cm]{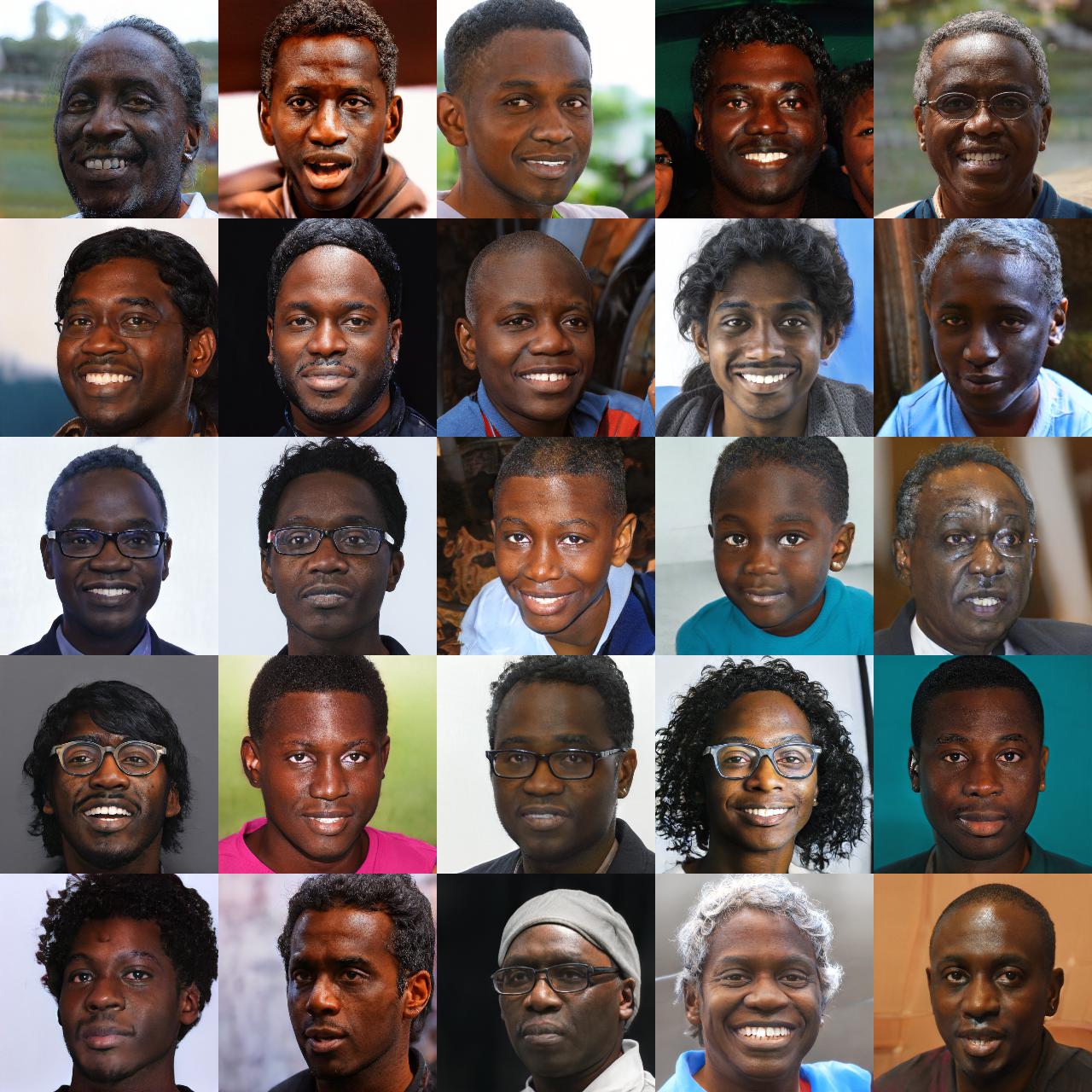}
}
\quad
\subfigure[Female Black]{
\includegraphics[width=8.32cm]{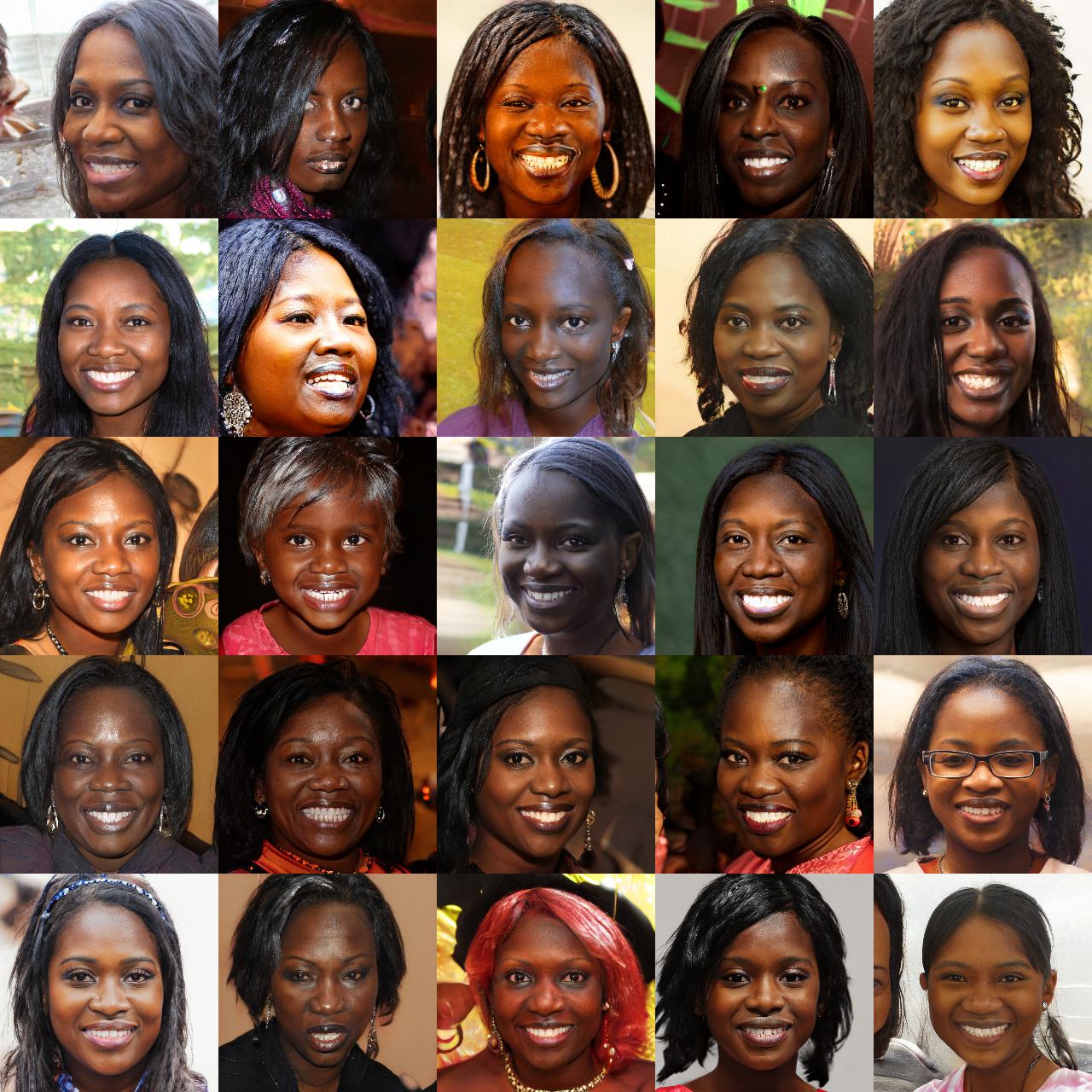}
}
\quad
\subfigure[Male Non-Black]{
\includegraphics[width=8.32cm]{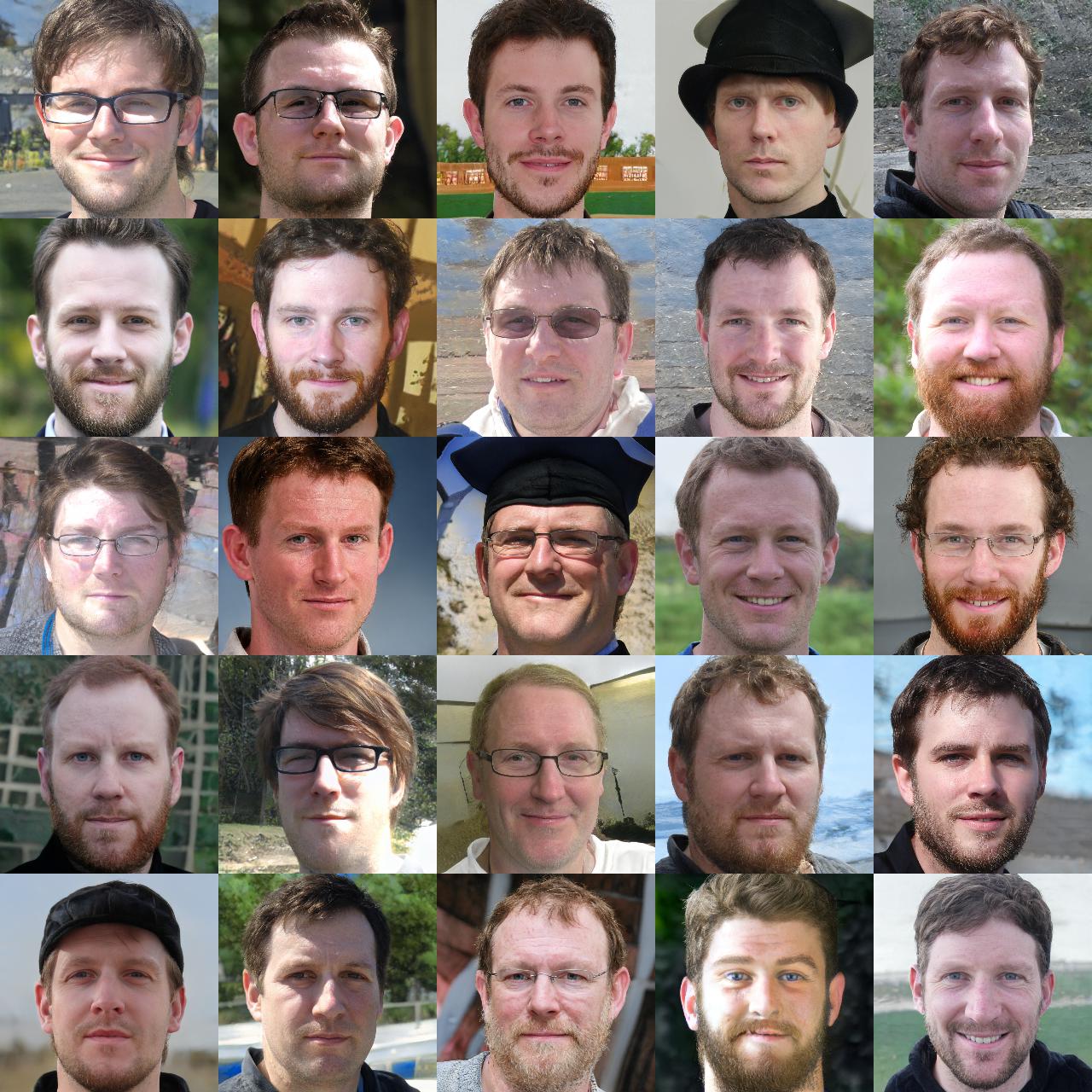}
}
\quad
\subfigure[Female Non-Black]{
\includegraphics[width=8.32cm]{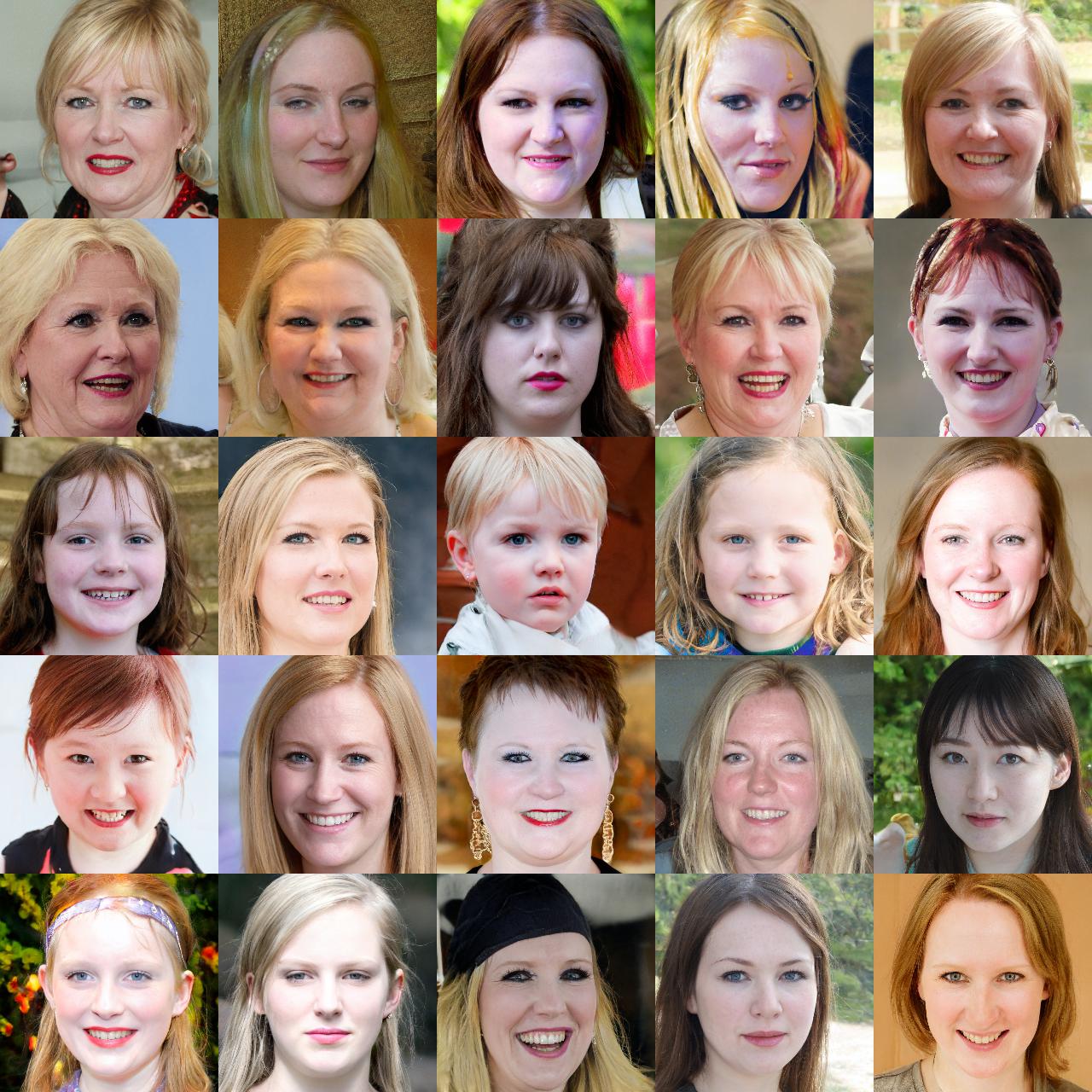}
}
\caption{Qualitative results for fair image generation in GANs with Black and Gender.}
\end{figure}

\vspace*{\fill}  

%% file: sections_supp/sample_plots_asian_age.tex
\vspace*{3.2em} 

\begin{figure}[H]
\hsize=\textwidth
\centering
\subfigure[Young Asian]{
\includegraphics[width=8.32cm]{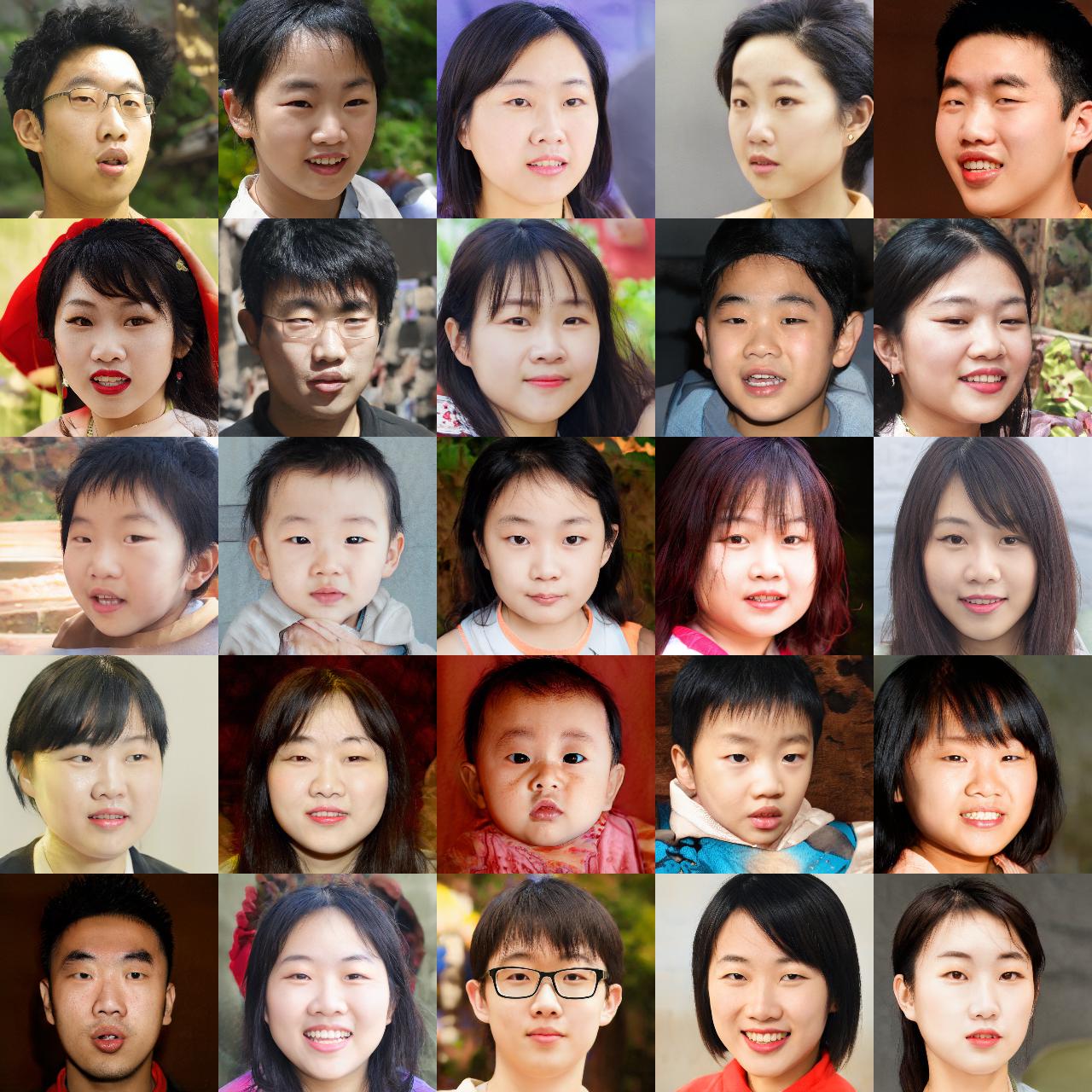}
}
\quad
\subfigure[Old Asian]{
\includegraphics[width=8.32cm]{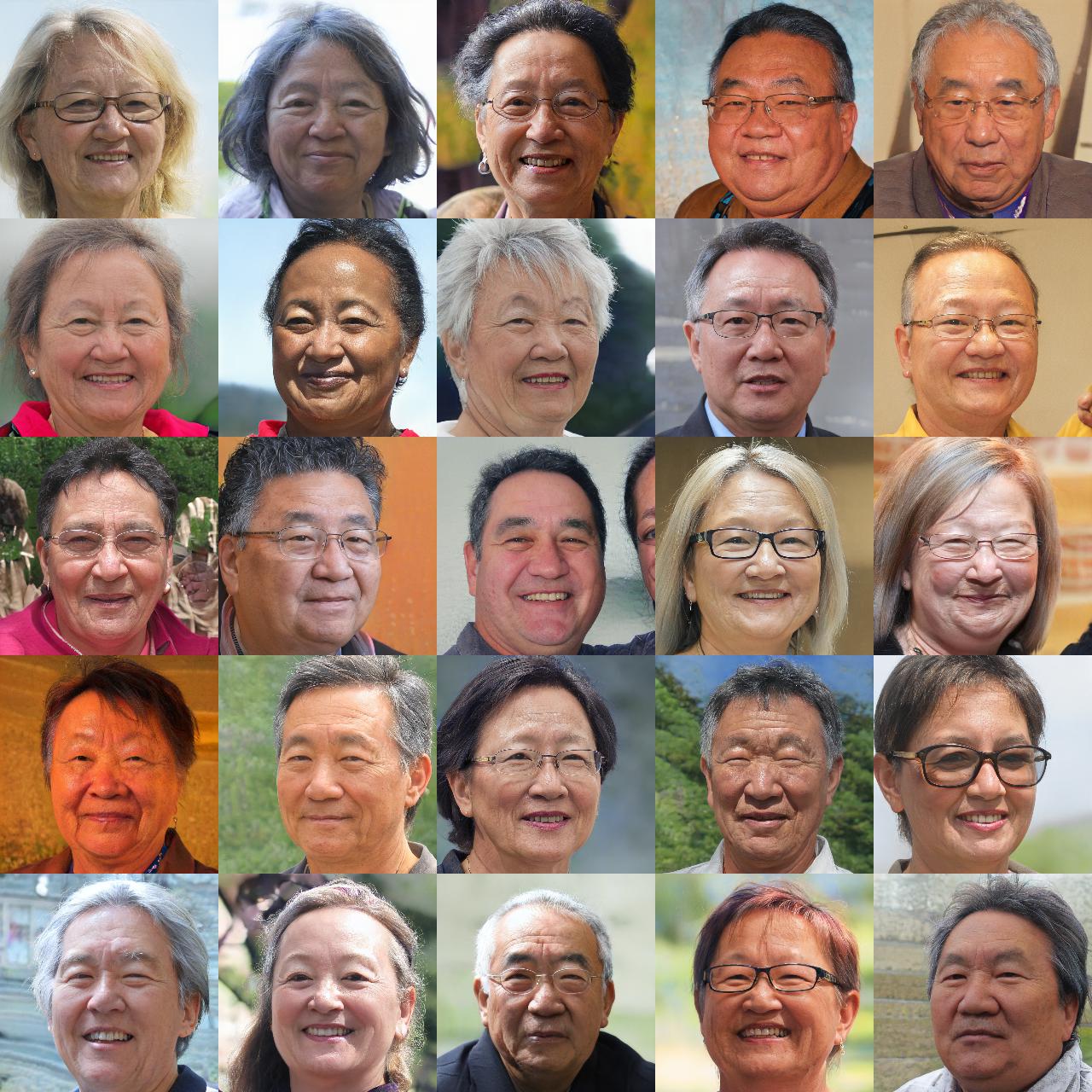}
}
\quad
\subfigure[Young Non-Asian]{
\includegraphics[width=8.32cm]{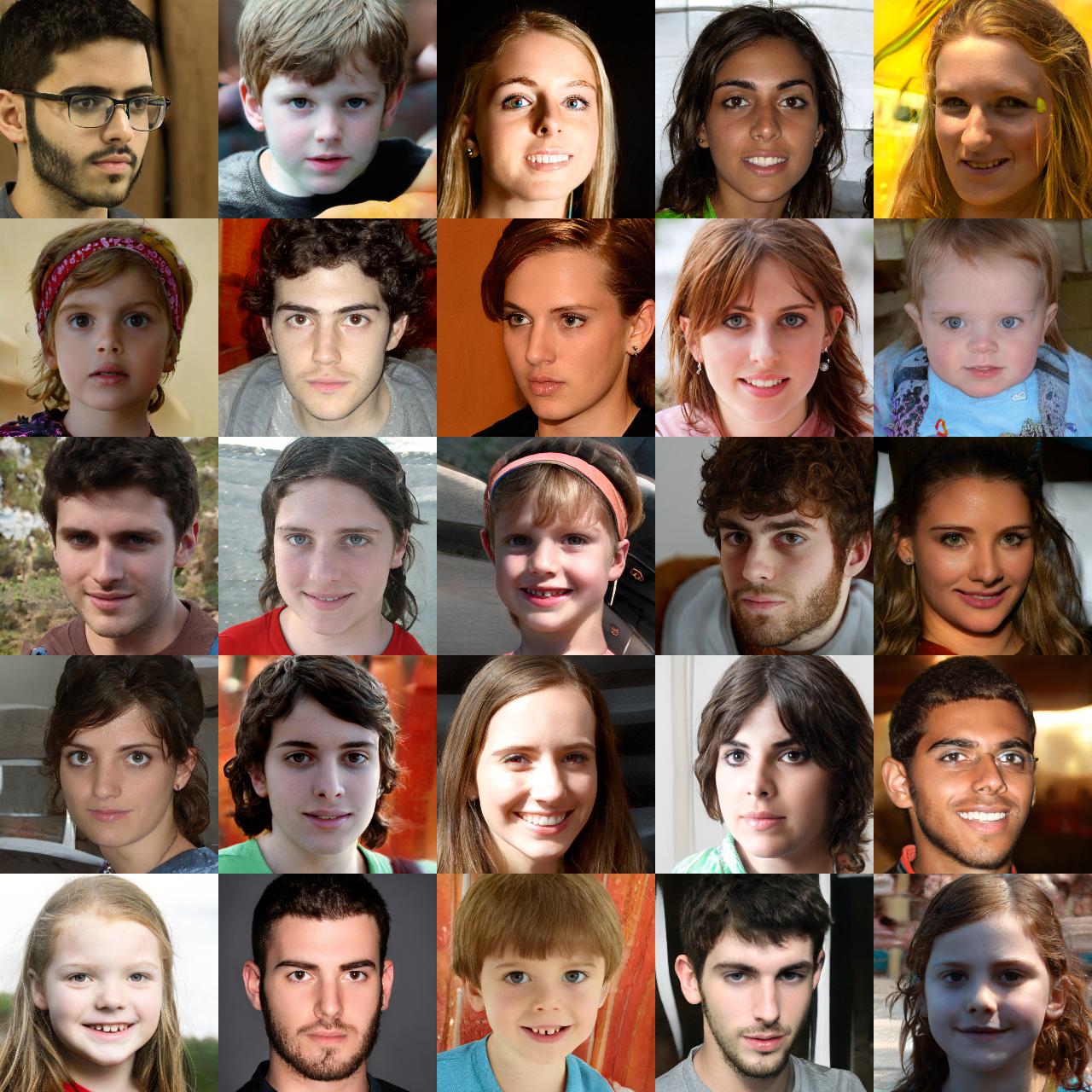}
}
\quad
\subfigure[Old Non-Asian]{
\includegraphics[width=8.32cm]{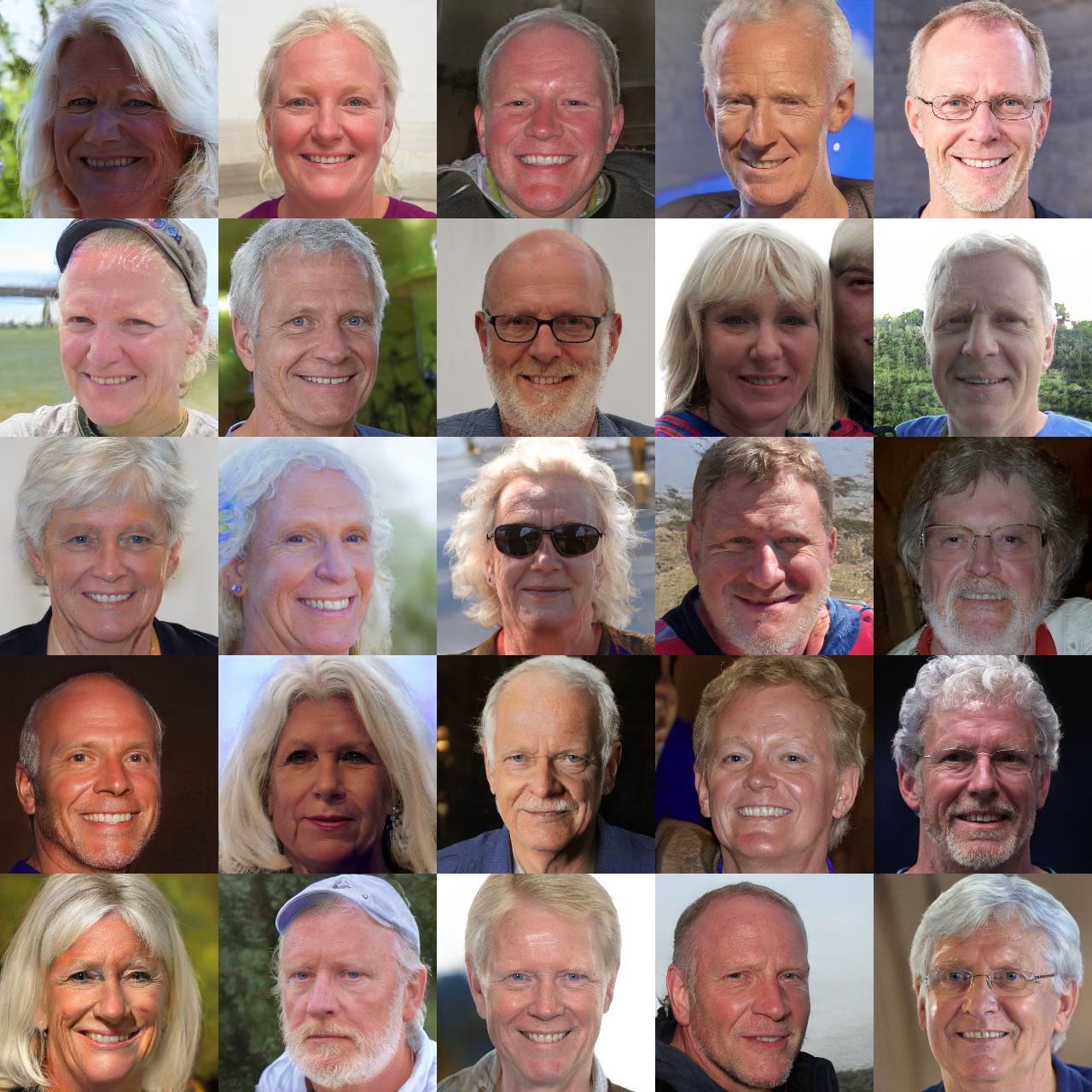}
}
\caption{Qualitative results for fair image generation in GANs with Asian and Age.}
\end{figure}

\vspace*{\fill}  

%% file: sections_supp/sample_plots_asian_gender.tex
\vspace*{3.2em} 

\begin{figure}[H]
\hsize=\textwidth
\centering
\subfigure[Male Asian]{
\includegraphics[width=8.32cm]{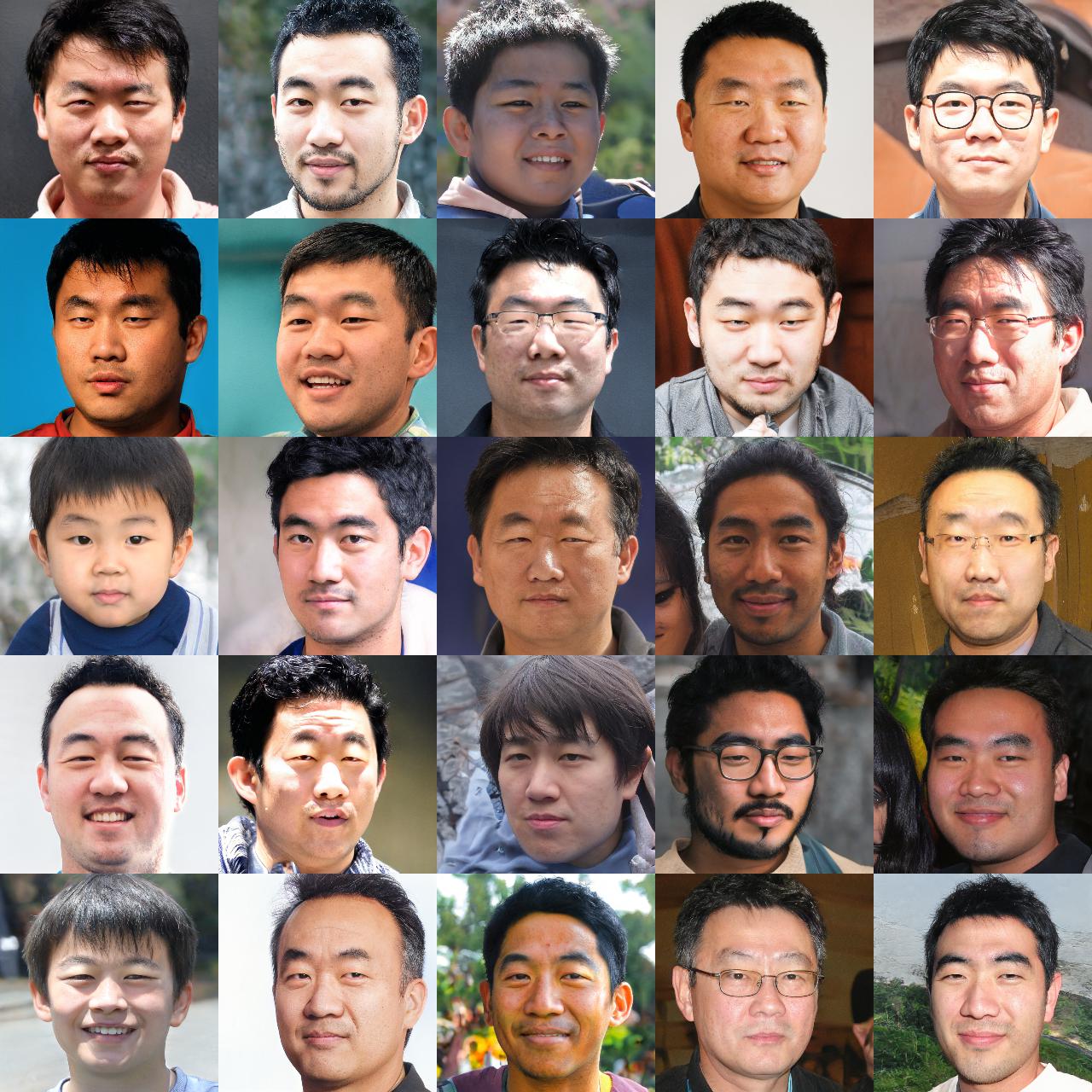}
}
\quad
\subfigure[Female Asian]{
\includegraphics[width=8.32cm]{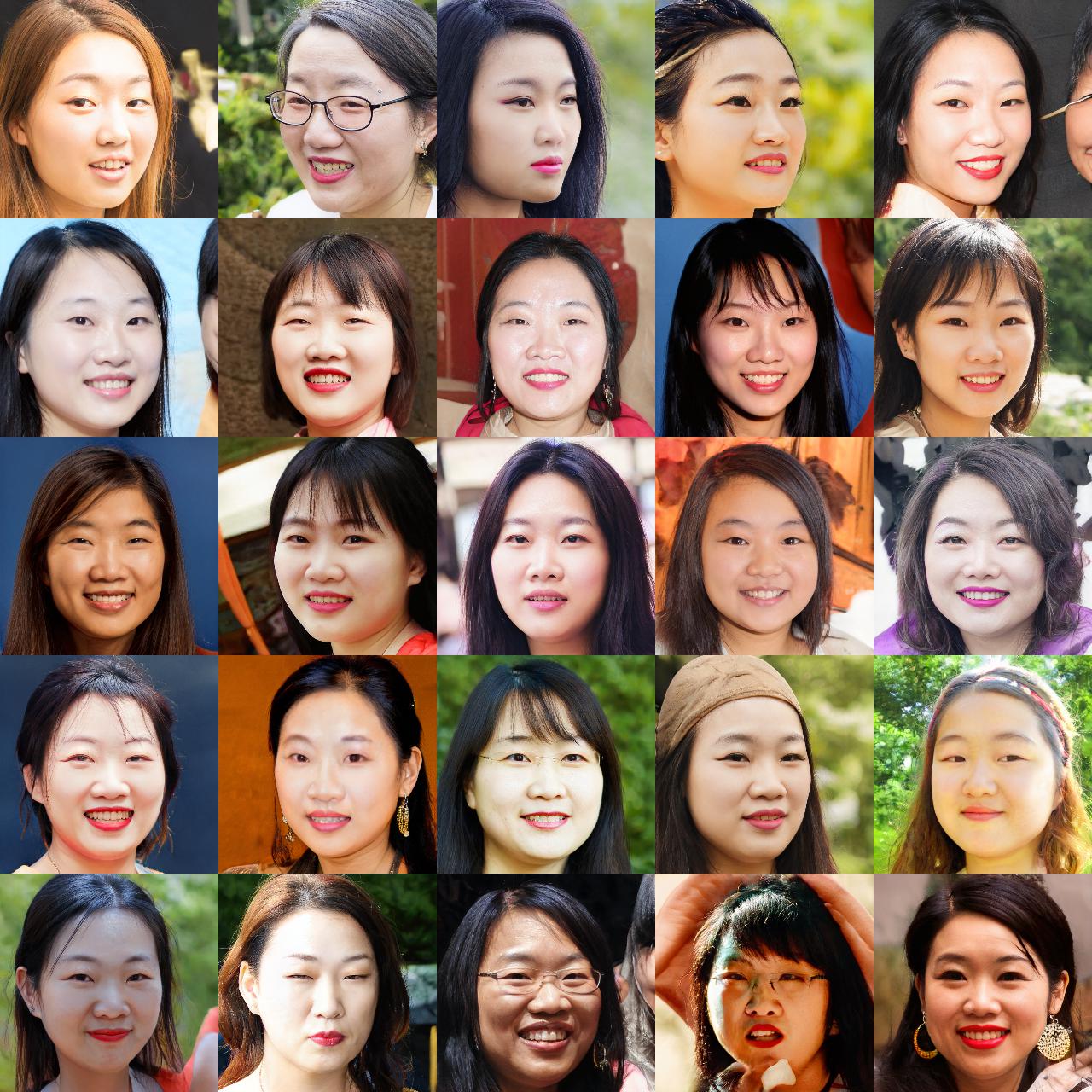}
}
\quad
\subfigure[Male Non-Asian]{
\includegraphics[width=8.32cm]{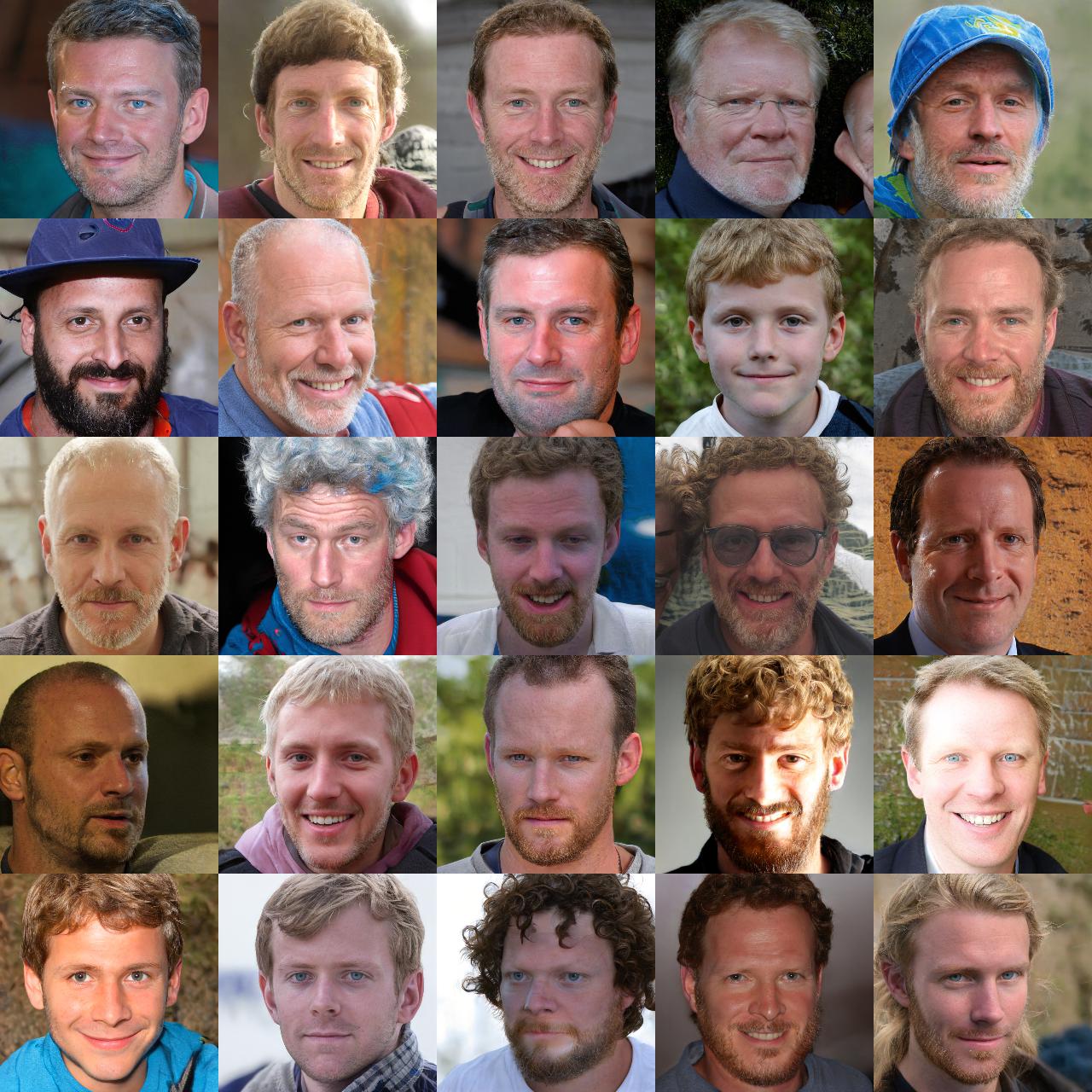}
}
\quad
\subfigure[Female Non-Asian]{
\includegraphics[width=8.32cm]{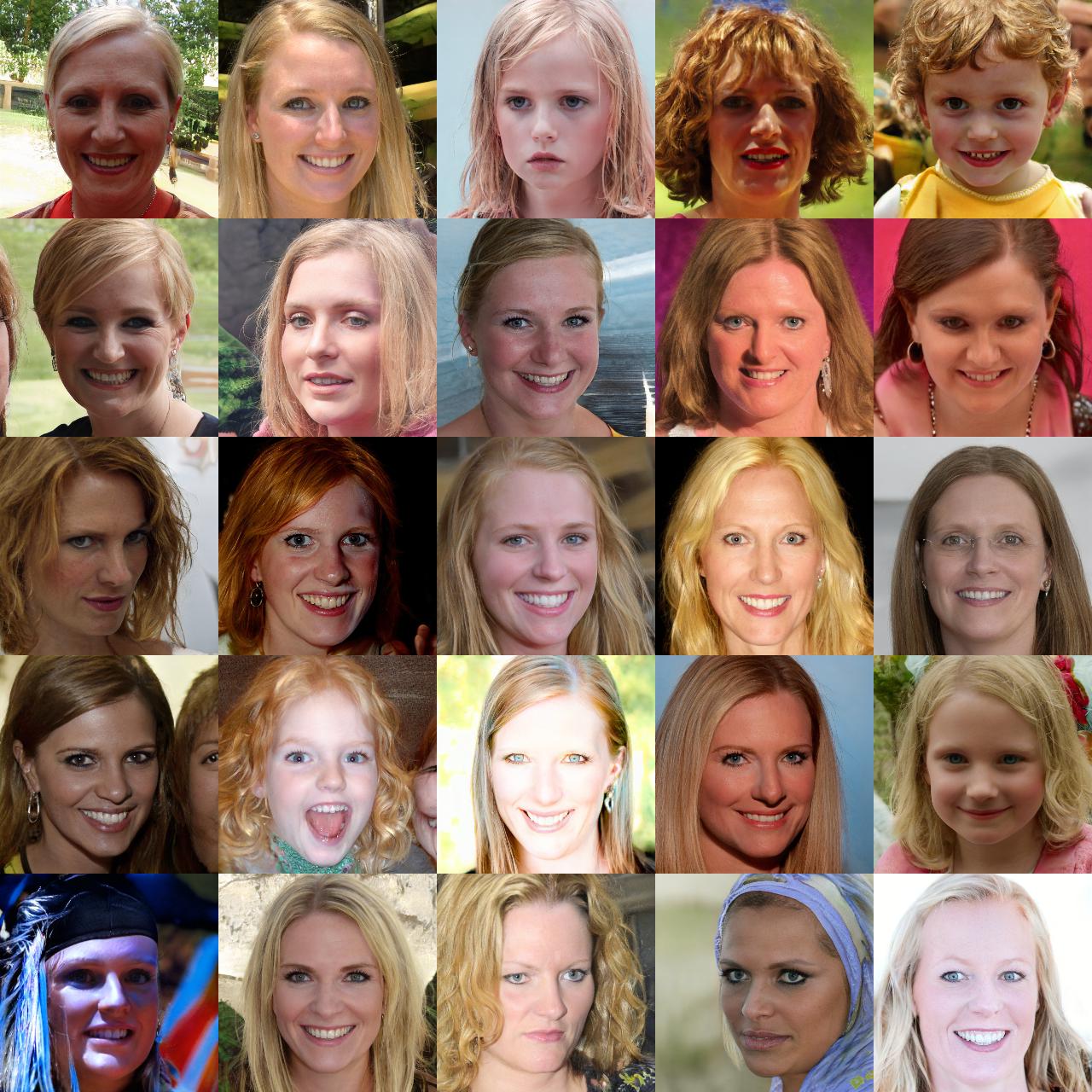}
}
\caption{Qualitative results for fair image generation in GANs with Asian and Gender.}
\end{figure}

\vspace*{\fill}  

%% file: sections_supp/sample_plots_gender_blackhair.tex
\vspace*{3.2em} 

\begin{figure}[H]
\hsize=\textwidth
\centering
\subfigure[Male with Black Hair]{
\includegraphics[width=8.32cm]{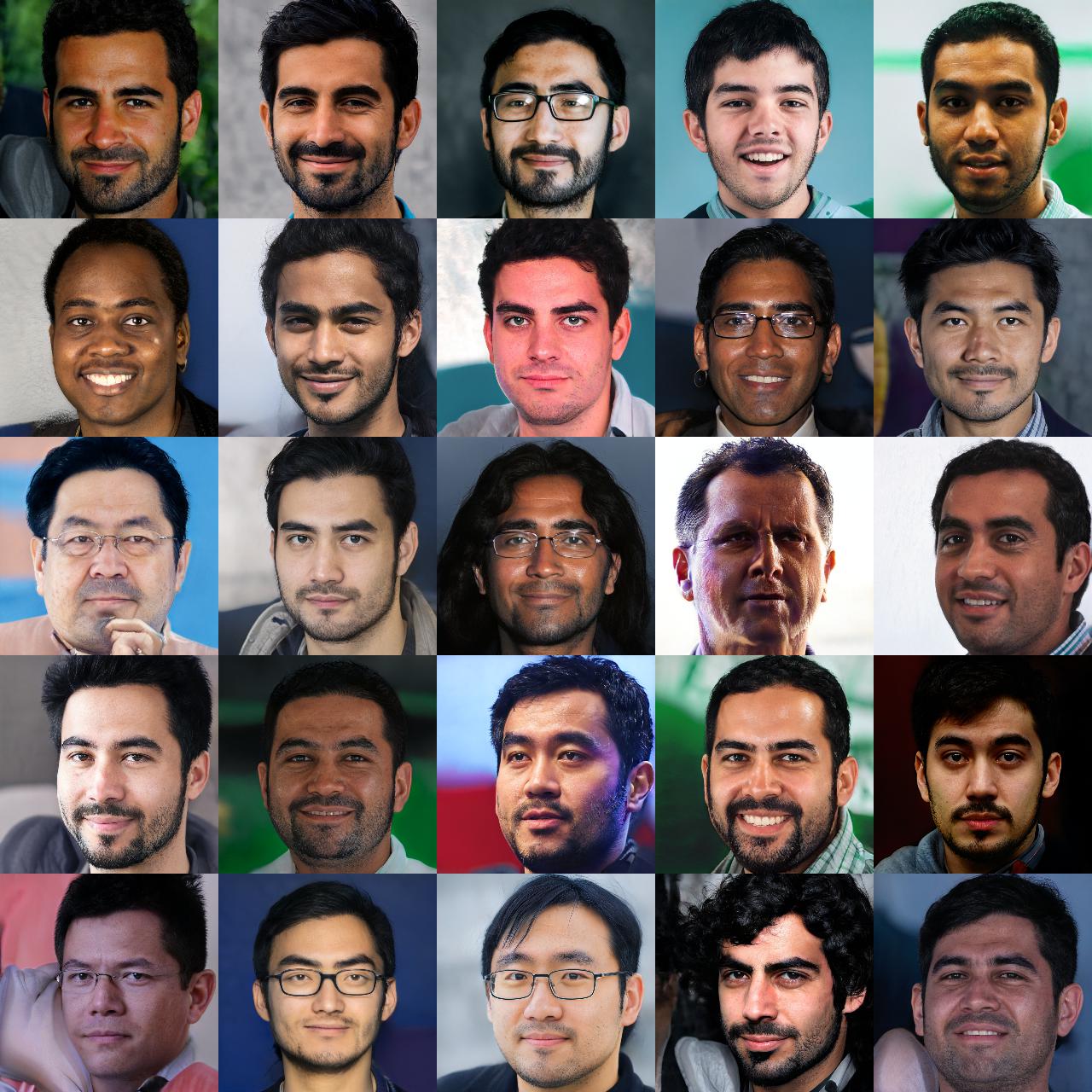}
}
\quad
\subfigure[Male without Black Hair]{
\includegraphics[width=8.32cm]{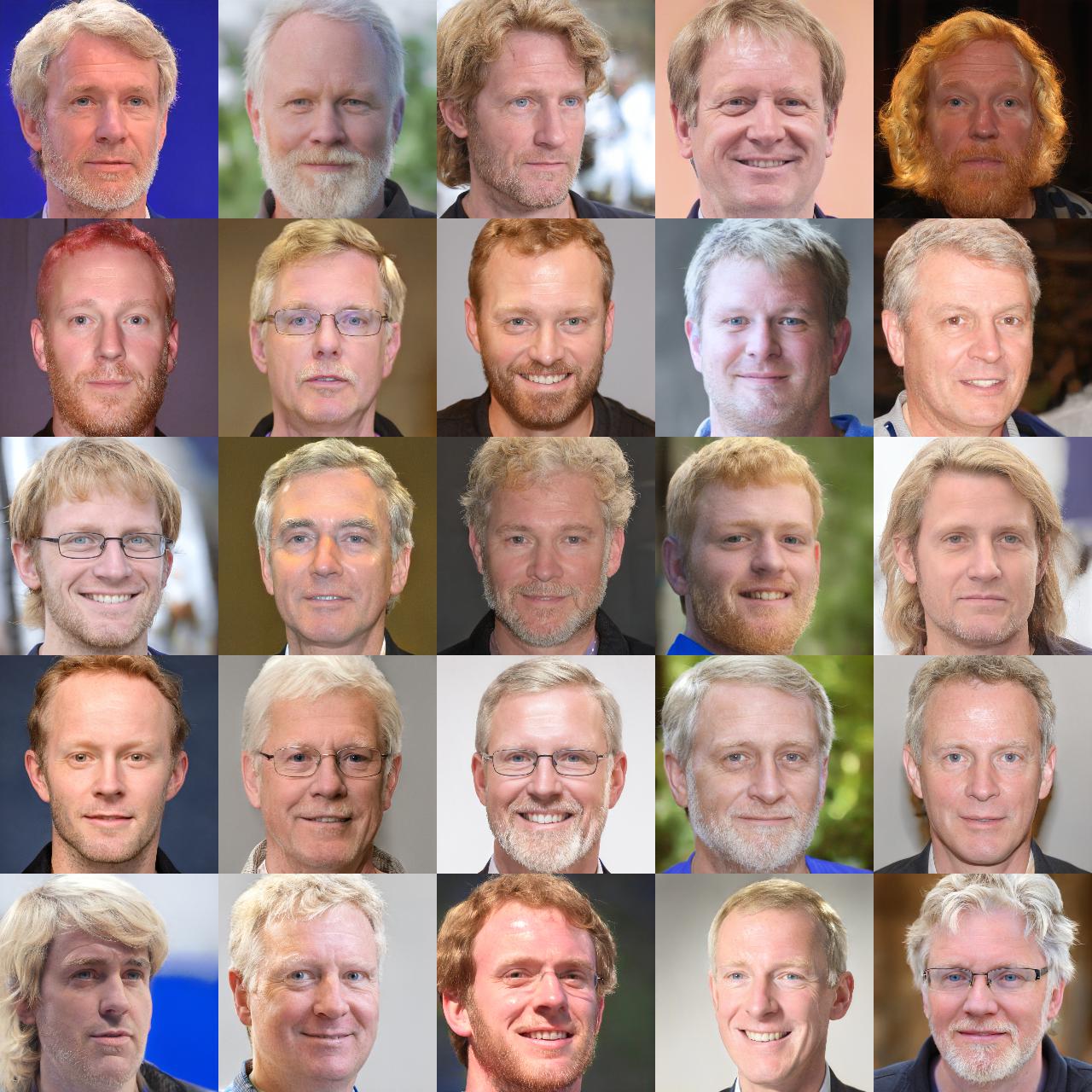}
}
\quad
\subfigure[Female with Black Hair]{
\includegraphics[width=8.32cm]{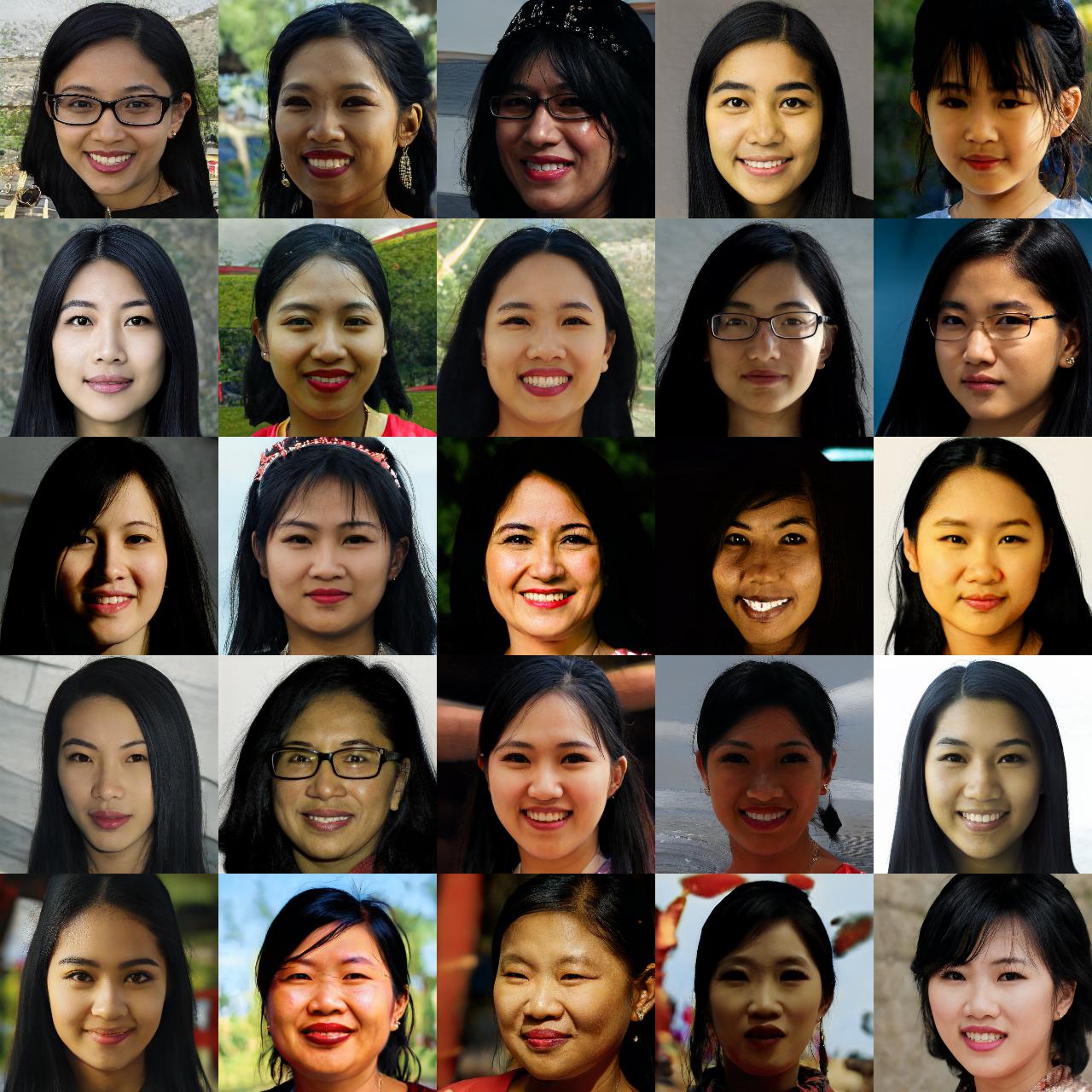}
}
\quad
\subfigure[Female without Black Hair]{
\includegraphics[width=8.32cm]{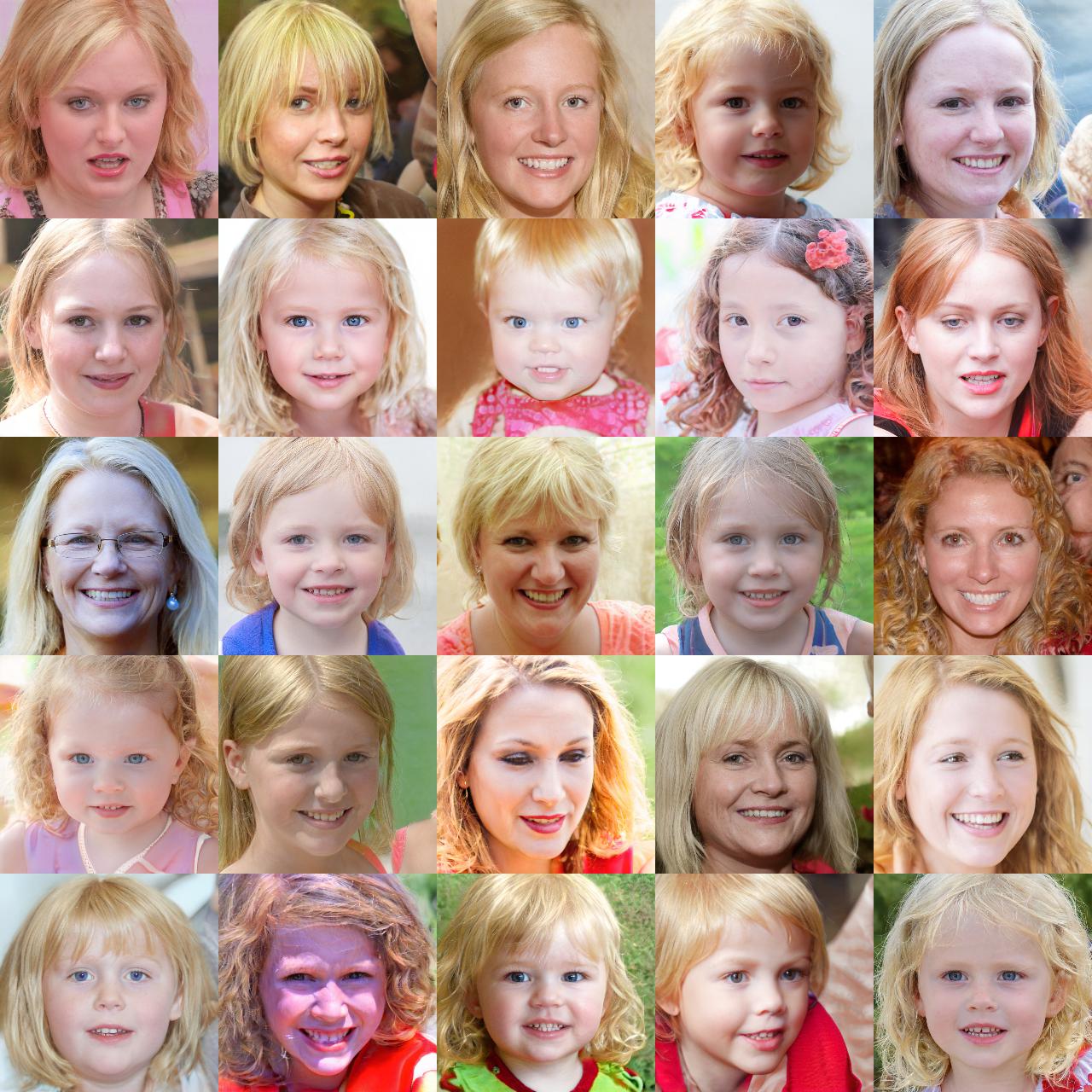}
}
\caption{Qualitative results for fair image generation in GANs with Gender and Black Hair.}
\end{figure}

\vspace*{\fill} 

%% file: sections_supp/sample_plots_age_smiling.tex
\vspace*{3.2em} 

\begin{figure}[H]
\hsize=\textwidth
\centering
\subfigure[Young with Smiling]{
\includegraphics[width=8.32cm]{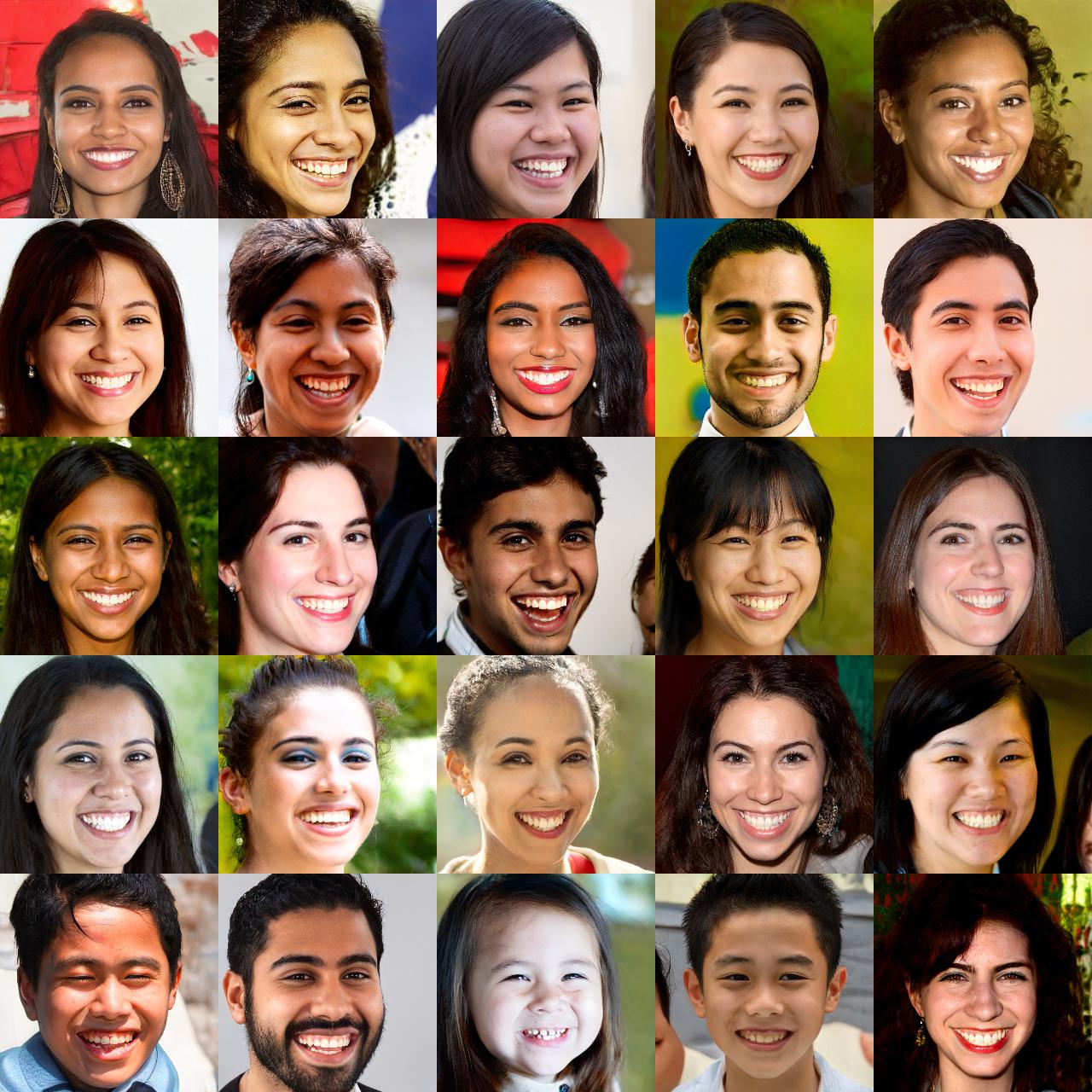}
}
\quad
\subfigure[Young without Smiling]{
\includegraphics[width=8.32cm]{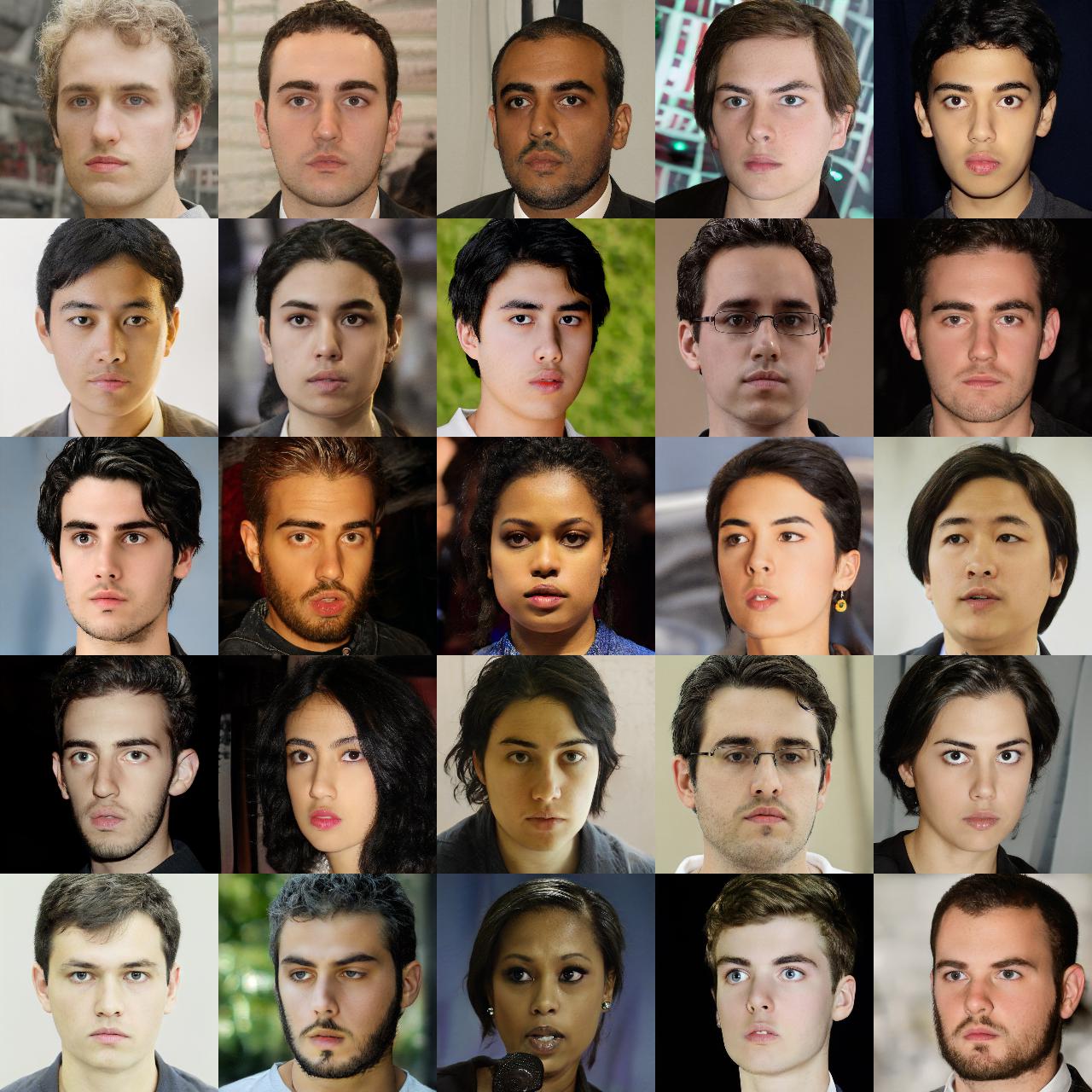}
}
\quad
\subfigure[Old with Smiling]{
\includegraphics[width=8.32cm]{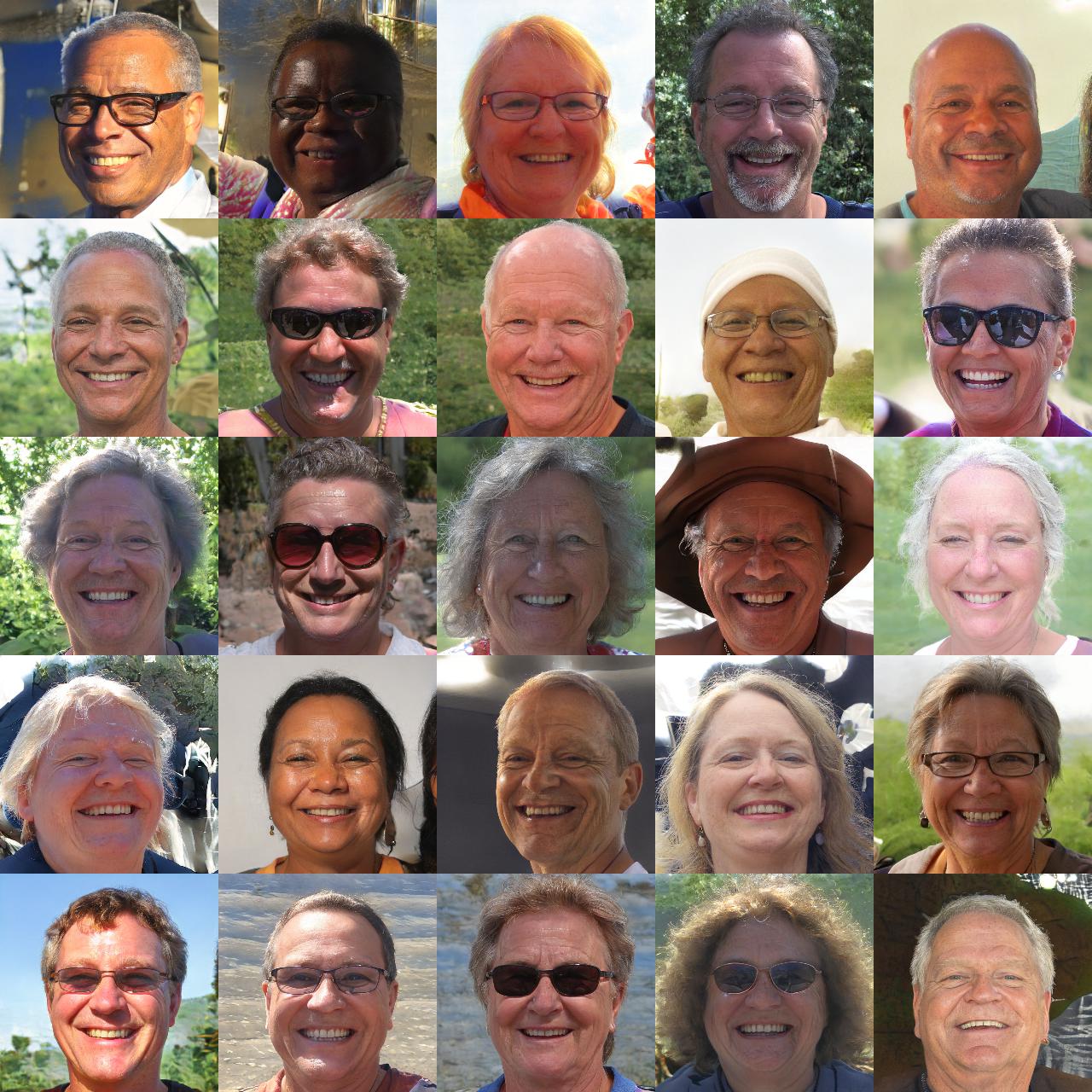}
}
\quad
\subfigure[Old without Smiling]{
\includegraphics[width=8.32cm]{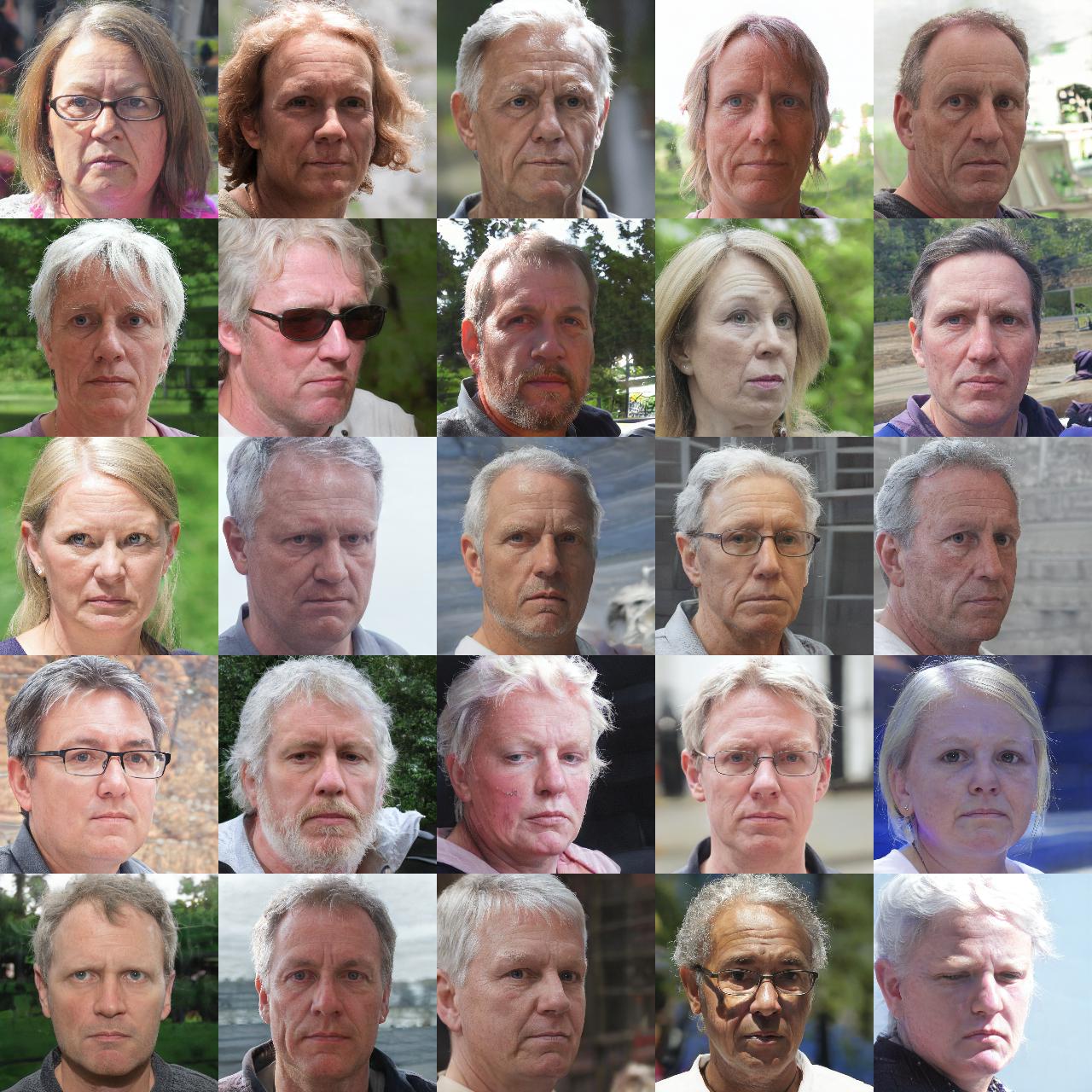}
}
\caption{Qualitative results for fair image generation in GANs with Age and Smiling.}
\end{figure}

\vspace*{\fill}  

%% file: sections_supp/pulse_samples.tex
\section{Bias in Super-Resolution Models}

\setlength{\parindent}{2em}
\par 
In this section, we show more examples of the failure cases we expose from the super-resolution model PULSE \cite{Menon_2020_CVPR}.

\setlength{\parindent}{2em}
\par 
For each set of images, from left to right we showcase 1) the original image generated by our method; 2) the LR image subsampled from the original image; 3) HR image output by PULSE given the LR image.

\begin{figure}[H]
\hsize=\textwidth
\centering
\subfigure{
\includegraphics[width=5.30cm]{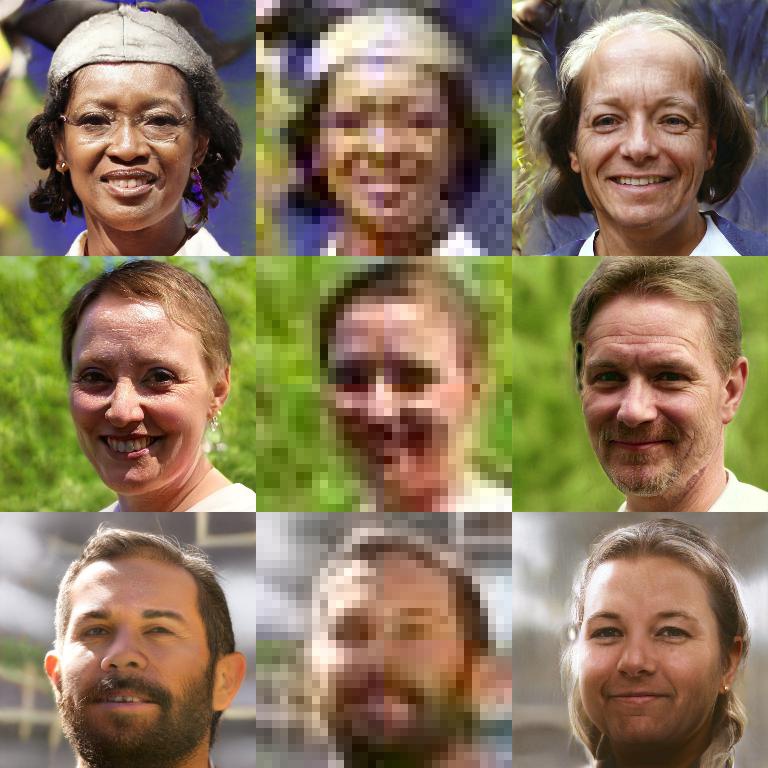}
}
\quad
\subfigure{
\includegraphics[width=5.30cm]{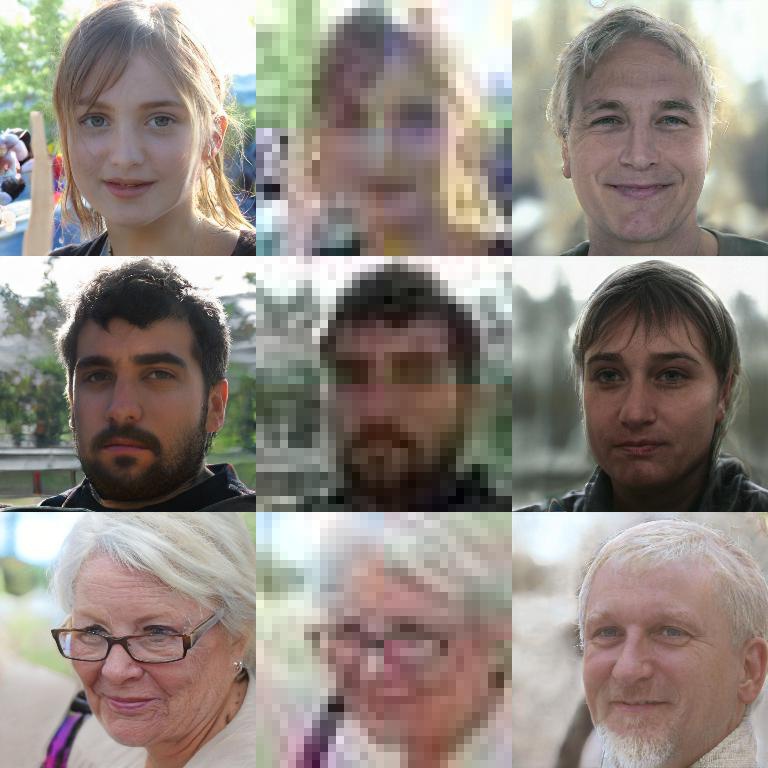}
}
\quad
\subfigure{
\includegraphics[width=5.30cm]{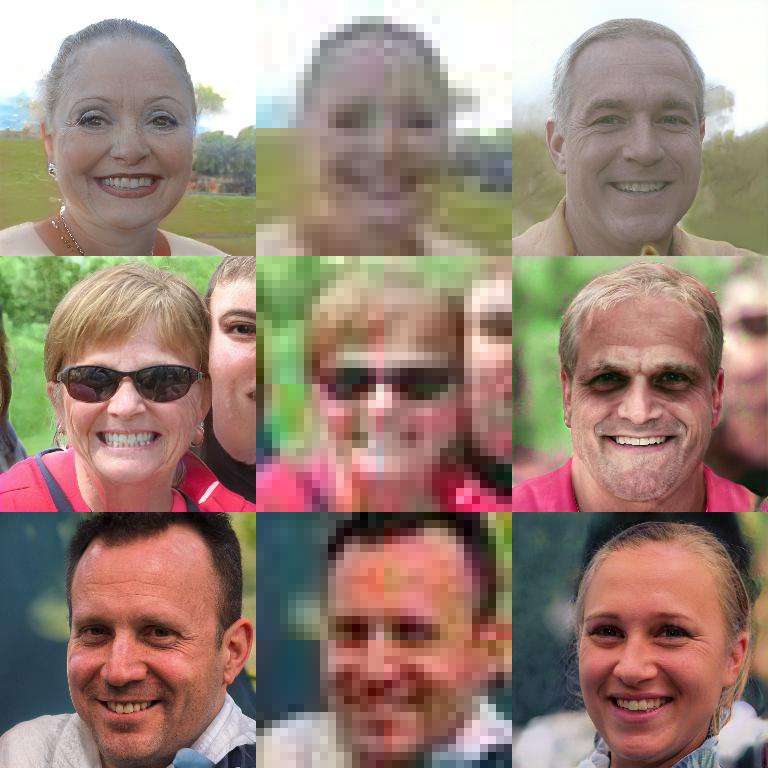}
}
\caption{Examples of Gender attribute alteration by the super-resolution model.}
\label{fig:PULSE_gender}
\end{figure}

\begin{figure}[H]
\hsize=\textwidth
\centering
\subfigure{
\includegraphics[width=5.30cm]{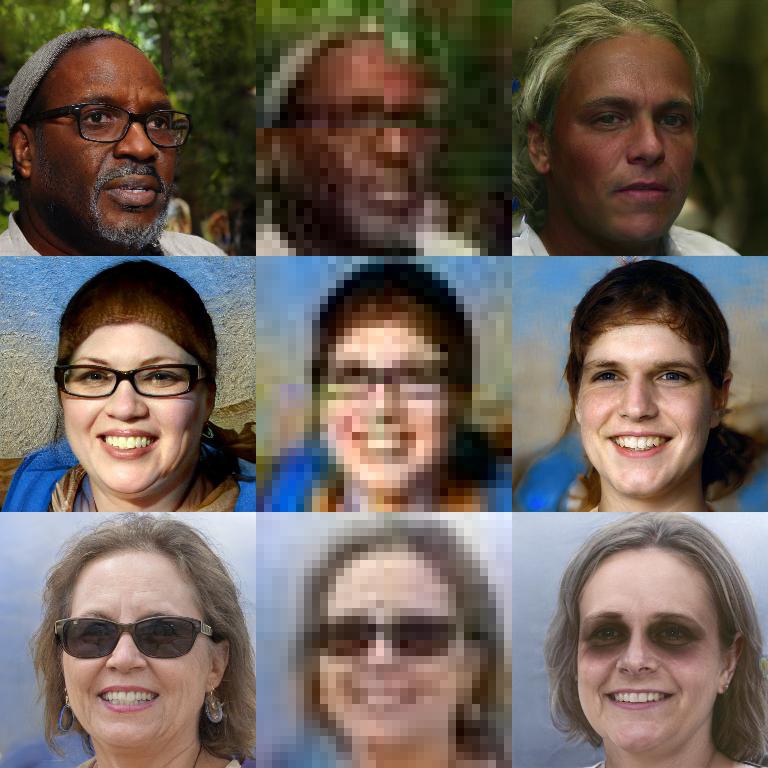}
}
\quad
\subfigure{
\includegraphics[width=5.30cm]{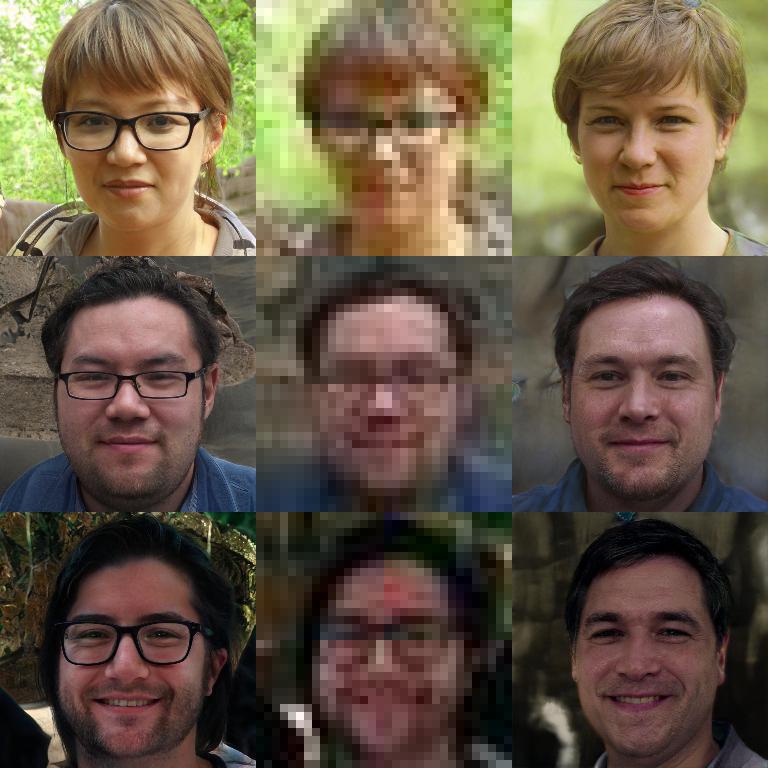}
}
\quad
\subfigure{
\includegraphics[width=5.30cm]{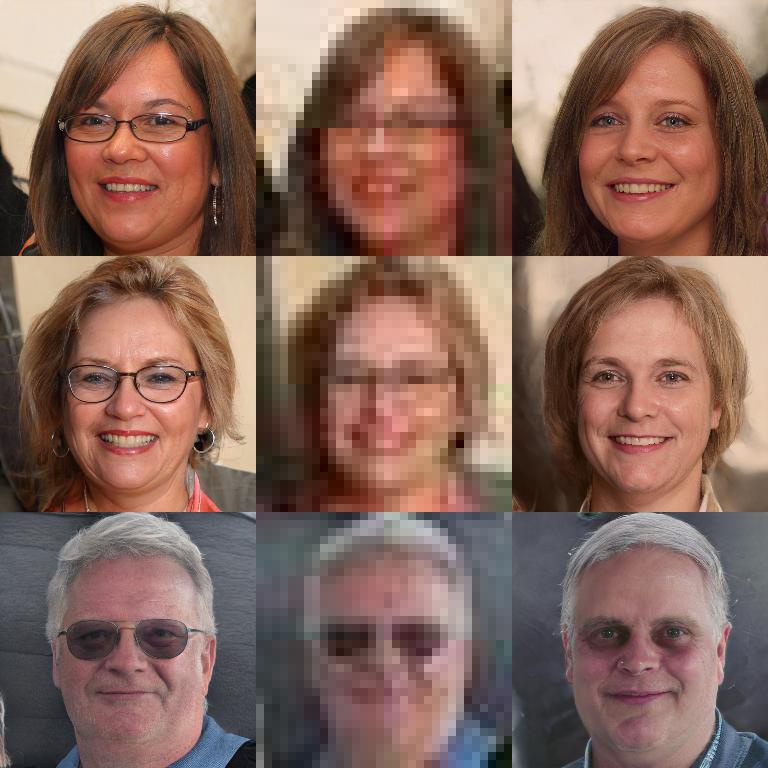}
}
\caption{Examples of Eyeglasses attribute alteration by the super-resolution model.}
\label{fig:PULSE_eyeglasses}
\end{figure}

\begin{figure}[H]
\hsize=\textwidth
\centering
\subfigure{
\includegraphics[width=5.30cm]{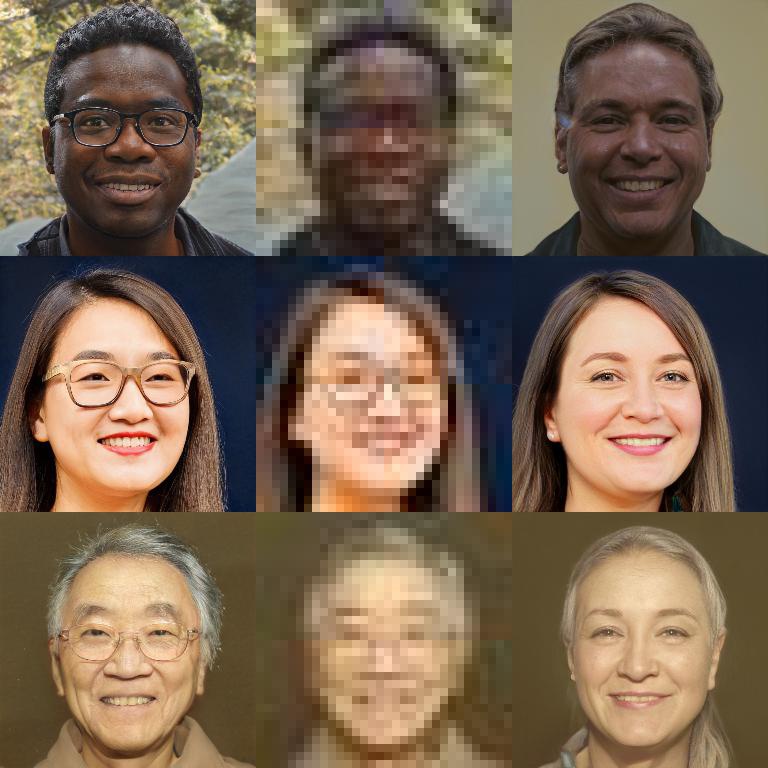}
}
\quad
\subfigure{
\includegraphics[width=5.30cm]{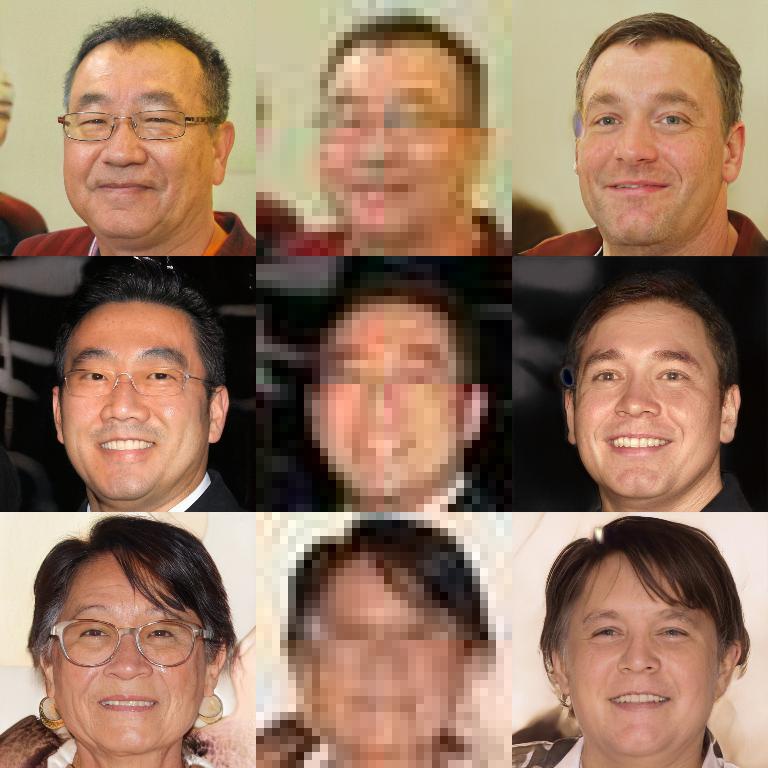}
}
\quad
\subfigure{
\includegraphics[width=5.30cm]{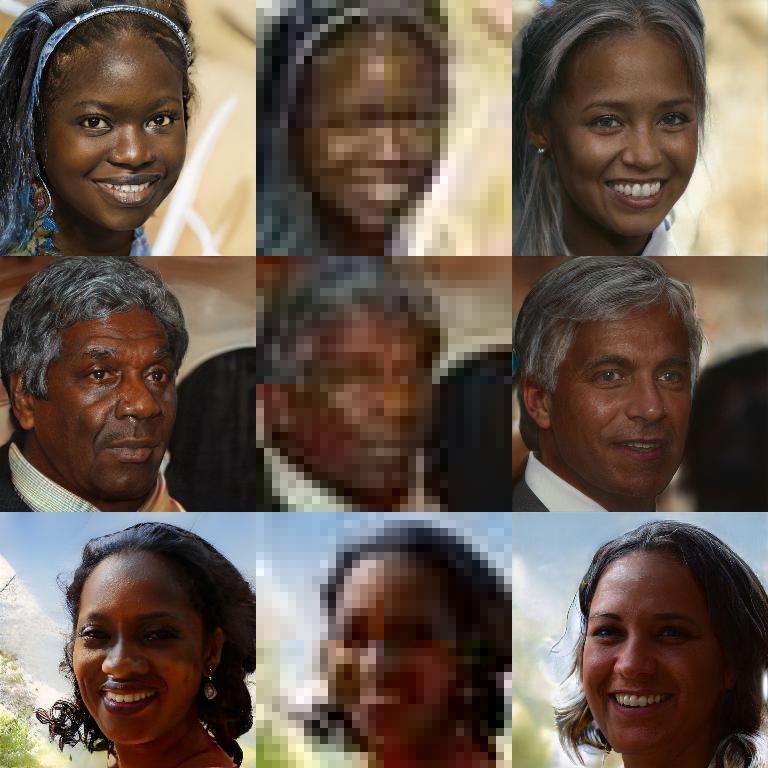}
}
\caption{Examples of Race attribute alteration by the super-resolution model.}
\label{fig:PULSE_age}
\end{figure}

%% file: sections_supp/api_samples_male.tex
\section{Bias in Gender Classification Models}

\setlength{\parindent}{2em}
\par 
In this section, we show more examples of the failure cases of gender classification for the two APIs we show in the paper. Specifically, we show the failure cases where our generated male faces are classified as female and vice versa.

\begin{figure}[H]
\hsize=\textwidth
\centering
\includegraphics[width=17cm]{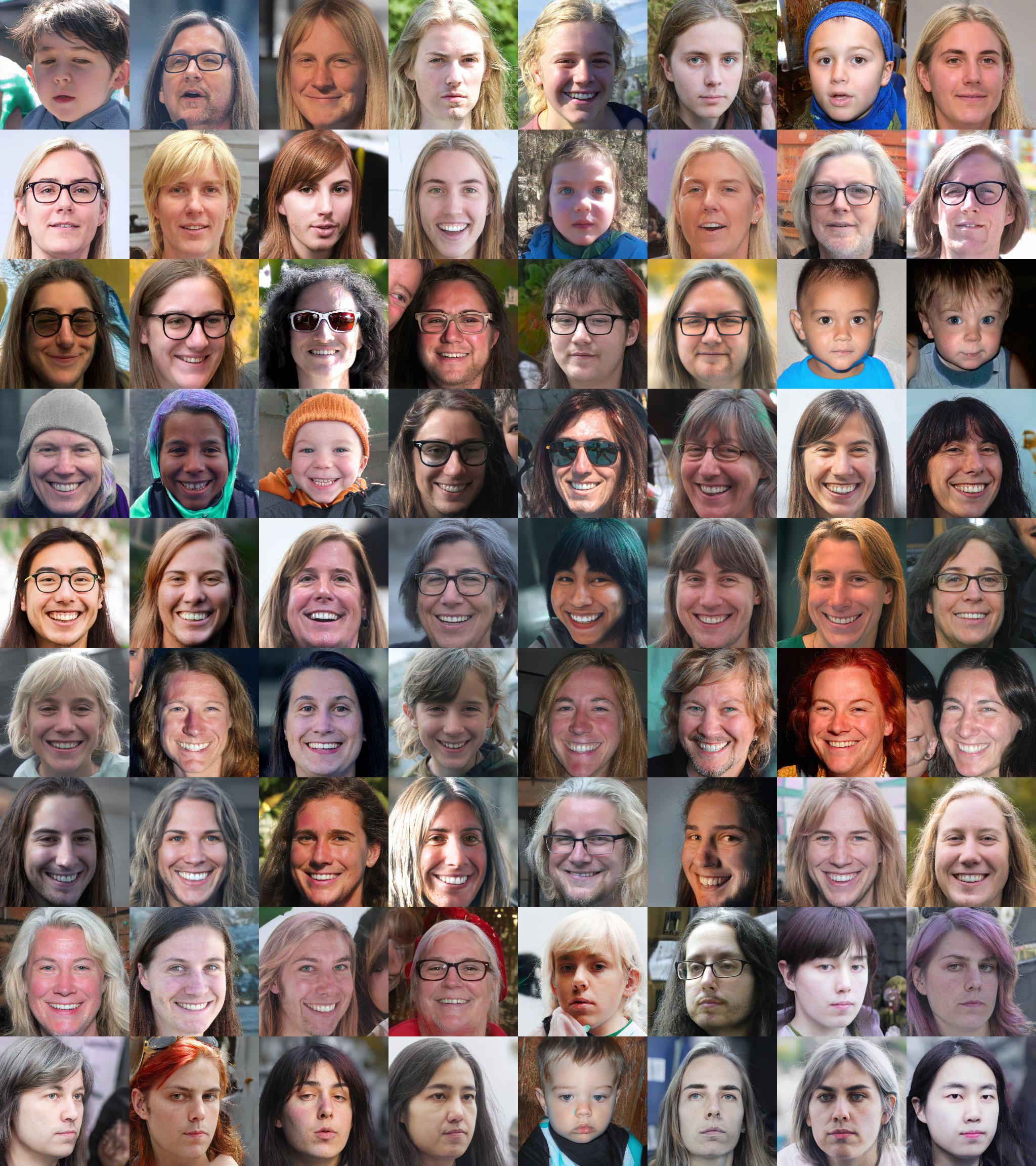}
\caption{Mis-classified Male Images by the Commercial Gender Classification APIs}
\label{fig:dist_plots_male}
\end{figure}

%% file: sections_supp/api_samples_female.tex
\vspace*{\fill} 

\begin{figure}[H]
\hsize=\textwidth
\centering
\includegraphics[width=17cm]{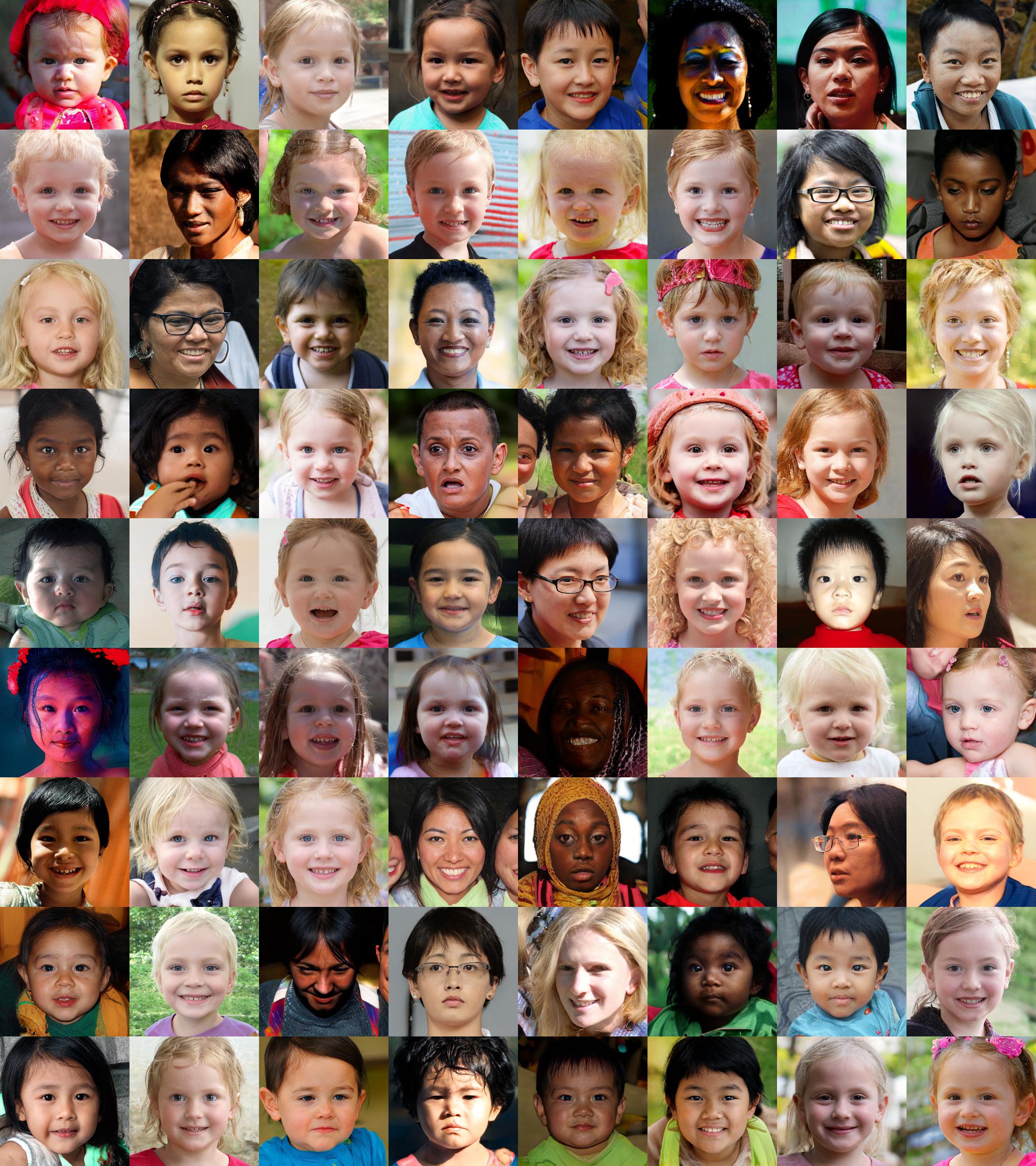}
\caption{Mis-classified Female Images by the Commercial Gender Classification APIs}
\label{fig:dist_plots_female}
\end{figure} 

\vspace*{\fill} 